\documentclass[11pt]{article}

\usepackage{graphicx}
\graphicspath{{.},{../}}

\usepackage{url}
\usepackage{subfigure}
\usepackage{cleveref}
\crefname{section}{Sec.}{Secs.}
\crefname{subsection}{Sec.}{Secs.}
\crefname{subsubsection}{Sec.}{Secs.}
\crefname{figure}{Fig.}{Figs.}
\crefname{subfigure}{Fig.}{Figs.}
\crefname{equation}{Eq.}{Eqns.}
\crefname{table}{Tab.}{Tables}
\crefname{appendix}{Appendix}{Appendices}

\usepackage{amsmath,amssymb}
\usepackage{framed}
\usepackage{array}
\usepackage{authblk}
\usepackage{fullpage}

\newcommand{\bs}{\boldsymbol}

\usepackage[disable]{todonotes}
\usepackage{color}
\def\dangle{{\theta}}
\def\dangletwo{{\varphi}}
\def\p{{p}}

\def\f{{f}}

\def\eone{{\boldsymbol{e}_1}}
\def\etwo{{\boldsymbol{e}_2}}
\def\ethree{{\boldsymbol{e}_3}}

\def\slmpat{\bs{\Phi}}
\def\vecm{\boldsymbol{\sigma}}
\def\elemx{x}
\def\vecx{{\bs{\elemx}}}
\def\vecs{{\boldsymbol{s}}}

\def\vecz{\bs z}

\def\elemspas{\psi}
\def\spasvec{\boldsymbol{\elemspas}}
\def\spasis{\boldsymbol{\Psi}}

\newcommand{\scp}[2]{\langle #1, #2 \rangle}

\def\sensvec{\bs \Gamma}
\def\sensis{\bs \Gamma}

\DeclareMathOperator*{\argmin}{\arg\min}
\DeclareMathOperator{\diag}{diag}

\def\elemalpha{\alpha}
\def\vecalpha{{\bs{\elemalpha}}}
\def\coh{\mu}
\def\senset{\Omega}
\def\vecy{\boldsymbol{y}}
\def\cmplx{\mathbb{C}}
\def\noisevec{\boldsymbol{n}}

\def\noistd{\epsilon}
\def\m{\sigma}

\def\matM{\boldsymbol{\Sigma}}
\def\H{\boldsymbol{H}}

\def\u{u}
\def\vecu{\bs u}

\def\n{{n}}
\def\vecn{{\bs \n}}

\def\matg{\mathbf{G}}
\def\vecg{{\boldsymbol{g}}}

\def\paramtrans{{\bs r}}

\def\prox{\text{prox}}

\def\k{k}

\def\K{K}
\def\matk{\bs K}
\def\M{M}
\def\N{N}
\def\nccd{N_{\rm C}}
\def\nslm{N}
\def\F{F}
\def\G{G}

\newcommand{\loneof}[1]{\|#1\|_1}
\newcommand{\ltwoof}[1]{\|#1\|_2}

\newcommand{\sq}{\vspace{0mm}}

\newcommand{\bb}{\mathbb}
\newcommand{\cl}{\mathcal}
\newcommand{\ie}{\emph{i.e.}, }
\newcommand{\eg}{\emph{e.g.}, }

\newcommand{\realno}{\mathbb{R}}

\title{Compressive Imaging and Characterization of Sparse Light Deflection Maps}

\author[1]{\small P. Sudhakar}
\author[1]{L Jacques}
\author[2]{X. Dubois}
\author[2]{P. Antoine}
\author[2]{L. Joannes}
\affil[1]{ELEN Department, ICTEAM, Universit\'{e} catholique de Louvain, Belgium.}
\affil[2]{Lambda-X, Nivelles, Belgium.}
\date{\small \today}

\begin{document}
	\maketitle
	%\slugger{mms}{xxxx}{xx}{x}{x--x}%slugger should be set to mms, siap, sicomp, sicon, sidma, sima, simax, sinum, siopt, sisc, or sirev
	
	%% Taking care of affiliations %%%
	\renewcommand{\thefootnote}{\fnsymbol{footnote}}
	%\footnotetext[2]{ICTEAM institute, Universit\'e catholique de Louvain, Louvain-la-Neuve, Belgium}
	%\footnotetext[3]{Lambda-X, Nivelles, Belgium}
	\renewcommand{\thefootnote}{\arabic{footnote}}

	\begin{abstract}
		{Light rays incident on a transparent object of uniform refractive
			index undergo deflections, which uniquely characterize the surface
			geometry of the object. Associated with each point on the surface is a
			\emph{deflection map} (or spectrum) which describes the pattern of
			deflections in various directions. This article presents a novel method to
			efficiently acquire and reconstruct sparse deflection spectra induced
			by smooth object surfaces. To this end, we leverage the framework of Compressed Sensing (CS)
			in a particular implementation of a schlieren deflectometer, \ie an
			optical system providing linear measurements of deflection spectra
			with programmable spatial light modulation patterns. In particular, we design those
			modulation patterns on the principle of spread spectrum CS
			for reducing the number of observations. Interestingly, the ability of our device to
			simultaneously observe the deflection spectra on a dense
			discretization of the object surface is related to a particular
			Multiple Measurement Vector (MMV) model. This scheme allows us to estimate
			both the noise power and the instrumental point spread function in a specific calibration procedure.
			
			We formulate the spectrum reconstruction task as the solving of a linear inverse
			problem regularized by an analysis sparsity prior using a translation invariant
			wavelet frame. Our results demonstrate the capability and advantages of using a CS based
			approach for deflectometric imaging both on simulated data and experimental deflectometric data. 
			
			Finally, the paper presents an extension of our method showing how we
			can extract the main deflection direction in each point of the object
			surface from a few compressive measurements, without needing any
			costly reconstruction procedures. This compressive characterization is
			then confirmed with experimental results on simple plano-convex and
			multifocal intra-ocular lenses studying the evolution of the main
			deflection as a function of the object point location. 
		}
	\end{abstract}
	

%	\begin{AMS}\end{AMS}

%	\pagestyle{myheadings}
%	\thispagestyle{plain}
%	\markboth{Compressive acquisition and characterization of sparse light
%		deflection maps}{SIAM Journal of Imaging}
	
\section{Introduction}
\label{sec:intro}

When light travels through a medium of varying refractive index, the refractive index gradient in the direction of light propagation causes light to deflect from its original path. By carefully studying the patterns of deflection, the composition of the medium under consideration can be discovered. A set of imaging techniques, known as \emph{schlieren imaging}, allows us to optically visualise the extent of light deflection~\cite{Settles:2001}. These \emph{schlieren deflectometers} operate by converting deflections into grayscale values and can be employed for any medium (solid, liquid and gases). Applications of schlieren techniques include flow modeling and computer graphics~\cite{Davies198137, Settles19853, Ihrke:2004:ITR}. In this article, we consider an instance of schlieren deflectometer, which is used for characterizing solid transparent objects such as optical lenses. 

Most of the alternative optical modalities for characterizing transparent objects are based on interferometry~\cite{Malacara:2007}. However, {these} techniques are very sensitive to vibrations and also need precise calibrations for the methods to work successfully. On the other hand, thanks to the nature of deflectometry, problems due to vibrations and sensitive calibrations are avoided and hence they make an excellent alternative for industrial deployment in applications such as optical metrology and quality control. 

\begin{figure}[t]
	\centering
	\raisebox{2mm}{%
		\includegraphics[width=.57\columnwidth]{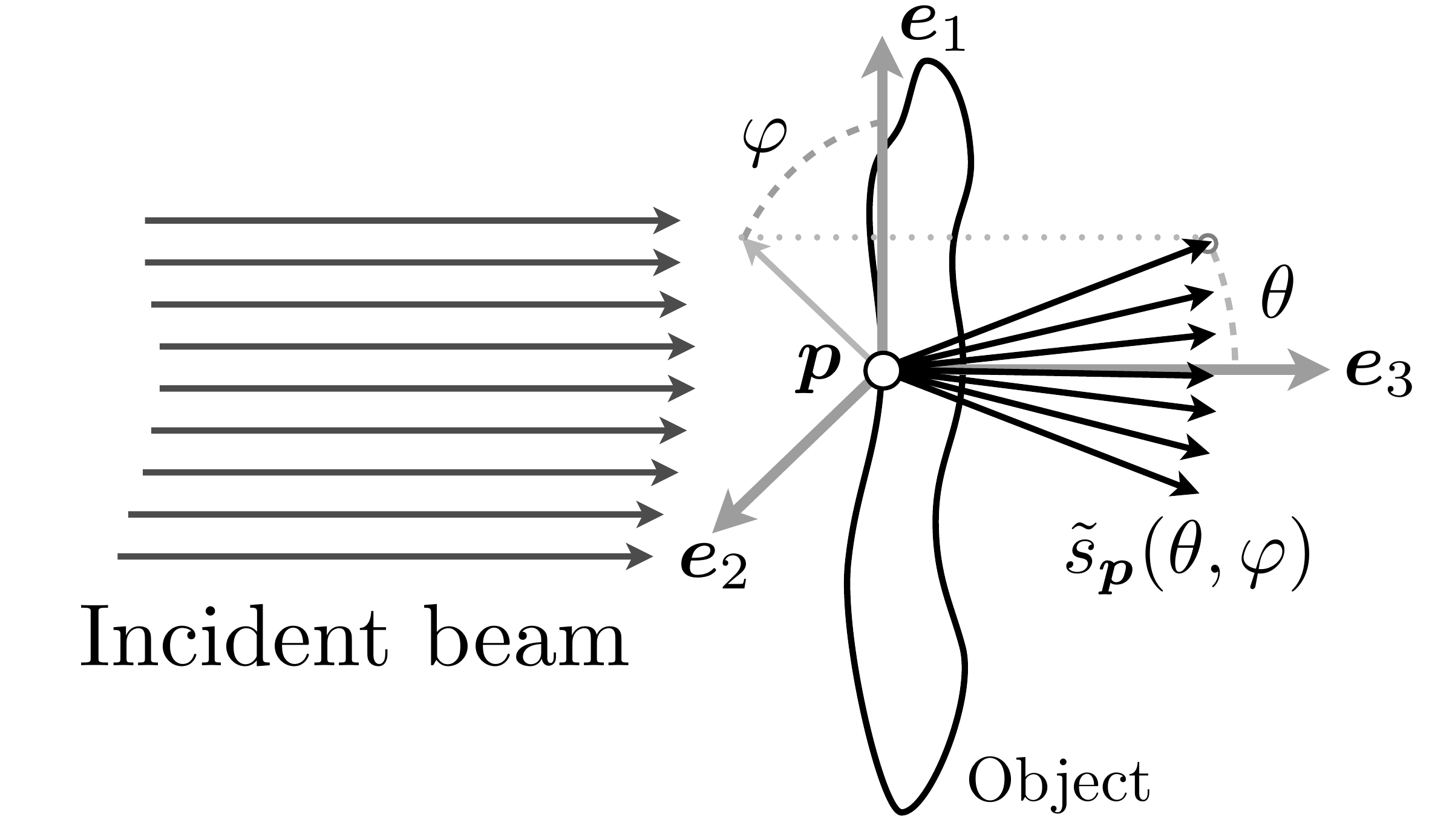}%
	}\!\!
	{
		\includegraphics[width=.4\columnwidth]{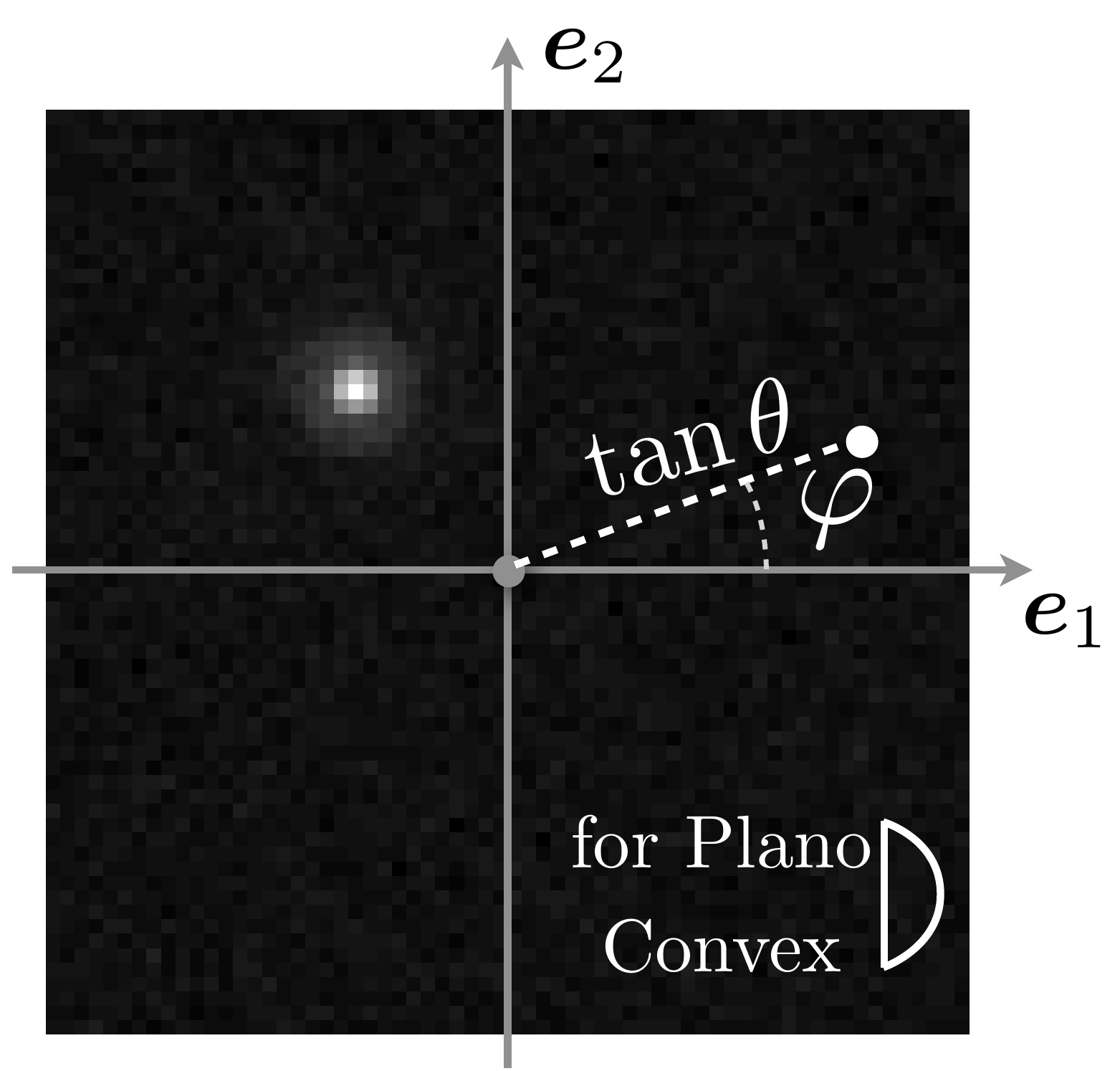}
	}
	\caption{\label{fig:2dspec-fullsetrecon} Left, illustration of a deflection spectrum. Right, a typical (projected) deflection spectrum $s_{\bs p}$ for a plano convex lens of optical power $25.12$D.\sq\sq\sq}
\end{figure}

Consider a (thin) transparent object with a beam of parallel light rays incident on one of its sides, as shown in Fig.~\ref{fig:2dspec-fullsetrecon}(left). At each surface location $\p$ (a non-zero area defined by the spatial resolution of the instrument), light deviates in multiple directions and they can be characterized in a local coordinate system $(\eone, \etwo, \ethree)$, with $\bs e_3$ being parallel to the incident light beam. Using the spherical coordinates $(\dangle, \dangletwo)$ in this system (see~\cref{fig:2dspec-fullsetrecon}(left)), the resulting \emph{deflection spectrum} $\tilde s_{\bs p}(\dangle, \dangletwo)$, which is non-negative in nature, represents the flux of light deviated in each direction $(\dangle, \dangletwo)$. 

As we consider a different location $\bs p'$ on the surface of the object, the local information about its shape is given by the corresponding deflection spectrum $\tilde s_{\bs p'}(\dangle, \dangletwo)$. Therefore, by studying deflection spectra across all the locations on the object, we can understand the overall shape of the object, thereby characterizing the same. 

In this article, the deflection spectrum $\tilde s_{\bs p}$ is conveniently represented by its projection on to the plane $\Pi$ that is normal to $\bs e_3$, \ie according to the projected function $s_{\bs p}(r(\theta),\varphi) = \tilde s_{\bs p}(\theta,\varphi)$ with $r(\theta) = \tan\theta$. Moreover, the object surface is assumed sufficiently smooth to be parameterized by a projection of the location $\bs p$ on to the same plane (with an arbitrarily fixed origin), so that $\bs p$ is basically parameterized as a 2-D vector in the coordinate system $\{\bs e_1, \bs e_2\}$.

For most objects (\eg with smooth surfaces), deflections at any location $\bs \p$ are well behaved and occur in a limited range of angles. The deflection spectra therefore tend to be naturally \emph{sparse} in plane $\Pi_{\bs p}$ or even in some appropriate basis of this domain, such as redundant wavelets. Fig.~\ref{fig:2dspec-fullsetrecon}(right) shows an example of a discretized deflection spectrum $s_{\bs p}$ for one location of a plano-convex lens, obtained using the setup that will be described in~\cref{sec:setup}. The bright spot in the illustrated deflection spectrum signifies that light deflects only in a few directions around a dominant deflection (as governed by classical optics) and deflections elsewhere are negligible.

Measuring deflection angles in a straightforward way using goniophotometer is a cumbersome and elaborate process, and it is only suited for large dynamic of deflection angles and high contrast~\cite{Sauter:1995}. However, in the present context, the deflection information is observed using an indirect method. The particular optical setup at our disposal, which will be described in~\cref{sec:setup}, measures the spectrum indirectly by optical comparison with a certain number of programmable modulation patterns. These optical comparisons are actually modeled as inner products between the modulation patterns and the underlying deflection spectrum. The discretization of the modulation pattern dictates the discretization of the deflection spectrum, and hence the inner products can also be envisaged as being discrete. Furthermore, the optical setup is capable of simultaneously performing the optical comparisons of a single modulation pattern with deflection spectra corresponding to a regular sampling of the test object's surface. These parallel optical comparisons, which are inner-products, are in turn collected in a Charged Coupled Device (CCD) array. By this arrangement, each CCD pixel probes one location on the object, and to reconstruct each spectrum one needs to solve a linear inverse problem from the inner products collected at the corresponding CCD pixel. 

Indirectly probing a signal through its inner products, with known patterns, is  a generalization of the classical sampling procedure where the modulation patterns are simply shifted delta functions~\cite{Baraniuk:2007p171}. Therefore, inner products with known modulation patterns are generalized samples of the signal. From now on, we will simply refer to these inner product samples as \emph{measurements}. In the context of sampling band limited signals using shifted delta functions, the number of samples per unit time needed to reconstruct the signal is dictated by the overall bandwidth of the considered signal space. This is the classical Shannon's sampling theorem. Alternatively, if the underlying signal can be expressed or approximated by a linear combination of a few basis vectors, then results in sparse signal recovery and in \emph{compressed sensing}~\cite{Elad:2010:SRR:1895005, Foucart:2013} show that with appropriate (random) patterns, one can successfully recover such a signal from a fewer number of measurements compared to the dimension of its ambient domain. This signal recovery procedure is non-linear in general and the performance greatly depends on the nature of the modulation patterns. 

In the context of our schlieren deflectometry, we have a situation where the unknown signals (deflection spectra) are sparse and can be observed only through their inner products with programmable modulation patterns. This is {an appropriate} situation for adopting compressed sensing for recovering deflection spectra.

Compressed sensing, despite some difficulty in implementation for optical systems~\cite{Willett:2011vz}, has found several applications in imaging, beginning with the famous \emph{single pixel camera}~\cite{Duarte:2008} to recent applications in, for example, magnetic resonance imaging~\cite{Lustig:2007wf, Lustig:he, Puy:2012p2362}, astronomical imaging~\cite{4703508}, radio interferometry~\cite{Wiaux21052009}, hyper spectral imaging~\cite{6595593} and biological imaging~\cite{Studer:2012wc}. 

\vspace{10pt}
\noindent
\textbf{Contributions and organization of the article:}
The contributions of this article are twofold. Firstly, we demonstrate a novel way to compressively acquire sparse deflection spectra in a schlieren deflectometer and present a numerical method for their reconstruction along with experimental results. To this end, we present a design of optical modulation patterns based on \emph{spread spectrum}\footnote{``Spread Spectrum'' is not related to the studied deflection ``spectrum'' but it refers to the signal frequency spectrum.} compressive sensing~\cite{Puy:2011p1751}. This framework not only enables us to leverage the power of random measurements, as advocated by compressed sensing theory, but also makes the numerical methods exploit the advantage of fast algorithms for matrix-vector multiplications. {The schlieren deflectometer that is used for all the work reported in this article is described in~\mbox{\cref{sec:setup}}.}

{At the outset, this paper does not provide a detailed theoretical treatment of the imaging process. However, the objective of this paper is to demonstrate how compressed sensing can be used to access detailed deflection information in an efficient manner, which is impossible with the existing Phase Shifting schlieren (PSS) technique, described in~\mbox{\cref{sec:pss}}.}

{Of course, in a brute force way, one can sift through each possible deflection direction and collect information, which amounts to performing Nyquist sampling of the deflection spectrum. This is very inefficient both in terms of the number of SLM patterns required and also the limited light throughput. Fortunately, due to the reason that deflection spectra are sparse for a large class of interesting optical objects, guided by compressed sensing principles, we can afford to reduce the number of SLM patterns.}

{We rely on spread spectrum compressed sensing for selecting suitable SLM patterns.~\mbox{\cref{subsec:cs}, \cref{subsec:sscs}} and \mbox{\cref{sec:opticalSensing}} provide the development of sensing SLM patterns and their adaptation to make them amenable for optical implementation. As described in the beginning, CCD pixel simultaneously collect measurements (of corresponding underlying spectra) for a given SLM pattern. This process is mathematically formalized as a Multiple Measurement Vector (MMV)~\cite{mishali:mmv} model in~\mbox{\cref{sec:opticalSensingnoises}}. This formalism greatly aids in modeling different sources of noises and also numerically estimate the quantities associated with them.}

{Even though we have remarked that spectra of smooth objects are sparse in its inherent domain, much can be gained by considering its sparse representation in an overcomplete dictionary such as undecimated wavelets~\cite{Mallat:2008:WTS:1525499}. As opposed to the classical synthesis based sparsity priors, we rely on analysis sparsity approach for solving the inverse problem~\cite{Elad:2007ub}, as described in~\mbox{\cref{sec:problemformulation}}.} In~{\cref{sec:numericalmethod}}, we detail the numerical method, based on a primal-dual approach, used to solve the three problems.

\cref{sec:recoveryResults} contains the experimental results, both on simulated data and actual data obtained using the deflectometer. Our experimental results show the potential of compressed sensing based deflectometric imaging in reducing the number of measurements. 

A description of the optical setup and a {preliminary} reconstruction method, along with the results, were briefly presented in our conference communications~\cite{Sudhakar:2012uk, Sudhakar:2013uk}. However, the present article contains several new material including the details of modulation pattern {selection}{, analysis of noise and its numerical estimation}, along with new contributions on how to use deflection information for object characterization. 

The second part of the contribution deals with the local characterization of lenses using compressive measurements, such as mapping local dioptric power (related to the local focal length) across the surface of the lens. The deflection spectrum provides detailed information about the deflection pattern at a particular location on the object. However, the price to be paid to obtain such rich information is the computation effort spent on its numerical reconstruction. When computational resource is scarce, we can still extract meaningful information about the deflection spectra, sufficient to characterize the shapes of objects under consideration. This work is on the lines of compressed domain signal processing and parameter estimation~\cite{Davenport:2007p2548, Davenport:2010p589, Fyhn:2013vk}.  

We develop a simplified description of deflection spectrum, which is characterized by a pair of translation parameters. This is inspired by the fact that when the surface of an object is smooth, the deflection spectra at consecutive locations tend to have similar shape and size but  translated on the spectral plane. Therefore, we propose a \emph{compressed domain matched filtering}, in~\cref{sec:CPE}, to extract these parameters directly from the measurements, without involving any expensive reconstruction method. 

A method for estimating parameters in compressed domain, up to sub-pixel accuracy, is presented and comparison is made with the parameters estimated from reconstructed deflection spectra. We then describe the relationship between the optical power of the object under consideration and the evolution of the parameters, using the schematic of the optical system. Our experiments show the capability of compressive method to provide relevant characterization of simple test objects such as plano-convex lenses and also complicated objects such as diffractive multifocal intra-ocular lenses~\cite{Klein:1993}, \ie lenses displaying multiple foci due to their special design of surfaces.

{In summary, this article presents an entire system level design of a compressive imager, able to simultaneously acquire a large number of deflection spectra. Our approach includes mathematically concrete procedures for tuning the design parameters and also presents concrete ways to utilize deflection information to characterize optical properties of transparent objects.} 

\medskip
\paragraph*{Conventions} We find useful to introduce here some of the mathematical notations used throughout the paper. Most of domain dimensions are denoted by capital roman letters, \eg $M, N, \ldots$ Vectors and matrices are associated to bold symbols, \eg $\bs \Phi \in \bb R^{M\times N}$ or $\bs u \in \bb R^M$, while lowercase light letters are associated to scalar values. The $i^{\rm th}$ component of a vector $\bs u$ reads either $u_i$ or $(\bs u)_i$, while the vector $\bs u_i$ may refer to the $i^{\rm th}$ element of a vector set. The vector of ones in $\bb R^D$ is denoted by $\bs 1_D = (1, \cdots, 1)^T$, with $(\cdot)^T$ denoting the transposition of a matrix. The set of indices in $\bb R^D$ is $[D] = \{1, \cdots, D\}$. The cardinality of a set $\cl C$, measuring the number of elements of the set, is denoted by $|\cl C|$. The (convex) indicator function $\imath_{\cl C}(\bs x)$ of the set $\cl C$ is equal to $0$ if $\bs x\in \cl C$ and $+\infty$ otherwise. For any $p\geq 1$, the $\ell_p$-norm of $\bs u$ is $\|\bs u\|_p^p = \sum_i |u_i|^p$. The Frobenius norm of $\bs A$ is given by $\|\bs A\|_F^2 = \sum_i \sum_j |A_{ij}|^2$. The support of a vector $\bs u \in \bb R^D$ is defined as $\textrm{supp} \ \bs u = \{ i \in \mathbb{N}^D : u_i \neq 0 \}$.

% % % %
\begin{figure}[t]
	\centering
	\subfigure[\label{subfig:pss}]{
		\includegraphics[width=0.75\textwidth]{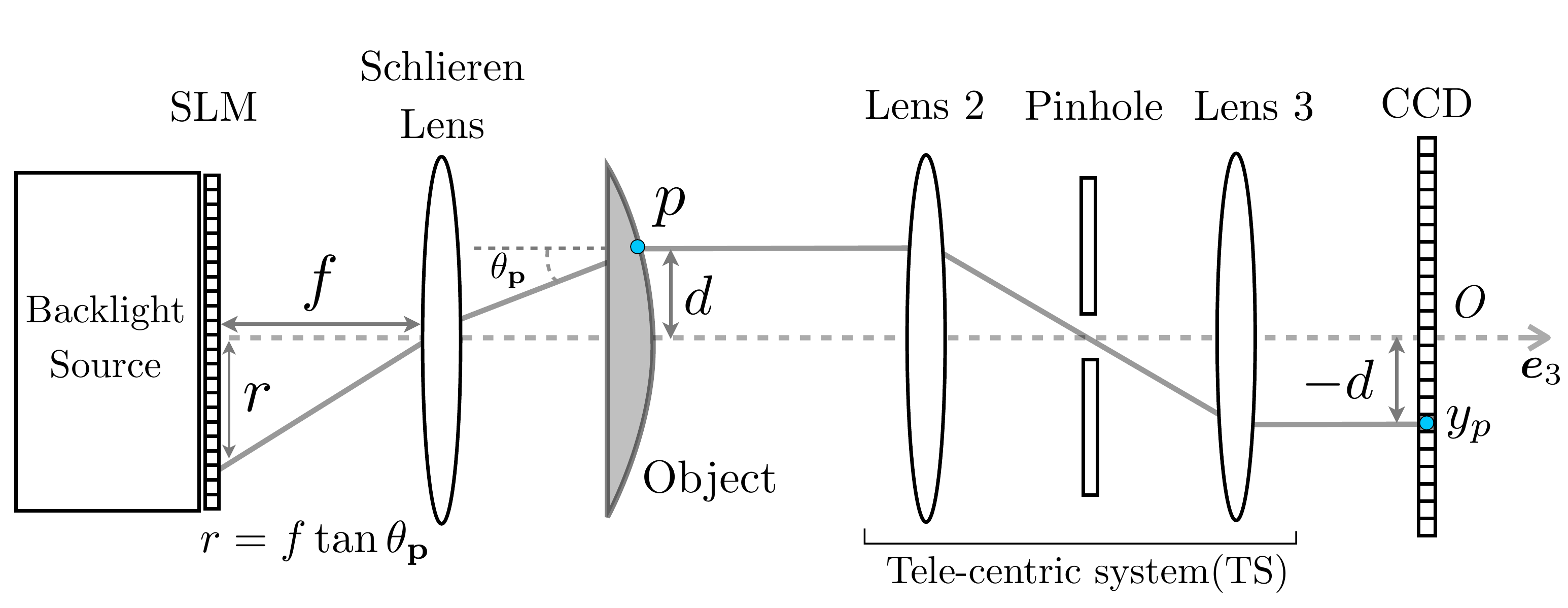}\sq
	}
	\subfigure[\label{subfig:nimo}]{
		\includegraphics[width=0.19\textwidth]{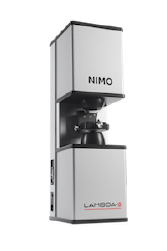}\sq
	}
	\caption{\label{fig:pssetnimo} \subref{subfig:pss} 2D schematic of schlieren deflectometer and \subref{subfig:nimo} a commercially available NIMO\texttrademark schlieren deflectometer.}
\end{figure}

\section{Optical setup and notations}
\label{sec:setup}

The schematic of the schlieren deflectometer used in our work to measure deflection spectra is shown in~\cref{subfig:pss}, which is a simplified 2D representation associated to one 2D slice.~\cref{subfig:nimo} shows a picture of a commercially available schlieren deflectometer sold under the name of NIMO\texttrademark  by Lambda-X. Its key components are \emph{(i)} a Spatial Light Modulator (SLM) based on {the Liquid-crystal Display (LCD) technology ($1024 \times 768$ pixel grid, 8 bits per pixel)}, \emph{(ii)} a schlieren lens with focal length $\f$, \emph{(iii)} a Telecentric System (TS) and \emph{(iv)} a Charged Coupled Device (CCD) camera to collect light {($1296 \times 966$ pixel grid, 14 bits per pixel)}. 

The object to be analyzed is placed in between the schlieren lens and the telecentric system. On its left side, light rays from a {incoherent backlight source (12W, $\lambda=546$ nm)} are incident. Due to the telecentric system, only those light rays emerging out of the object that are parallel to the optical axis pass through and get collected by the CCD. Up to a coordinate system inversion around the optical axis $O$, each location $\bs \p$ on the object, at a distance $d$ from the optical axis $O$ (dashed line), is probed by a corresponding CCD pixel also at a distance of $d$ from $O$. Each location $\bs \p$ is thus in one-to-one correspondence with a CCD pixel and therefore, the spatial resolution of the system is dictated by the resolution of the CCD array and the size of the pinhole of the telecentric system. 

From classical optics, a light ray that is incident on location $\bs \p$ at an angle $\dangle_{\bs \p}$ originates from the light source at a distance of $r = \f\tan\dangle_{\bs \p}$ from the optical axis. Likewise, the light rays originating from different locations of the source have different incident angles at $\bs \p$. Since we can always invert the direction of light propagation virtually in an optical system, we can view the system as though parallel rays of light were incident from the right side of the object, and the light undergoes deflections at $\bs p$ on the object, and they form an image on the SLM. Therefore, up to a global scaling by $f$, the SLM plane is exactly the plane $\Pi$, that was discussed in~\cref{sec:intro}, on which the deflection spectra corresponding to every location $\bs p$ is observed. With this arrangement, modulating the SLM amounts to modulating the deflection spectrum $s_{\bs p}$, while the light collected in CCD pixel $\bs p$ is just an inner product of $s_{\bs p}$ with the modulation pattern. 

Even though deflection spectrum is a continuous domain object, the use of a discrete SLM, and hence discrete modulation patterns, renders the inner products to be discrete in nature, \ie we observe the continuous spectrum through a discrete representation where each sample is associated to the integration of light in a single SLM pixel. Hence, the recoverable deflection spectra is also discrete and has the same resolution as the SLM. Similarly, the discreteness of the CCD also limits the spatial resolution of the locations that can be probed. Henceforth, we shall use the notation $k$ to indicate the discrete locations on the object and the corresponding CCD pixel locations. 

Even though the deflection spectrum and the modulation patterns are 2D quantities, we shall represent them as 1D vectors for brevity of notation so that the action of modulation patterns on a deflection spectrum can be written as a matrix vector product. The 1D vector can be obtained by simply stacking all the columns of the 2D representation. If we generate $\M$ modulation patterns $\phi_{i}\in\bb R^\nslm$ with $1\leq i \leq \M$ in the SLM of $\N$ pixels, considering the discrete nature of the CCD camera (having $\nccd$ pixels), the discretized deflection spectra are observed through\sq 
\begin{equation}
\label{eq:schlieren-forward-model}
\bs y_k = \cl C(\bs \Phi \bs s_k),\quad 1\leq k \leq \nccd,\vspace{-.5mm}
\end{equation}
where $\bs \Phi^T = (\bs \phi_1, \cdots, \bs \phi_M) \in \bb R^{N \times M}$ is the \emph{sensing matrix}, $k$ is the CCD pixel index, $\bs s_k\in\bb R^N$ is the discretized spectrum at the $k^{\rm th}$ pixel/object location, and $\cl C:\bb R^M \to \bb R^M$ models a corruption of the measurement process, \eg by shot/photon noise or by additive signal or measurement noises of finite power (see Sec.~\ref{sec:sscs}). 

For each modulation pattern, the CCD collects one measurement of the underlying corresponding deflection spectrum. The system is designed in such a way that the CCD pixels are independent of each other and collect only the samples of its corresponding deflection spectrum. The {next step is to choose a suitable} {sensing matrix} $\bs \Phi$ in order to maximize the information captured in each sample. To this end, we {select} the modulation patterns relying on the theory of spread spectrum compressed sensing~\cite{4960224, Puy:2011p1751}. As the forward measurement model in Eq.~\eqref{eq:schlieren-forward-model} and the recovery problem formulation (developed in future sections) are the same for all the locations of the CCD, we suspend the use of the subscript $\k$ to simplify the notations, and resume the usage when the situation demands.

\section{Phase shifting schlieren}
\label{sec:pss}
The schlieren deflectometer which we consider in our work has been already used, outside the context of compressive sensing, to measure light deflections in transparent objects. For the purpose of emphasizing the advantage of the compressive sensing method, we shall briefly explain the existing method. For a detailed description of the same, the interested readers can consult~\cite{Joannes:2003p1444, Joannes2004}. The system is configured to work using several phase shifted sinusoidal patterns (for modulation) in order to measure deflections and this configuration is named Phase Shifting schlieren (PSS). In PSS, the deflection information is directly encoded into the intensity of the CCD through sinusoidal modulation. 

The key assumption in PSS is that the deflection spectrum for each location consists of a single peak. {Mathematically, if $(x_1,x_2)$ represents the coordinates in the SLM plane $\{\bs e_1, \bs e_2\}$ orthogonal to the optical axis $\bs e_3$ (see Fig.~\ref{fig:pssetnimo}) and if we do not consider the impact of the SLM/CCD discretization for the sake of simplicity, a spectrum of unique deflection direction $(r_1, r_2)^T \in \bb R^2$ observed on object point $\k$ is modeled as
	$$
	s_k(x_1,x_2) = a\,\delta^{(2)}(x_1 - r_1, x_2 - r_2),
	$$
	with $\delta^{(2)}$ the Dirac distribution and given an amplitude $a>0$.} The PSS measurements are then made using multiple phase shifted sinusoidal modulations in both horizontal and vertical directions, for making the measurements. For example, {a~vertical sinusoidal pattern with period $\Lambda$ and phase $\xi_1$ is given by 
	\begin{equation}
	h^{\rm v}_{\xi_1}(x_1, x_2) = 1+\sin(\tfrac{2\pi}{\Lambda}x_1+\xi_1).
	\label{eqn:sinupatt}
	\end{equation}
	With such a pattern and suitable approximations of the optical system, the intensity recorded on a CCD pixel at location $\k$ in the absence of noise reads
	\begin{equation}
	y^{\rm v}_\k(\xi_1) = \scp{h^{\rm v}_{\xi_1}}{s_k} = a+a\sin(\tfrac{2\pi}{\Lambda} r_1+\xi_1).
	\label{eqn:pssmeas}
	\end{equation}
	The PSS method consists then in measuring $y^{\rm v}_\k$ over different $\xi_1$, in a \emph{phase shifting} stage, and extracting the values $r_1$ from Eq.~\eqref{eqn:pssmeas} by using the $n$-step algorithms~\cite{Almazan:2003,Hariharan:87}. The procedure is repeated with sinusoids of different periods in order to remove the implicit modulo ambiguity over $r_1$, while remaining ambiguities are removed with spatial unwrapping procedures \cite{bioucas2007phase,gonzalez2014robust}. The vertical component $r_2$ of the displacement is recovered similarly by phase shifting of horizontal sinusoidal patterns.} 

PSS is a simple and effective method for measuring deflection angles and has been successfully used in tomographic applications such as refractive index map reconstruction~\cite{Beghuin:2010p2208, AGonzalez:2012}. However, this method is not stable for deflection spectra that are spread out, unlike a point or even of estimating several main deflection angles for each object location. Therefore, the richness of deflection spectra to reveal interesting information is lost. Moreover, the use of non-binary modulation patterns in Eq.~\eqref{eqn:sinupatt} brings in the problem of non-linearity in the SLM response and without a careful calibration, it is prone to computational errors. Our proposed method is capable of recovering full deflection spectra and also it uses binary modulation patterns to avoid SLM non-linearities.

% %

\section{{Spread spectrum compressive schlieren deflectometry}}
\label{sec:sscs}

{The program\-mable nature of the schlieren deflectometer described in Sec.~\ref{sec:setup} permits its transformation in a compressive imager of light deflection spectra. This section explains how we adjust the sensing procedures provided by the Compressed Sensing theory to an actual sensing scheme respecting our optical constraints. Interestingly, we also show that the ability of our device to simultaneously observe the deflection spectra of each object surface point, as discretized by the CCD pixel grid, is related to a particular Multiple Measurement Model (MMV) \cite{mishali:mmv} of those spectra.  This allows us to establish a specific calibration procedure for studying the main (additive) sources of noises in our compressive observations and for estimating the point spread function (PSF) of the instrument.}

\subsection{Compressive sensing}
\label{subsec:cs}
Compressive Sensing (CS)~\cite{Donoho:2006p558, Candes:2006tc, Baraniuk:2007p171, Kutyniok:2012p1804, Foucart:2013} is a new paradigm in signal sampling which envisages that any structured (sparse) signal $\vecx=\spasis\vecalpha\in\cmplx^\N$ that can be decomposed or compressed with only few important coefficients in a basis $\bs\Psi$, can be tractably recovered from a few corrupted linear measurements of the form
\begin{equation}
\label{eqn:csmeasurement}
\vecy =\bs\Phi \vecx\; + \;\noisevec = \bs\Theta \vecalpha\; + \;\noisevec,
\end{equation}
where $\bs\Phi$ is a $\M\times\N$ {\em measurement (or sensing) matrix}, $\bs\Theta = \bs \Phi\spasis$ and $\bs n$ is an additive noise such that $\|\bs n\|_2\leq \epsilon$ for some known bound $\epsilon>0$. For instance, if $\bs \Phi$ is a random Gaussian matrix with $\bs\Phi_{ij} \sim_{\rm iid} \mathcal N(0,1/M)$ and if the number of measurements $M$ is of the order of $\K\log(\N/\K)$, then, with high probability over any possible signal $\bs x$, the solution $\widehat\vecalpha$ of the following convex optimization problem 
\begin{equation}
\label{eqn:bpdn}
\widehat{\bs\alpha} := \argmin_{\vecalpha\in\cmplx^\N}\;\loneof{\vecalpha}\;\text{subject to}\; \ltwoof{\vecy - \bs\Theta 
	\vecalpha}\leq\noistd,
\end{equation}
satisfies
\begin{equation}
\ltwoof{\vecalpha - \widehat{\vecalpha}} = O\left(\tfrac{\loneof{\vecalpha - \vecalpha_\K}}{\sqrt{\K}}\;+\;\noistd\right),
\label{eqn:bpdnreconerr}
\end{equation} 
with $\vecalpha_\K $ is the best $\K$-term approximation of $\vecalpha$~\cite{candes2007sparsity, Rauhut:2010p2556}.

Similar type of result also holds when the measurement matrix is derived from an orthonormal basis $\sensis\in\cmplx^{\N\times\N}$. In such a case, a $\M\times\N$ measurement matrix is of the form $\sensis_\senset^\ast$, the conjugate transpose of the submatrix matrix $\sensis_\senset$ formed by restricting the columns of $\sensis$ to those index by subset $\senset\subset [N] :=\{1,\,\cdots, N\}$ randomly chosen at uniform. If the {\em coherence} $\coh :=\sqrt{\N} \max_{1\leq i,j\leq \N} |\langle \sensvec_j, \spasvec_i\rangle|\in[1,\sqrt N]$ between $\sensis$ and $\spasis$
is very close to~1, then 
\begin{equation*}
\M = O\big(\coh^2\K\log^4(\N)\big)
\end{equation*}
measurements are enough for the optimization problem (with $\bs\Theta=\sensis_\senset^\ast\spasis$ and $\slmpat = \sensis_\senset^\ast$) to provide an estimate $\widehat\vecalpha$ that satisfies Eq.~\eqref{eqn:bpdnreconerr}~\cite{candes2007sparsity, Rauhut:2010p2556}. The number of the measurements for successful recovery scales quadratically with respect to the coherence $\coh$ and hence it is desirable to have the two bases as incoherent as possible to make $\coh$ close to~$1$.

While fully random matrices are optimal in terms of sampling efficiency, measurement matrices derived from orthonormal bases are often equipped with fast matrix-vector multiplication making them attractive for practically solving Eq.~\eqref{eqn:bpdn}. In the next section, we select a method that achieves a trade-off between optimal random measurement strategy and structured computations.

\subsection{Spread spectrum sensing}
\label{subsec:sscs}
The key issue in using orthonormal bases as measurement matrices is that of its {possibly high} coherence with the signal sparsity basis. In order to statistically minimize coherence between the two bases, we employ \emph{spread spectrum} modulation of the data vector $\vecx$~\cite{4960224, Puy:2011p1751}. 

Spread spectrum randomly scrambles the phases of signal by point-wise multiplication of the signal samples with another modulator signal whose samples have random phases but unit magnitude. If the sensing basis is a Fourier or Fourier-like, then this point-wise modulation amounts to performing a convolution of the signal and the modulator in the spectral domain. As a result of this, the energy of the signal is spread over its entire spectrum, while preserving its norm, and hence the name ``spread spectrum". 

Further, whenever all the entries of the sensing basis $\bs \sensis$ have the same amplitude, then with high probability, spread spectrum technique ensures that the number of measurements required for successful reconstruction of the signal is comparable to that of fully random measurements, the optimal measurement strategy according to compressive sensing. Such bases are called \emph{universal} sensing basis and Fourier and Hadamard bases are some examples of them. 

Mathematically, the spread spectrum vector is a random vector $\vecm \in \bb C^N$ with the amplitudes of each of its entry $|\m_i|=1$, \eg a Steinhaus or Rademacher sequence. The sensing matrix that incorporates modulation by $\vecm$ is $\bs\Phi = \sensis_\senset^\ast\matM$, where $\matM=\diag(\vecm)$ is a diagonal matrix. In this case, we need\sq
\begin{equation*}
\M\geq C_\rho\,\K\log^5(\N) \sq
\end{equation*}  
measurements in order to recover a solution $\vecalpha^\star$ of~\eqref{eqn:bpdn} satisfying~\eqref{eqn:bpdnreconerr} with a probability at least
$1-O(\N^{-\rho})$, for some $0<\rho<\log^3(\N)$. Noticeably, the coherence $\mu$ has disappeared from the condition implying that with spread spectrum and universal sensing basis, the recovery guarantee is {\em universal}, irrespective of the sparsity basis. From the perspective of computations, spread spectrum involves only point-wise multiplication and hence does not break down the structure of the fast algorithms used for the transforms. 

On a related note, some researchers have investigated other ways of designing structured random sensing matrices. Notable of them are the ones based on convolution with random sequences~\cite{Haupt:uy, Tropp:2006tq, romberg2009compressive, Yin:2010wq}, where the signal is convolved with a random sequence before subsampling. Even this scheme is shown to be universal and works well with any sparsity basis~\cite{romberg2009compressive}. Convolutions can be implemented as multiplication in the Fourier domain and hence they are computationally efficient too. However, as our application is for an optical system, we would like to have a sensing matrix that is binary valued in nature to avoid non-linearities of the system and hence we stick to spread spectrum. 

% %
\subsection{Adaptation to optical compressive sensing}
\label{sec:opticalSensing}

As the sensing operation has to be finally implemented in a physical deflectometric system, it is essential to have the sensing basis and the spread spectrum vector $\vecm$ to be non-negative real valued, and also preferably binary in nature to avoid optical non-linearities. 

{
	A natural choice for a Fourier-like basis having binary valued entries is the Hadamard basis {$\sensis=\H \in \{\pm 1/\sqrt N\}^{N\times N}$}, which is also a universal basis as remarked earlier~\cite{Tsaig:2006}. As the modulation vector $\vecm$ should be real valued and having unit magnitude in each of its entry, they are chosen randomly from $\{\pm1\}$ with equal probability. The spread spectrum sensing matrix acting on the signal domain could be composed as
	\begin{equation}
	\bs \Phi_{\rm ss} = \H_\senset^T\matM,
	\label{eqn:senspatt}
	\end{equation} 
	where as before $\senset$ selects a random subset of $\M$ sensing vectors in $\bs H$ and $\matM = \diag(\vecm)$.}

{As the resulting matrix $\bs \Phi_{\rm ss}$ belongs to $\{\pm 1/\sqrt N\}^{\M\times\N}$, it has to be properly \emph{biased} to make it non-negative so that it can be implemented as a on-off pattern in the SLM \cite{Studer:2012wc,Willett:2011vz}. Additionally, a scaling of the matrix by $\sqrt N$ is also required in order to fix the coding of the SLM pixel conventionally\footnote{Associating the values 0 and 1 to an opaque and to a fully transparent SLM pixel, respectively.} in $\{0,1\}$. The \emph{optical} sensing matrix becomes then 
	\begin{equation}
	\label{eqn:sensingbias}
	\bs \Phi_{\rm opt} = \tfrac{1}{2} \left(\sqrt N \bs \Phi_{\rm ss} + \bs 1_M\bs 1_N^T\right),
	\end{equation} 
	where $\bs 1_D$ is a $D$ dimensional vector of all ones. {Notice that when $\M=\N$, \ie $\Omega = [\N]$, $\bs \Phi_{\rm opt}$ is invertible only if $\sigma_1 = 1$, with 
		\begin{equation}
		\label{eq:inverse-phi}
		\bs \Phi_{\rm opt}^{-1} = \tfrac{1}{\sqrt N}(2 \bs I_{N} - \bs c_1\bs 1_N^T)\bs \Phi_{\rm ss}^T,   
		\end{equation}
		otherwise $\bs c_1=(1,\,0,\,\cdots, 0)^T \in \bb R^N$ is a non-trivial element of its kernel. For all the experiments reported in this article, the spread spectrum vector had $\sigma_1$ = 1.}  
	
	In a noiseless setting, the optical bias introduce above can be easily removed in our schlieren deflectometer by completing the compressive observations of a spectrum $\bs s_k$ in pixel $k$ with one associated to a pattern $\bs \phi_{\rm on} = \bs 1_N$, \ie simulating the observation of $\bs s_k$ with the sensing $\bs \Phi_{\rm ss}$.  Indeed, we verify easily that 
	\begin{equation}
	\label{eq:debiased-sensing}
	\bs z_\k\ =\ N^{-1/2}\,(2\,\bs \Phi_{\rm opt} \vecs_\k\ -\ (\bs \phi_{\rm on}^T \vecs_\k) \bs 1_M )\ =\  \bs \Phi_{\rm ss} \vecs_\k.  
	\end{equation}
	As will be clear in the next section, the sensing noise in the model \eqref{eq:schlieren-forward-model} that corrupts both $\bs \Phi_{\rm opt} \vecs_\k$ and $\bs \phi_{\rm on}^T \vecs_\k$ impacts this bias removal. Actually, the sensing noise impacting $\bs z_\k$ is bigger than the one present in $2N^{-1/2}\bs \Phi_{\rm opt} \vecs_\k$ and it could seem then more appropriate to base the reconstruction of $\bs s_k$ on these last observations rather than on $\bs z_k$. However, we found empirically that the reconstruction procedure detailed in Sec.~\ref{sec:problemformulation} is much more stable\footnote{We have also reproduced the same phenomenon on 1-D sparse signal reconstruction with unbiased and positively biased Bernoulli sensing matrices using a different reconstruction solver included in the SPGL1 toolbox \cite{spgl1:2007}.} when realized from $\bs z_k$, \ie for $\bs \Phi = \bs \Phi_{\rm ss}$, despite the additional noise introduce by the extra observation of $\bs \phi_{\rm on}^T \vecs_\k$. We must also acknowledge the fact that $\bs z_k$ is based on $M+1$ observations of $\bs s_k$ and not $M$, but we will omit this point as $M \gg 1$ in all our experiments.}

\subsection{Noisy sensing model}
\label{sec:opticalSensingnoises}

{Our schlieren deflectometer described in Sec.~\ref{sec:setup} is obviously corrupted by several sources of noise. These are due, for instance, to backlight source inhomogeneities, light leakage in SLM (\eg due to incomplete pixel fill factor), optical stray light effect, CCD quantization/readout noise and photon or shot noise in the CCD pixels collecting the optical compressive measurements. Our sensing model must account for these noises and bound their impacts in order to define a robust reconstruction of deflection spectra. However, as will become clearer below, we can summarize the impact of all these by a global additive noise model both on the signal and on the measurements.}

{\paragraph*{Limited influence of the photon noise} As noted in \cite{Studer:2012wc} in the general case of compressive optical observations of sparse images, the bias introduced in \eqref{eqn:sensingbias} for ensuring a non-negative sensing matrix has an impact on the spectrum reconstruction. When the measurements are corrupted by a Poisson noise, as in the low-photon counting regime, the mean square error (MSE) of typical CS reconstruction methods (such as \eqref{eqn:analysisbpdnNN}) is proportional to the mean of the observed image and inversely proportional to both the compressive \emph{oversampling} ratio $M/K$ and to the light intensity, where $K$ is the expected sparsity level of the image. Moreover, there exist intrinsic upper bounds on the image reconstruction quality discovered in \cite{raginsky2010compressed} for compressive optical systems subject to Poisson noise.}

{We can make two observations relatively to those prior works. First, the limitations explained in \cite{raginsky2010compressed} do not apply readily here. Conversely to the assumption made in this work, our system is not ``flux-preserving'', \ie the reconstruction relies on an accumulation of measurements obtained by multiple SLM patterns and, as for the single-pixel CS camera~\cite{Duarte:2008}, the recorded light flux can be made arbitrarily large, independently of the incoming light flux.}
{Second, the schlieren deflectometer analyzed in this paper is not so much affected by the Poisson noise effect analyzed in \cite{Studer:2012wc} since the CCD camera records data in a high-photon counting mode. Our system is globally characterized by an important light throughput induced by
	the selection of our binary SLM patterns, \ie about 50\% of the SLM pixels are ``on'' for any rows of $\bs \Phi_{\rm opt}$ in \eqref{eqn:sensingbias}. Even with a SLM throughput close to 23\% (as governed by the LCD technology), the light intensity collected per pixel in the CCD focal plane is still important, thanks to a 12W monochromatic backlight source illumination (see Fig.~\ref{fig:pssetnimo}). Conversely to fluorescent imaging techniques, the system also observes the direct transmission of the light source, \eg with no additional losses of light conversion by the material. A conservative calculation shows that for a typical exposure time of 50\,ms per observation, considering the source wavelength (546 nm) and an angular acceptance of the telecentric system of 3.14 $10^{-4}$ steradian, more than 10$^5$ photons per CCD pixel and per observation are collected. At such a high-photon counting regime the sensing noises are therefore dominated by additive noises or by noises well approximated by an additive model (\eg due to high-resolution quantization distortion or to CCD readout noise). Consequently, the sensing corruption $\cl C$ in \eqref{eq:schlieren-forward-model} will be considered as mainly additive in our developments.} 

{\paragraph*{Noisy multiple measurement vector model} We propose a model that covers the impacts of all the additive noises by splitting their origins into two sources: a \emph{signal} noise corrupting the signal before its sensing, and a measurement noise impacting only the observation process independently of the signal. Moreover, we adopt a (compressive) Multiple Measurement Vector (MMV) model (see \eg \cite{mishali:mmv}) that illustrates the implicit (massive) parallelism of our measurement process induced by the $N_C$ pixels of the CCD camera. This model is also common in other high-dimensional compressive acquisition imaging procedures such as in compressive hyperspectral imaging \cite{gehm2007single,6595593}.}

{Mathematically, given a positive sensing matrix $\bs \Phi \in \bb R_+^{M \times N}$, if we gather the $N_C$ measurement vectors $\bs y_k$ (one per CCD pixel) in a single matrix $\bs Y = (\bs y_1, \cdots, \bs y_{N_C}) \in \bb R_+^{M \times N_C}$, the model \eqref{eq:schlieren-forward-model} becomes 
	\begin{equation}
	\label{eq:MMV-noisy-model}
	\bs Y = \bs \Phi(\bs{\sf S} + \bs{\sf N}_{\rm s}) + \bs {\sf N},
	\end{equation}
	where $\bs{\sf S} = (\bs s_1, \cdots, \bs s_{N_C}) \in \bb R_+^{N \times N_C}$ is the matrix containing the $N_C$ deflection spectra, $\bs {\sf N}_{\rm s} = (\bs n_{\rm s, 1}, \cdots, \bs n_{\rm s, N_C}) \in \bb R_+^{N \times N_C}$ is the signal noise corrupting $\bs{\sf S}$ before the sensing process and $\bs {\sf N} = (\bs n_{1}, \cdots, \bs n_{ N_C}) \in \bb R_+^{M \times N_C}$ is the measurement noise impacting only the sensing procedure independently of the signals.}

{To aid our further analysis, we assume that the noises are all independent and that all entries of $\bs{\sf N}_{\rm s}$, and those of $\bs{\sf N}$, are independently and identically distributed, with possibly different distributions between the two independent matrices. 
	This means that the distribution of those noises are independent of the programmed SLM patterns (\ie with respect to the row indices of $\bs{\sf N}$), of the CCD pixel location (for the column indices of both $\bs{\sf N}_{\rm s}$ and $\bs{\sf N}$) and of the deflection spectrum locations (for the row indices of $\bs{\sf N}_{\rm s}$). Hereafter, we denote by $\mu_{\rm s}$ and~$\varsigma_{\rm s}$ (respectively by $\mu$ and~$\varsigma$) the mean and standard deviation of any of the entry of $\bs{\sf N}_{\rm s}$ (resp. of $\bs{\sf N}$). 
	
	{In this context, the debiased measurements $\bs Z = (z_1, \cdots, z_{N_C})$ associated to \eqref{eq:debiased-sensing} are issued from the noisy model
		\begin{align}
		\bs Z&= N^{-1/2}\big(2\bs \Phi_{\rm opt}(\bs{\sf S} + \bs{\sf N}_{\rm s}) + \bs {\sf N} - \bs \phi^T_{\rm on} (\bs{\sf S} + \bs{\sf N}'_{\rm s}) \bs 1_M \bs 1^T_{N_C} - \bs 1_M (\bs n')^T \big),\nonumber\\
		&=\bs\Phi_{\rm ss}(\bs{\sf S} + \mu_{\rm s} \bs 1_N \bs 1^T_{N_C}) + \mu \bs 1_M \bs 1^T_{N_C} + \bar{\bs{\sf N}},
		\label{eq:Z-MMV-noisy-model}
		\end{align}
		where the entries of $\bs{\sf N}'_{\rm s}$ and of $\bs n' \in \bb R^{N_C}$ are distributed as those of $\bs{\sf N}_{\rm s}$ and $\bs{\sf N}$, respectively, and with $\bar{\bs{\sf N}}$ a colored noise with zero mean.
	}
	
	\begin{table}[t]
		\centering
		\begin{tabular}{|c|c|c|}
			\hline  
			&Signal Noise&Measurement Noise\\[1mm]
			\hline
			&&\\[-3mm]
			Mean&$\hat{\mu}_{\rm s}=1.1313\ 10^{-4}$&$\hat{\mu}=3.0999$\\[1mm]
			\hline
			&&\\[-3mm]
			Variance&$\hat{\varsigma}^2_{\rm s}=7.8747\ 10^{-4}$&$\hat{\varsigma}^2=1.0936$\\[1mm]
			\hline
			&&\\[-3mm]
			Dynamic&$\|\bs s^{\rm no}\|_{\infty} = 0.63$&$\|\bs\Phi_{\rm opt}\bs s^{\rm no}\|_\infty = 40.10$\\[1mm]
			\hline
		\end{tabular}
		\ \\[2mm]
		\caption{Measurement and signal noises parameters. Following the calibration methods of Secs.~\ref{subsubsec:measnoise}-~\ref{sec:uppo-spre-funct}, those have been estimated over a CCD area of 4096 pixels. As a point of comparison, the last line of this table provides the dynamic range of the spectrum $\bs s^{\rm no}$ obtained in the absence of object and of its compressive observations (see Sec.~\ref{sec:upcal-sign-noise}).}
		\label{table:noise-param}
	\end{table}
}

{We present below how the estimation of these four parameters can been obtained by benefiting of the properties of our imaging system and of the MMV model above. The result of these estimations is already summarized in Table~\ref{table:noise-param}. As an aside, we also show how to measure the PSF of the instrument, \ie its response when light deflection is measured in the absence of object. These calibrations allow us to finally estimate the impact of these noises on each measurement vector $\bs y_k$, which is of course mandatory to stabilize the reconstruction of $\bs s_k$ described in Sec.~\ref{sec:problemformulation}. }

\subsubsection{{Calibration of the measurement noise}}
\label{subsubsec:measnoise}

{The parameters of this noise are easily estimated by programming the SLM with a fully opaque pattern. In other words, by taking a single pattern $\bs\Phi_{\rm opt} = (0, \cdots, 0) \in \bb R_+^{1\times N}$ in \eqref{eq:MMV-noisy-model} (\ie with $M=1$), irrespectively of the observed spectra $\bs{\sf S}$, the MMV model \eqref{eq:MMV-noisy-model} reduces simply to
	$$
	\bs Y^0 :=  (y^0_1,\cdots, y^0_{N_C}) = \bs \Phi_{\rm opt}(\bs{\sf S} + \bs{\sf N}_{\rm s}) + \bs {\sf N} = \bs {\sf N}.
	$$ 
	with $\bs {\sf N} = (n_1,\,\cdots, n_{N_C})$ since $M=1$.} 

{Assuming as above that the sensing noise is independent of the SLM pattern, these observations correspond then to $N_C$ samples of the distribution characterizing the entries of $\bs {\sf N}$. The appropriate (unbiased) estimators of $\mu$ and $\varsigma$ corresponds then simply to 
	$$
	\hat{\mu} = \bb E_{\rm exp} \bs Y^0 := \tfrac{1}{N_C} \sum_j y^0_j\quad \text{and}\quad \hat{\varsigma}^2 = {\rm Var}_{\rm exp} \bs Y^0 := \tfrac{1}{N_C-1} \sum_j(y^0_j - \hat{\mu})^2,
	$$
	where $\bb E_{\rm exp}$ and ${\rm Var}_{\rm exp}$ are the experimental mean and variance, respectively, computed from the samples collected on the CCD pixels.
}

\subsubsection{{Calibration of the signal noise variance}}
\label{sec:upcal-sign-noise}

{For estimating the variance of the signal noise, we must exploit some specific properties of our schlieren deflectometer. We notice first that in the absence of any test object the incident light does not undergo any deflections and hence the deflection spectrum should ideally consist of a single peak at its origin, independently of the CCD locations~$k$. Such an ideal spectrum $\bs s^{\rm no}$ is of course not observed in reality. The schlieren deflectometer displays a non-trivial \emph{point spread function} (PSF) induced by the finite size of the TS pinhole in the schematic~\cref{subfig:pss}. Incidentally, we provide in Sec.~\ref{sec:uppo-spre-funct} a method for estimating~$\bs s^{\rm no}$.} 

{The constancy of $\bs s^{\rm no}$ with respect to $\k$, \ie the fact that all spectra in \eqref{eq:MMV-noisy-model} are given by $\bs{\sf S}^{\rm no} = \bs s^{\rm no}\bs 1^T_{N_C}$, allows us to estimate the variance $\varsigma^2_{\rm s}$. This is achieved by proceeding as in the previous estimations of $\mu$ and $\varsigma$, \ie we set \eqref{eq:MMV-noisy-model} in a single measurement process with $\bs \Phi_{\rm op} = \bs 1_N^T$ and $\bs {\sf N} = (n_1,\cdots, n_{N_C})$ (\ie $M=1$), so that 
	$$
	\bs Y^{1,\rm no} :=  (y^{1,\rm no}_1,\cdots, y^{1,\rm no}_{N_C}) = \bs 1_N^T (\bs s^{\rm no}\bs 1^T_{N_C} + \bs{\sf N}_{\rm s}) + \bs {\sf N}.
	$$ 
	Since the first term is constant, we find 
	$$
	{\rm Var}_{\rm exp} \bs Y^{1,\rm no} = {\rm Var}_{\rm exp}(\bs 1_N^T \bs{\sf N}_{\rm s} + \bs {\sf N}),
	$$
	with ${\rm Var}_{\rm exp} \bs Y^{1,\rm no}$ a good estimator of ${\rm Var}\,y_\k^{1,\rm no}$ on each pixel index $\k$ when $N_C$ is large.
	However, from the independence of $\bs{\sf N}_{\rm s}$ and $\bs {\sf N}$, we have also on each pixel index~$\k$
	$$
	{\rm Var}\,y_\k^{1,\rm no} = {\rm Var}( \bs 1_N^T \bs{n}_{\rm s,k}) + \varsigma^2 = N\varsigma^2_{\rm s} + \varsigma^2.
	$$
	Consequently, having established in the previous section an estimator $\hat{\varsigma}^2$ on $\varsigma^2$, we obtain another one on $\hat\varsigma_{\rm s}$ such that
	$$
	N\hat\varsigma^2_{\rm s} := {\rm Var}_{\rm exp} \bs Y^{1,\rm no} - \hat\varsigma^2.
	$$
}

\if 0
{\emph{Remark:} In cases where we do observe some test objects, the above computation still hold if the underlying spectrum $\vecs_{k}$ maintains its overall shape across $k$ but only gets translated depending upon $k$, thereby rendering $\bs 1_N^T\vecs_{k}$ a constant with respect to $\k$. This assumption is true for smooth objects such as plano-convex lenses on large regions of the CDD, and on restricted CCD areas for complicated objects such as MIOLs.}
\fi

%%%%%%%%%%%%%%%%%%%%%%%%%%%%%%%%%%%%%

\subsubsection{{Signal noise mean and point spread function estimations}}
\label{sec:uppo-spre-funct}

\begin{figure}[!tb]
	\centering
	\subfigure[\label{fig:s-no-singl-meas}]{
		\includegraphics[width=.3\textwidth]{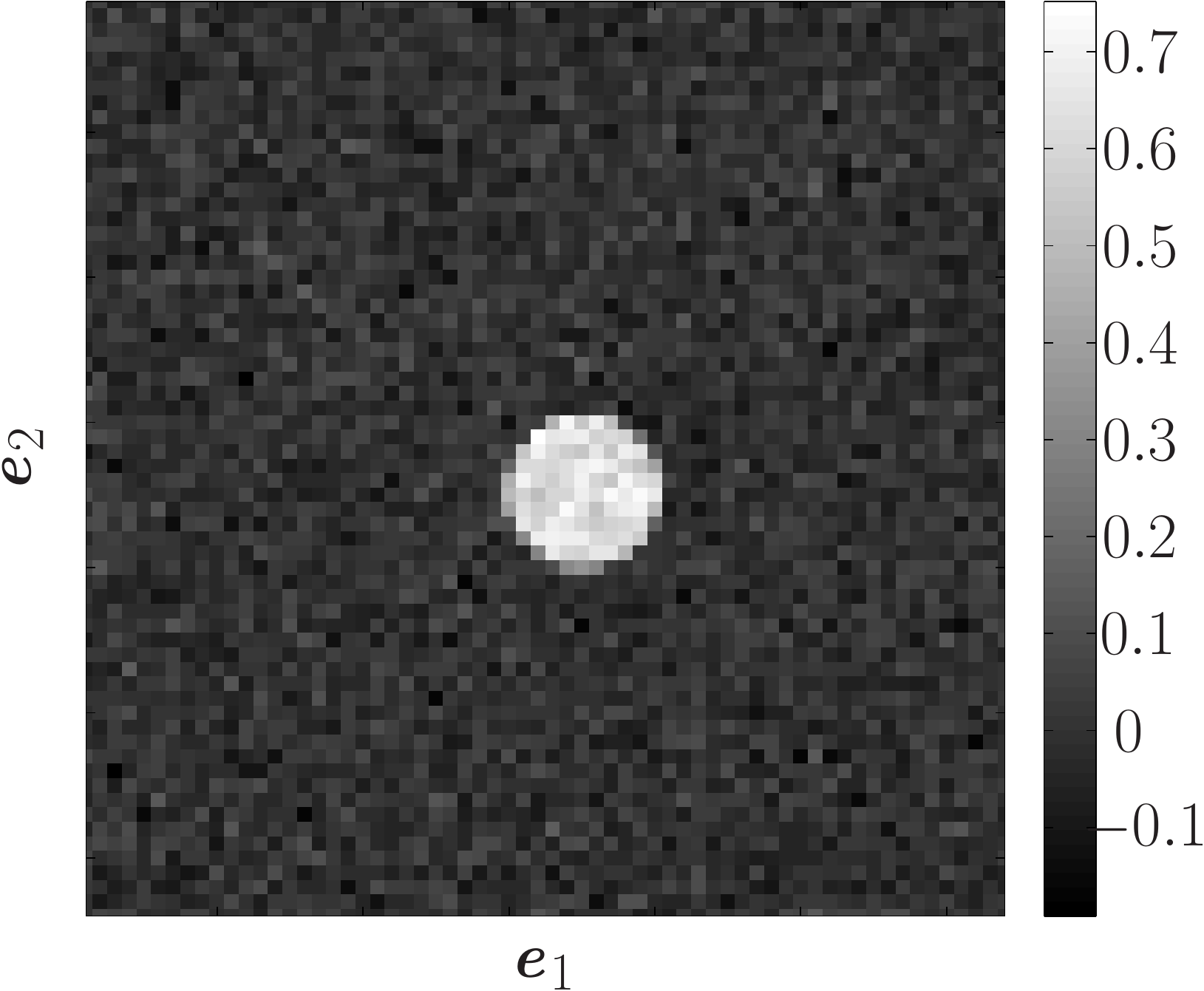}
	}
	\subfigure[\label{fig:psf-estim}]{
		\includegraphics[width=.3\textwidth]{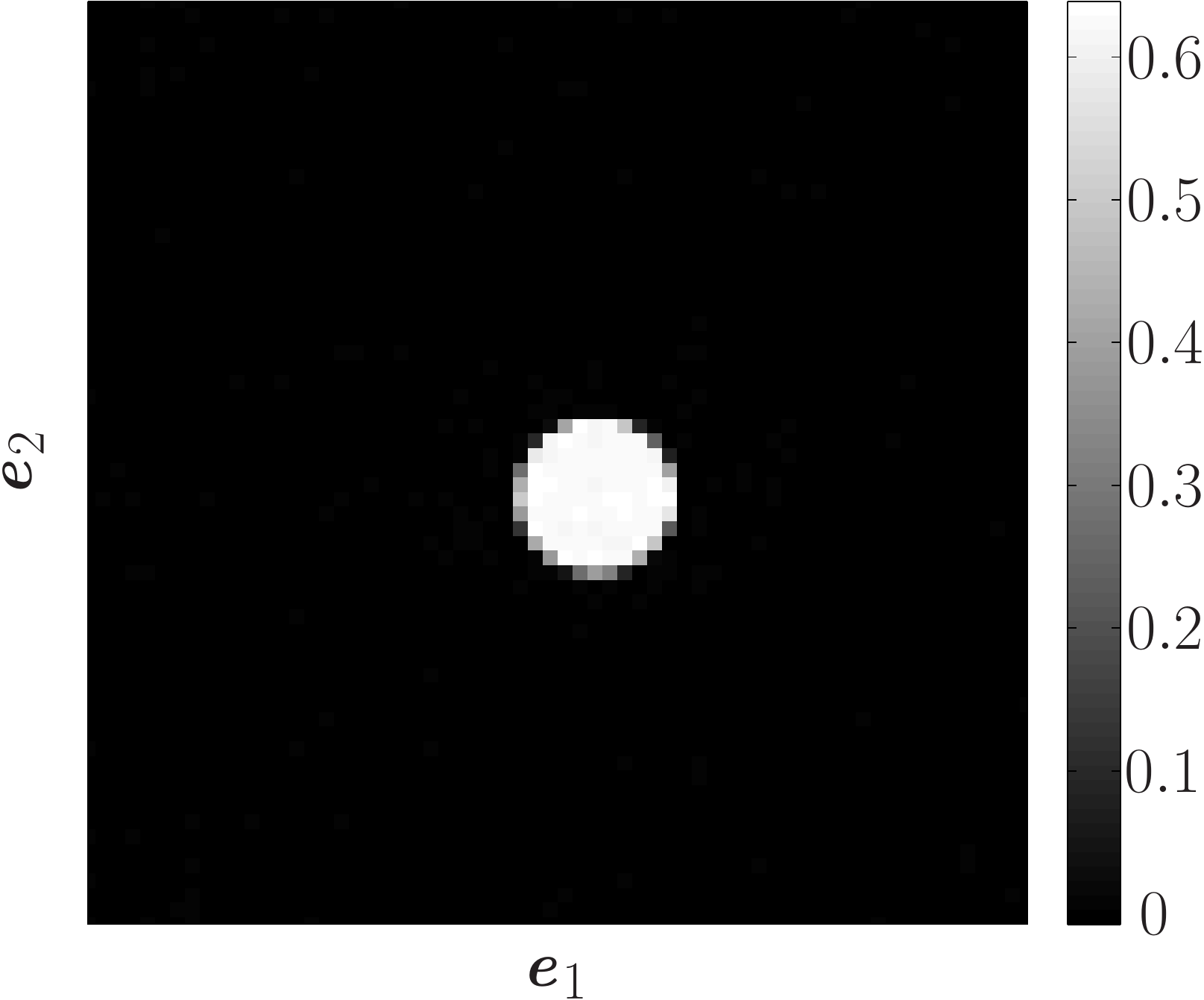}
	}
	\caption{{PSF estimation obtained from \eqref{eq:psf-estim} with $M=N$ measurements: (a) on a single CCD pixel $\k$ (setting $N_C=1$) and (b) with $N_C=4096$.}}
	\label{fig:psf-estimation}
\end{figure}

{This section shows that estimating the signal noise mean passes through the calibration of the instrument point spread function (PSF), \ie the spectrum observed in the absence of object. This can be done by configuring the sensing matrix as in \eqref{eqn:sensingbias} with\footnote{In order to limit the number of measurements, the SLM can be programmed on a limited area around the expected spectrum location, \ie inserting the rows of $\bs \Phi_{\rm opt}$ only in a limited zone and filling the rest of the SLM with opaque pixels. For the ``no object'' spectrum, we have set $N=M=64^2=4096$.} $M=N$ and $\sigma_1=1$. Since in the absence of any object the spectra are independent of the location $\k$, $\bs{\sf S}^{\rm no} = \bs s^{\rm no} \bs 1_{N_C}^T$ and the sensing model reads
	$$
	\bs Y^{\rm no} = \bs \Phi_{\rm opt}(\bs s^{\rm no} \bs 1_{N_C} + \bs{\sf N}_{\rm s}) + \bs{\sf N}.
	$$ 
	Projecting the left and the right hand sides of this equation on the vector $N_C^{-1}\bs 1_{N_C}$ and applying the inverse sensing matrix on the result, we get a biased estimation of $\bs s^{\rm no}$
	$$
	\tilde{\bs s}^{\rm no} := N_C^{-1} \bs \Phi^{-1}_{\rm opt}\bs Y^{\rm no} \bs 1_{N_C} = \bs s^{\rm no}  + \bar{\bs n}_{\rm s} + \bs \Phi^{-1}_{\rm opt} \bar{\bs n},
	$$
	with $\bar{\bs n}_{\rm s} = N_C^{-1} \bs{\sf N}_{\rm s} \bs 1_{N_C}$ and $\bar{\bs n} = N_C^{-1} \bs{\sf N}\bs 1_{N_C}$. 
	
	It is easy to see that, up to a constant bias that will be subtracted later, $\tilde{\bs s}_{\rm no}$ provides a good estimate of $\bs s_{\rm no}$ when $N_C$ is large. First, each component of $\bar{\bs n}_{\rm s}$ has a mean $\mu_{\rm s}$ and a variance $\varsigma^2_{\rm s}/N_C$. Concerning $\bs \Phi^{-1}_{\rm opt} \bar{\bs n}$ we find $\bb E \bs \Phi^{-1}_{\rm opt} \bar{\bs n} = \mu \bs \Phi^{-1}_{\rm opt} \bs 1_{N}$. From \eqref{eq:inverse-phi} and $\bs H \bs 1_N = \sqrt N \bs c_1$, we get $\bs \Phi_{\rm opt}^{-1} \bs 1_N = \tfrac{1}{\sqrt N}(2 \bs I_{N} - \bs c_1\bs 1_N^T)\bs \Sigma \bs H \bs 1_N = \bs c_1$, so that $\bb E \bs \Phi^{-1}_{\rm opt} \bar{\bs n} = \mu \bs c_1$.}

{Second, we can show that the covariance of $\bs\Phi^{-1}_{\rm opt} \bar{\bs n}$ vanishes as $O(1/N_C)$. Indeed,
	\begin{align*}
	\bb E (\bs \Phi^{-1}_{\rm opt} \bar{\bs n} - \mu \bs c_1) (\bs \Phi^{-1}_{\rm opt} \bar{\bs n} - \mu \bs c_1)^T&=\bs \Phi^{-1}_{\rm opt} \bb E\big( (\bar{\bs n} - \mu \bs 1_N) (\bar{\bs n} - \mu \bs 1_N)^T\big) (\bs \Phi_{\rm opt}^{-1})^T\\
	&=\tfrac{1}{N_C}\varsigma^2 \bs \Phi^{-1}_{\rm opt} (\bs \Phi_{\rm opt}^{-1})^T\\  
	% &=\tfrac{1}{N_C}\varsigma^2 \tfrac{1}{\sqrt N}(2 \bs I_{N} - \bs c_1\bs 1_N^T)\bs \Sigma \bs H  
	% \bs H^T \bs \Sigma \tfrac{1}{\sqrt N}(2 \bs I_{N} - \bs 1_N \bs c_1^T)\\
	&=\tfrac{1}{N_C}\varsigma^2 \tfrac{1}{N}(2 \bs I_{N} - \bs c_1\bs 1_N^T)(2 \bs I_{N} - \bs 1_N \bs c_1^T),  
	\end{align*}
	but for any $\bs u \in \bb R^N$, a crude bound gives $\bs u^T (2 \bs I_{N} - \bs c_1\bs 1_N^T)(2 \bs I_{N} - \bs 1_N \bs c_1^T) \bs u \leq (N+8) \|\bs u\|^2$, so that ${\rm Cov}(\bs \Phi^{-1}_{\rm opt} \bar{\bs n}) \preccurlyeq \tfrac{1}{N_C}\varsigma^2 (1+ \tfrac{8}{N}) \bs I_{N} = O(\bs I_{N} / N_C)$.
	
	Consequently, for large $N_C$, we find  
	$$
	\tilde{\bs s}^{\rm no} \approx \bs s^{\rm no}  + \mu_{\rm s} \bs 1_N + \mu \bs c_1.
	$$
	Moreover, since $\bs c_1$ is non-zero only on the first pixel of the spectrum domain, assuming reasonably that the support of $\bs s^{\rm no}$ does not contain this first pixel (\ie $\bs c_1^T \bs s^{\rm no} = 0$) and is small compared to its complementary set, we can define the two following estimators:
	\begin{align}
	\label{eq:signal-noise-mean}
	\hat{\mu} _{\rm s} &= \rm median \,(\tilde{\bs s}^{\rm no})\\
	\label{eq:psf-estim}
	\hat{\bs s}^{\rm no}&= H_1(\tilde{\bs s}^{\rm no} - \hat{\mu}_{\rm s} \bs 1_N),
	\end{align}
	where $H_1(\bs u)$ sets to zero the first value of $\bs u \in \bb R^N$. Anticipating slightly over the experimental setup described in Sec.~\ref{subsec:deflectoExperiments}, Fig.~\ref{fig:psf-estimation} compares the estimation of $\bs s^{\rm no}$  obtained by simply inverting $\bs\Phi_{\rm opt}$ (with $M=N$) on $\bs y_\k$ for a arbitrary pixel $k$ with the one associated to \eqref{eq:psf-estim}.  We clearly observe that the level of noise vanishes from the averaging procedure given in \eqref{eq:psf-estim}. 
}

\subsubsection{{Noise impact on random measurements}}
\label{sec:upno-impact-rand}

{
	Having established above how to estimate the main parameters of noises $\bs{\sf N}_{\rm s}$ and $\bs{\sf N}$, it is now easy to bound the noise power on the observations of deflection spectra. 
	
	Focusing our analysis on a single pixel~$\k$ and on the spread-spectrum CS matrix $\bs\Phi_{\rm opt}$ given is Eq.~\eqref{eqn:sensingbias}, the model Eq.~\eqref{eq:MMV-noisy-model} becomes
	%$$
	%\bs z_{\k} = 2\bs\Phi_{\rm opt} \bs s_\k + 2\bs\Phi_{\rm opt} \bs{n}_{\rm s} + 2\bs n_k - \bs 1_\M(\bs 1^T_\N\bs n_s') - \bs 1_\M\bs n'.
	%$$
	
	$$
	\bs z_\k = 2\vecy(\vecs_\k, \Phi_{\rm opt}) - y(\vecs_\k, \bs A^1)\bs 1_\M,
	$$
	where $\vecy(\vecs_\k, \Phi_{\rm opt})$ is a vector of noisy measurements of $\vecs_\k$ of size $\M$ and $y(\vecs_\k, \bs A^1)$ is the single measurement of $\vecs_\k$ using an all ones pattern $\bs A^1 = \bs 1_\N$, with a different noise realization.
	
	After expanding the two terms, we have 
	$$
	\bs z_\k = \sqrt{\N}\Phi_{\rm ss}\vecs_\k +2\Phi_{\rm opt}\vecn_s + 2\vecn - \bs 1_\M (\bs 1_\N^T\vecn_s') -  n'\bs 1_\M.
	$$
	The noise terms in the above expression can be further split into zero mean terms and systematic bias terms. Recalling that the signal noise and measurement noise have means $\mu_s$ and $\mu$, we obtain
	$$
	\bs z_\k = \sqrt{\N}\Phi_{\rm ss}\vecs_\k + \underbrace{\sqrt{\N}\Phi_{\rm ss}\mu_s\bs 1_\N + \mu\bs 1_\M}_{\text{systematic bias}} + \underbrace{2\Phi_{\rm opt}\tilde{\vecn}_s + 2\tilde{\vecn} - \bs 1_\M (\bs 1^T_\N\tilde{\vecn}'_s) - \tilde{n}'_1\bs 1_\M}_{\text{zero mean noise}}, 
	$$ 
	where $\tilde{\vecn}_s, \tilde{\vecn}'_s, \tilde{\vecn}$ and $\tilde{n}'_1$ are zero mean noises.
	
	First, the systematic bias 
	\begin{equation}
	\bs b:= \sqrt{\N}\Phi_{\rm ss}\mu_s\bs 1_\N + \mu\bs 1_\M
	\label{eq:noise-bias-estim}
	\end{equation}
	must be subtracted from the observations $\bs z_\k$.

	Second, bounding the observation noise power amounts to finding an appropriate bound $\epsilon$ such that
	$$
	\|\bs z_\k - \bs \Phi_{\rm ss} \bs s_\k - \bs b\|^2 \leq \epsilon^2
	$$
	holds with high probability with respect to the noise randomness. This is mandatory for adjusting the fidelity constraints of any reconstruction method (such as the variation of BPDN \eqref{eqn:bpdn} introduced in the next section). With $\bs \xi:= 2\Phi_{\rm opt}\tilde{\vecn}_s + 2\tilde{\vecn} - \bs 1_\M (\bs 1^T_\N\tilde{\vecn}'_s) - \tilde{n}'_1\bs 1_\M$, we have
	$$
	\bs z_\k = \sqrt{\N}\Phi_{\rm ss}\vecs_\k + \bs b + \bs \xi.
	$$  
	Replacing $\bs b$ by $\hat{\bs b} := \sqrt{\N}\Phi_{\rm ss}\hat{\mu}_s\bs 1_\N + \hat\mu\bs 1_\M$, we obtain $\bs \xi \approx\bs z_k - \sqrt{\N}\Phi_{\rm ss}\vecs_\k - \hat{\bs b}.$
	
	Estimating the noise bound $\epsilon^2$ amounts to estimating $\ltwoof{\xi}^2$, which is hard in general but thanks to the measurements of the no object spectrum we can rely on the estimate $\hat{\vecs}^{\rm no}$ in Eq.~\eqref{eq:psf-estim} and the MMV model in Eq.~\eqref{eq:MMV-noisy-model} of the deflectometric system. For each $\M$, we consider the no object measurement vectors $\vecz_\k = 2\vecy(\vecs^{\rm no}, \Phi_{\rm opt}) - y(\vecs^{\rm no}, \bs A^1)\bs 1_\M$ at all CCD locations and construct the empirical distribution (across CCD locations) of $\ltwoof{\vecz_\k - \sqrt{\N}\Phi_{\rm ss}\hat{\vecs}^{\rm no} - \bs b}^2$. We then simply set $\epsilon^2$ to be the $75^{\rm th}$ percentile of this empirical distribution.

\section{{Deflection spectrum recovery and numerical method}}
\label{sec:problemformulation}

{While solving the optimization problem in Eq.~\eqref{eqn:bpdn} the choice of the sparsity basis $\spasis$ plays an important role in terms of the quality of the solution. Even though it is remarked earlier that the spectra we wish to reconstruct are sparse in the canonical domain itself, due to the nature of the objects we investigate, it is advantageous nevertheless to work in some other domain. One choice is to use wavelets as they are suitable for signals which are piecewise regular. 
	
	For sparse representation of deflection spectra, we use translation invariant frame of wavelet associated with the UnDecimated Wavelet Transform (UDWT), \ie the usual wavelet transform without the decimation~\cite{Mallat:2008:WTS:1525499}. The non-uniqueness of the decomposition of a signal in such a redundant frame allows us to represent signals in their sparsest possible way~\cite{Elad:2010:SRR:1895005, Candes:2010p581}, which may not be possible with a wavelet basis. 
	Moreover, for (locally) smooth objects, the deflection spectra associated to neighboring CCD pixels are mainly differentiated by a small translation while their shapes are almost identical. This justifies the use of a translation invariant sparsity prior in order to ensure a reconstruction error for the different deflection spectra that is spatially constant. In our work, we use a UDWT built on the Daubechies 16 tap filter in all our reconstruction experiments. 
	
	Notice that, rather than estimating the deflection spectra with the sparse synthesis prior used in Eq.~\eqref{eqn:bpdn} \cite{Sudhakar:2013uk,Sudhakar:2012uk}, we have preferred an analysis based prior model \cite{Elad:2010:SRR:1895005, Puy:2012p2362} for regularizing our inverse problem. Consequently, as our deflection spectra are positive by nature, the optimization problem we solve for estimating each of them in each pixel $\k$ reads
	\begin{equation}
	\hat{\vecs}_k:=\argmin_{\vecs}\loneof{\spasis^\ast\vecs}\;\text{subject to}\;\ltwoof{\vecy_k - \slmpat_{\rm opt}\vecs - \bs b}\leq\noistd\;\text{and}\;\vecs \in\realno_+^\N,
	\label{eqn:analysisbpdnNN}
	\end{equation}
	where $\bs b$ is the noise bias estimated in {Eq.~\eqref{eq:noise-bias-estim}} and $\epsilon$ is the noise power estimated in Sec.~\ref{sec:upno-impact-rand}.
	
	% Eq.~\eqref{eq:noise-power-estim}, respectively. 
	
	In the absence of the positivity constraint, if the analysis dictionary $\spasis$ is a tight frame and $\sensis$ satisfies Dictionary-Restricted Isometry Property (D-RIP) with the isometry constant $\delta_{2\K}<0.08$, then the solution $\hat{\vecs}$ satisfies~\cite[Thm. 1.4]{Candes:2010p581} 
	\begin{equation}
	\ltwoof{\vecs _k - \widehat{\vecs}_k} = O\left(\tfrac{\loneof{\spasis^\ast\vecs_k - (\spasis^\ast \vecs _k)_\K}}{\sqrt{\K}}\;+\;\noistd\right).
	\label{eqn:analysisbpdnreconerr}
	\end{equation}
	Recovery guarantees when $\spasis$ is not a tight frame and measurements are Gaussian can be found in~\cite{Kabanava:2013}. In the reported work, $\spasis$ is always a UDWT based tight frame. 
	
}
\subsection*{Convex optimization algorithm}
\label{sec:numericalmethod}
To solve the program~\eqref{eqn:analysisbpdnNN}, we make use of a primal-dual method called the Chambolle-Pock (CP) algorithm~\cite{Chambolle:2011p1809}. Chambolle-Pock algorithm solves primal-dual forms of unconstrained convex problems and it relies on \emph{proximal operators} of the functions involved in the objective. It has a flexible structure, which allows easy inclusion of additional terms in the objective function. Furthermore, it has guarantees of convergence under under broad conditions on the objective function. 

{Chambolle-Pock algorithm is used to solve primal-dual formulations of the primal problems of type}
\begin{equation}
\min_{\vecu\in\realno^\N} \F(\matk\vecu)+\G(\vecu),
\label{eqn:primal}
\end{equation}
{where $\matk:\realno^\N\rightarrow\realno^\M$ be a continuous linear operator  with a bounded norm. Let $\F:\realno^\M\rightarrow[0, +\infty]$ and $\G:\realno^\N\rightarrow[0, +\infty]$ be two proper, convex, lower-semicontinuous functions~{\cite{bauschke:book}}.}

{The algorithm primarily relies on the \emph{proximal} operators of the two functions in{ Eq.~\eqref{eqn:primal}~\cite{Parikh:2013vb}}. Proximal operators generalize simple gradient descent when the functions are not differentiable.  Moreover, proximal operators of several functions commonly used in signal processing have closed forms which are easy to evaluate{~\cite{Combettes:2011wd, Parikh:2013vb}}. More Details on the algorithm and its convergence can be found in~{\cite{Chambolle:2011p1809}}.} 

{The constrained problem Eq.~\eqref{eqn:analysisbpdnNN} is converted into the unconstrained form Eq.~\eqref{eqn:primal} with the help of convex indicator function of the convex sets associated with each constraint.} The convex indicator function $\imath_{\mathcal{C}}$ of a convex set $\mathcal{C}$ (see Introduction) is proper, convex and lower-semicontinuous function and hence it satisfies the requirements of the Chambolle-Pock algorithm~\cite{Boyd:2004uz}. The $\ell_2$ ball of radius $\epsilon$, centred on the measurement vector $\vecy_k$, $\mathcal{B}_k  = \left\{\vecz\in\realno^\N\;|\;\|(\vecy_k - \bs b) - \vecz\|_2\leq\epsilon\right\}$, is a convex set. Hence the constraint $\ltwoof{\vecy_k - \slmpat_{\rm opt} \vecs - \bs b}\leq\epsilon$ can be inserted into unconstrained problem by including the convex indicator function of the set $\mathcal{B}_k$ into the objective.

The non-negative constraint in~\eqref{eqn:analysisbpdnNN} can be included through the convex indicator function of the non-negative orthant $\realno_+^\N$. With these considerations, the unconstrained formulation of the reconstruction problem~\eqref{eqn:analysisbpdnNN} takes the form:
\begin{align*}
\widehat{\vecs}_k&:= \argmin_{\vecs\in\realno^\N}\big( \loneof{\spasis^\ast\vecs}  + \imath_{\mathcal{B}_k}({\slmpat_{\rm opt}}\vecs) + \imath_{\realno^\N_+}(\vecs)\big).
\end{align*}

{The proximal operator of $\ell_1$ norm is the soft-thresholding operator $\prox_{\gamma\|\cdot\|_1}(\bs u) = (|\bs u| - \gamma)_+\,{\rm sign}(\bs u)$, and the one of the indicator functions of $\mathcal{B}_k$ and $\realno^\N_+$ are the projections $\prox_{\imath_{\mathcal{B}_k}}(\vecu) := \vecy + (\vecu - \vecy)\min(1,
	\epsilon\|\vecu - \vecy\|^{-1})$ and $\prox_{\imath_{\realno^\N_+}}(\vecu) := (\u)_+$ respectively.} {The adaptation of CP algorithm to handle more than two functions (as against two functions in its standard formulation), based on product space expansion, is described, \eg in~{\cite{AGonzalez:2012}}}.

\section{Experiments on Deflection Spectra Reconstructions}
\label{sec:recoveryResults}
{The spread spectrum Hadamard sensing approach to measure and reconstruct deflection spectra was first evaluated with synthetic experiments in an ideal noiseless condition. Subsequently, it was implemented and evaluated on an actual deflectometric system.

	For all the experiments reported in this article (synthetic as well as actual deflectometric), the spread-spectrum vector $\vecm$ was generated once and fixed\footnote{Fixing $\sigma_1$ to $1$ for allowing a full inversion of $\bs\Phi_{\rm opt}$ with Eq.~\eqref{eq:inverse-phi} when $M=N$ (see Sec.~\ref{sec:opticalSensing}).}, thereby fixing the full sized spread-spectrum Hadamard measurement matrix $\bs H^T\bs \Sigma$ in Eq. \eqref{eqn:sensingbias} throughout. For a given set $\senset\subseteq[\N]$ of $\M$ random indices, a typical vector of compressive measurements was generated using the sensing matrix ${\slmpat_{\rm opt}}$, defined in Eq.~\eqref{eqn:sensingbias}.} 

\subsection{Spectrum reconstruction from synthetic measurements}
\label{subset:syntheticexperiments}

{In~\cref{sec:intro}, we discussed that light deflections induced by smooth surfaces tend to be strong around a mean deflection angle and they gradually decay away from the mean angle. Moreover, when the object surface is not too curved the deflection spectral spot is also symmetric, while for more complicated surfaces we expect the spectrum to be decomposed over several spectral spots resulting both from the instrument resolution limit and PSF.We describe the resolution limit aspect in~\cref{sec:resolution}. Therefore, our compressive schlieren imaging scheme has been first verified with synthetic spectra composed of $Q$ spectral spots with $1\leq Q \leq 5$. These were synthesized by simply adding $Q$ 2-D Gaussian functions uniformly and independently placed at random over a 2-D spectrum domain of size $64\times 64$ ($N=4096$). The height and the width of each Gaussian pattern were set to 1 and to a standard deviation of 3 pixels, respectively, the width being fixed by the typical width of the instrumental PSF observed in Sec.~\ref{sec:uppo-spre-funct}. For reasons that will become clearer later, this is size is also close to those of actual deflection spectra.} 

Those spectra have been numerically observed with the sensing matrix $\bs \Phi_{\rm opt}$ in \eqref{eqn:sensingbias} under the conditions described at the beginning of this section. We select a noiseless sensing model as we focus here on the reconstruction capability of the compressive scheme as a function of the number of measurements $M$ and of the synthetic spectra complexity. The spectra were reconstructed by the method described in Sec.~\ref{sec:problemformulation}. The iterations were chosen to be stopped when the relative error of successive iterates dropped below a threshold $10^{-4}$. 

In case of synthetic experiments, as the ground truth $\vecs$ was available to us, the reconstruction performance for a given number of measurements $\M$ was evaluated by the \emph{output} SNR defined by $20\,\log_{10} (\ltwoof{\vecs}/\ltwoof{\vecs - \hat{\vecs}})$, where $\hat{\vecs}$ is the reconstructed spectrum obtained by solving \eqref{eqn:analysisbpdnNN} and $\vecs$ is the true spectrum. 

\begin{figure}[t]
	\centering
	\subfigure[]{
		\includegraphics[width=0.2\textwidth]{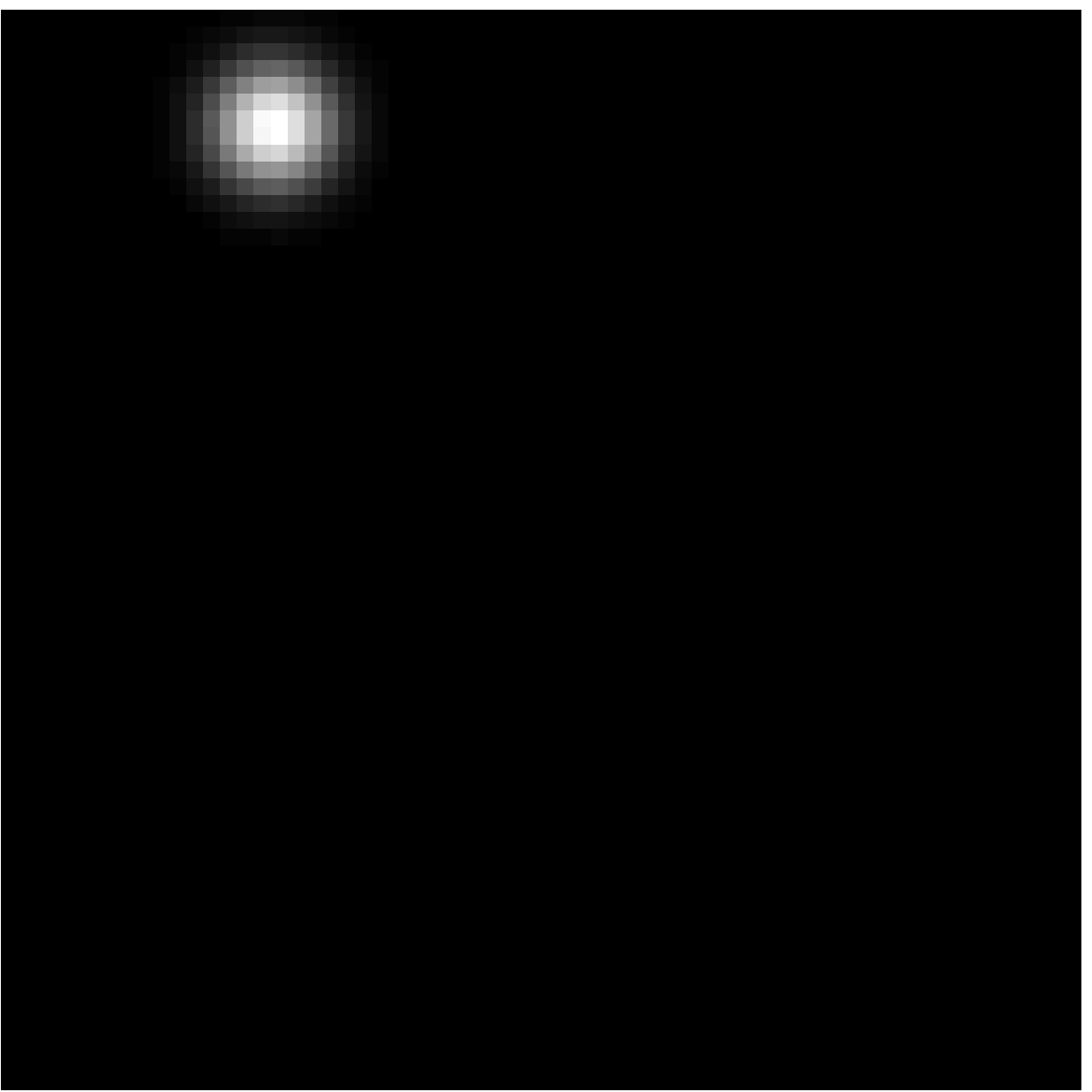}
		\label{subfig:onespot}
	}
	\subfigure[]{
		\includegraphics[width=0.2\textwidth]{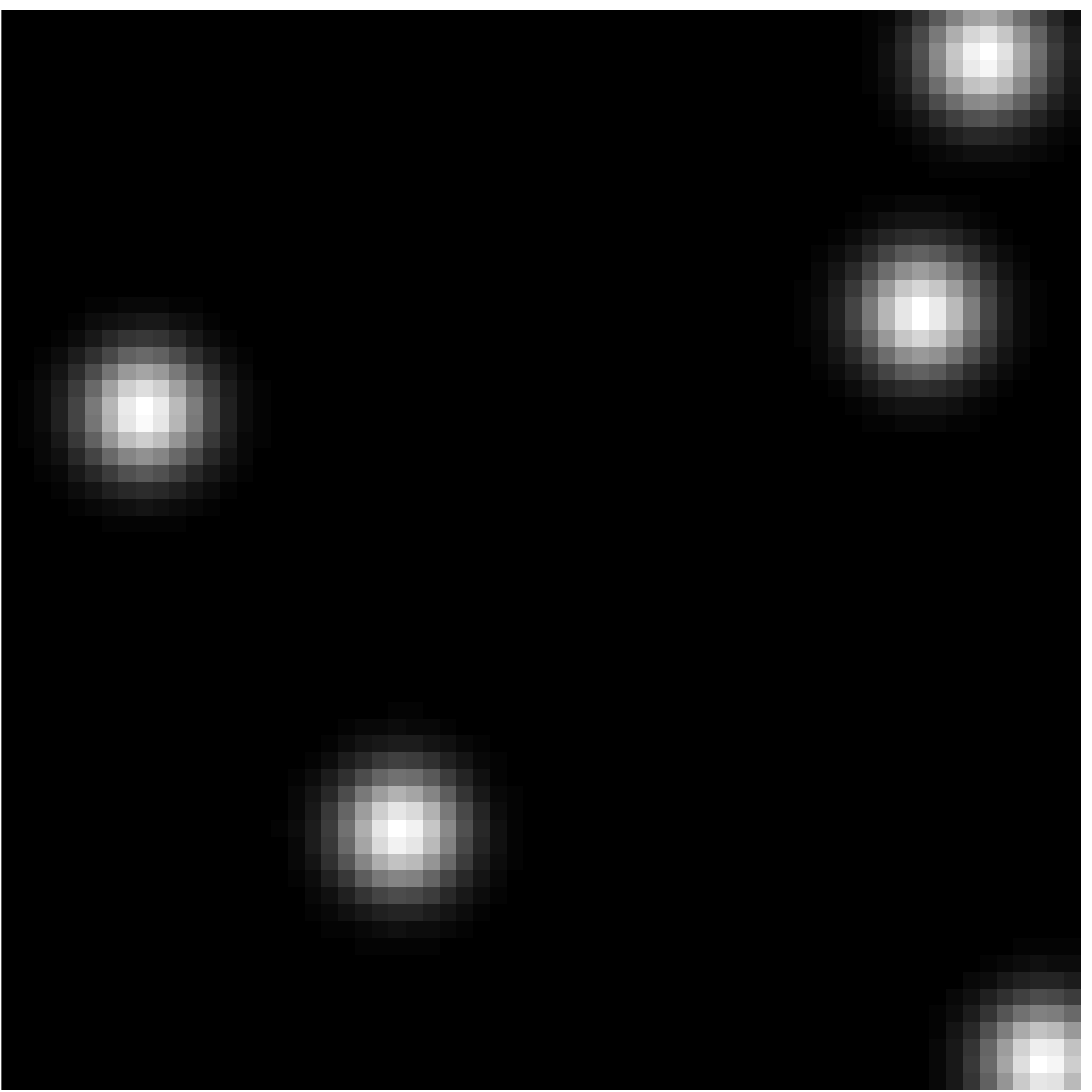}
		\label{subfig:twospot}
	}
	\caption{Examples of synthetic deflection spectrum: \subref{subfig:onespot} one spectral spot and \subref{subfig:twospot} {five} spectral spots }
	\label{fig:syntheticspectra}
\end{figure}

\begin{figure}
	\centering
	\subfigure[]{
		\centering
		\includegraphics[width=0.2\textwidth]{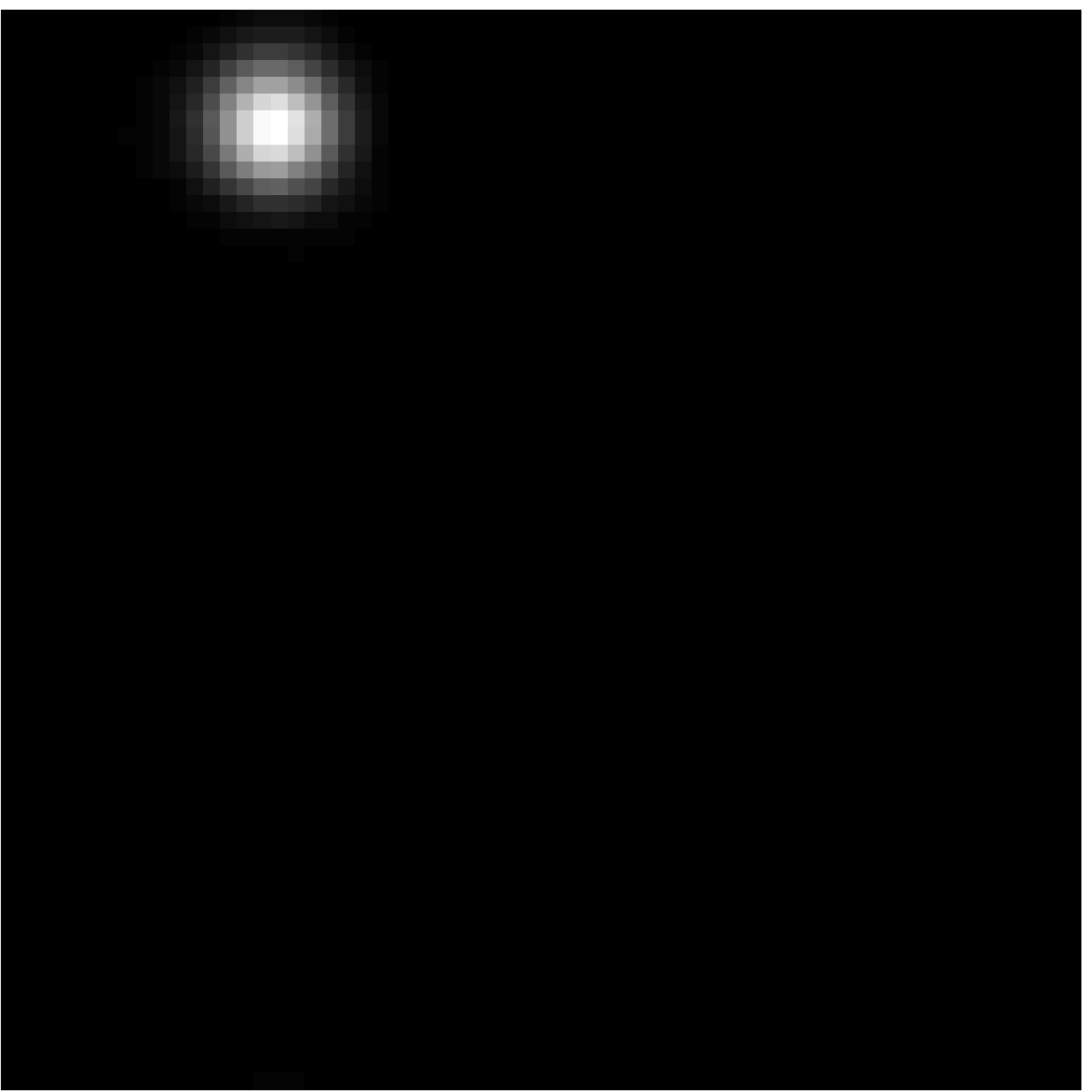}
		\label{subfig:M150Syn1AnaUDWT}
	}
	\subfigure[]{
		\centering
		\includegraphics[width=0.2\textwidth]{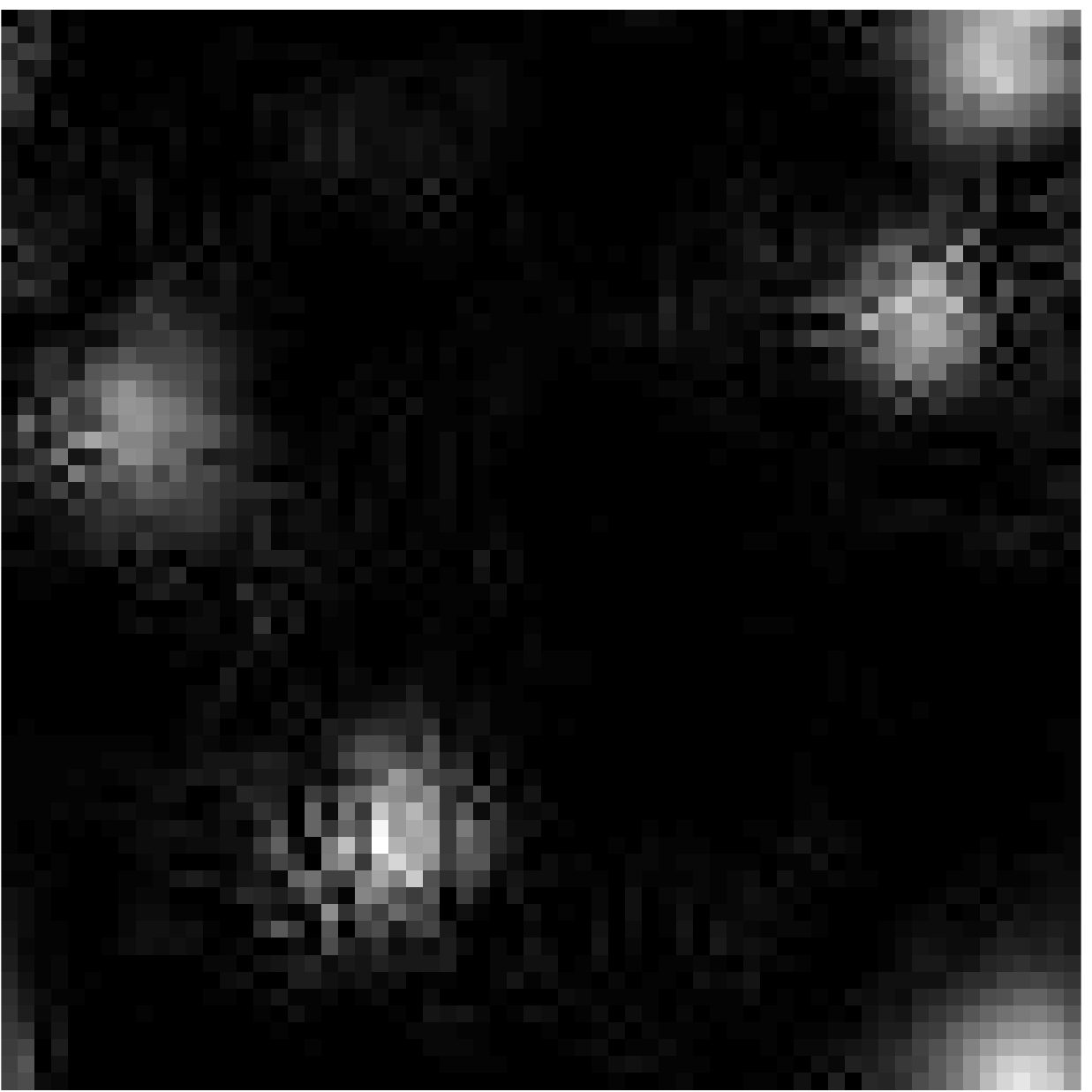}
		\label{subfig:M150Syn5AnaUDWT}
	}
	\caption{Reconstruction examples of synthetic spectra with $3.6\%$ of measurements:~\subref{subfig:M150Syn1AnaUDWT} corresponds to one spectral spot and~\subref{subfig:M150Syn5AnaUDWT} corresponds to five spectral spots.}
	\label{fig:synReconExamples}
\end{figure}

{The input spectrum for a given number of spots was fixed in all the experiments. For each value of $\M/\N$, 20 reconstruction trials were performed, each of them consisting of a new vector of measurements obtained using a new measurement matrix ${\bs \Phi}_{\rm opt}$ drawn according to \eqref{eqn:sensingbias}. The output SNR was averaged across the 20 trials respectively.}

\cref{fig:syntheticspectra} shows a typical synthetic deflection spectrum with~\subref{subfig:onespot} one and~\subref{subfig:twospot} {five} deflection spots, without any noise. \cref{fig:synReconExamples} contains the results of reconstruction using $\M/\N=3.6\%$ of the measurements. \cref{subfig:oSNRVsMeas_spots} shows the SNR curve for the three reconstruction modes on synthetically generated data as a function of the ratio $\M/\N$ in percentage, for deflection spectra containing one and {five} spots (as depicted in~\cref{fig:syntheticspectra}). 

\begin{figure}
	\centering
	\subfigure[]{
		\centering
		\includegraphics[width=0.47\textwidth]{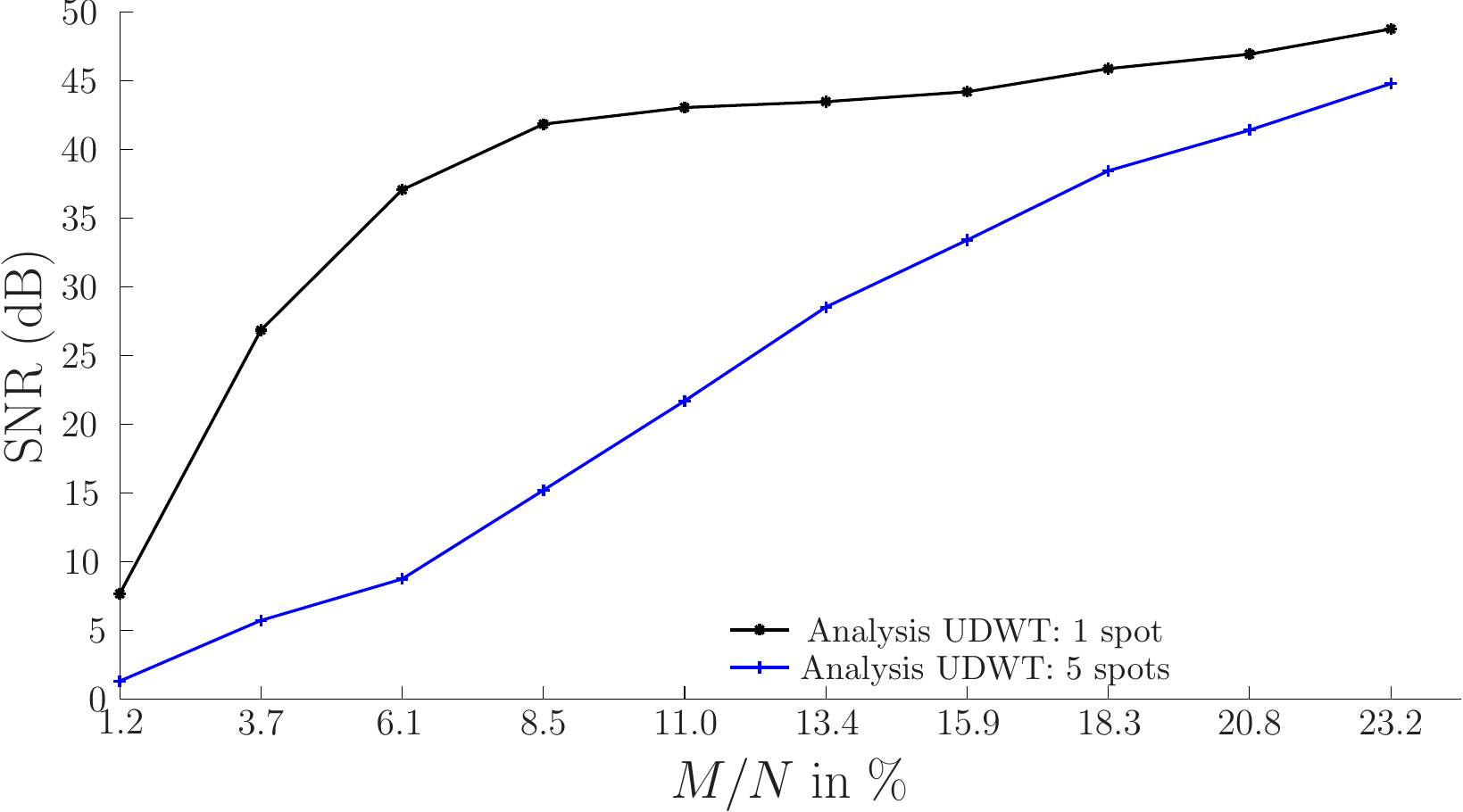}
		\label{subfig:oSNRVsMeas_spots}
	}
	\subfigure[]{
		\centering
		\includegraphics[width=0.47\textwidth]{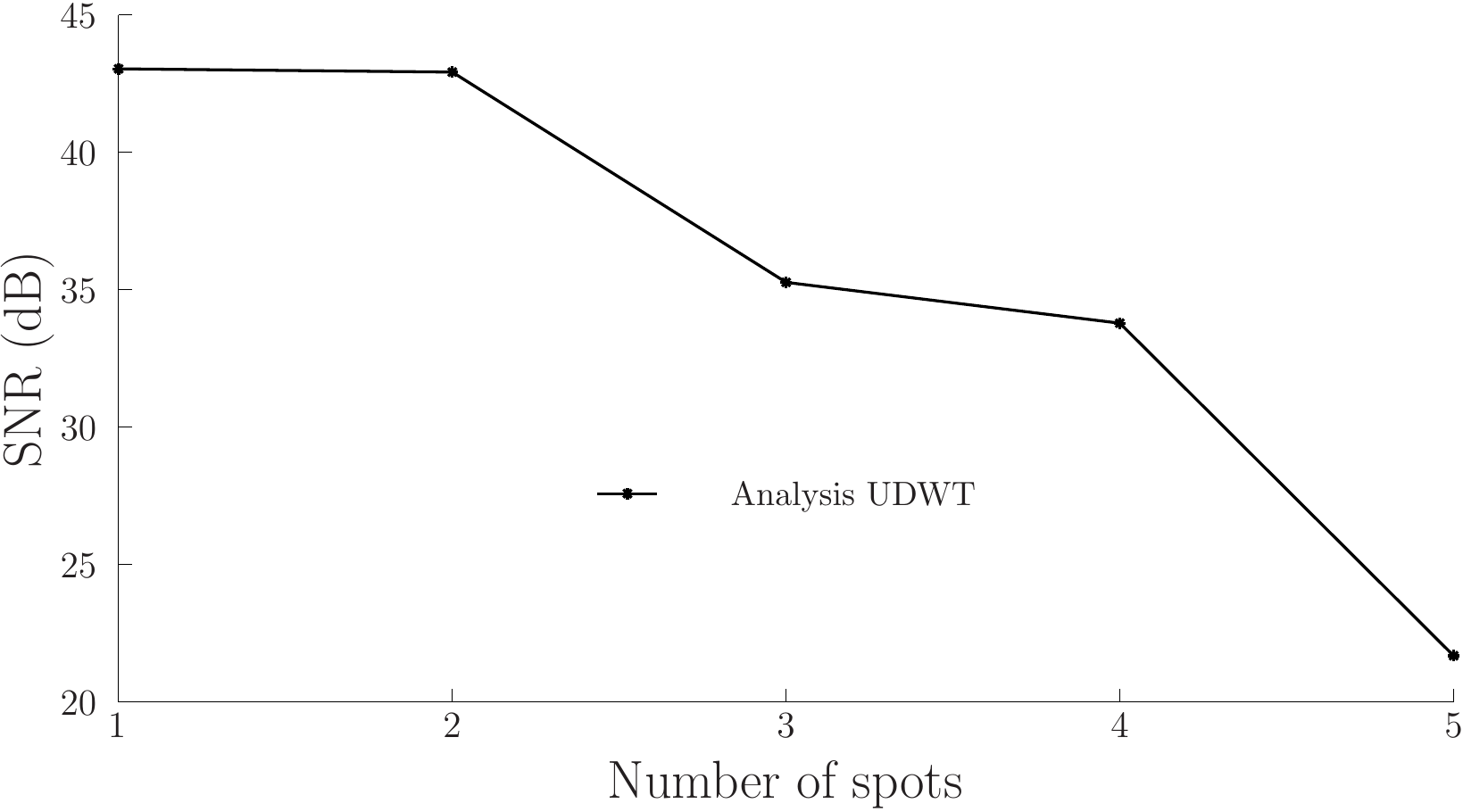}
		\label{subfig:oSNRVsSpots}
	}
	\caption{\subref{subfig:oSNRVsMeas_spots} Output SNR versus number of measurements for synthetic deflection spectra with one and five spots. \subref{subfig:oSNRVsSpots} Output SNR versus number of spots in the deflection spectrum for $\M / \N = 10.99\%$.}
	\label{fig:syntheticPerformance}
\end{figure}

\cref{subfig:oSNRVsSpots} shows the reconstruction performance as a function of the number of spots in the deflection spectrum, for a fixed $\M/\N=10.99\%$. As expected, the performance decreases as the number of spots increases. This is in concurrence with the fact that as the number of spots increase in the spectrum, its sparsity (number of non-zero coefficients) increases and hence would actually require larger number of measurements to have a comparable reconstruction quality. In other words, the error term $\|\spasis^\ast\vecs - \left(\spasis^\ast\vecs\right)_\K\|_1/\sqrt{\K}$  in \eqref{eqn:analysisbpdnreconerr} decays more slowly if $\K$ is high. 

{Additional reconstruction experiments were carried out to assess the benefit of spread spectrum sensing over conventional Hadamard sensing. For Hadamard sensing, the spread spectrum sequence $\bs \sigma$ was simply set to $\bs 1_N$ in the generation of $\bs \Phi_{\rm opt}$. \cref{fig:MultiSense} compares the reconstruction performances of both the methods on synthetic spectrum (with 4 spots) reconstruction. The spread spectrum sensing clearly outperforms the plain Hadamard sensing as already reported by \cite{Puy:2011p1751} in different settings.} 

\begin{figure}
	\centering
	\includegraphics[width=0.5\textwidth]{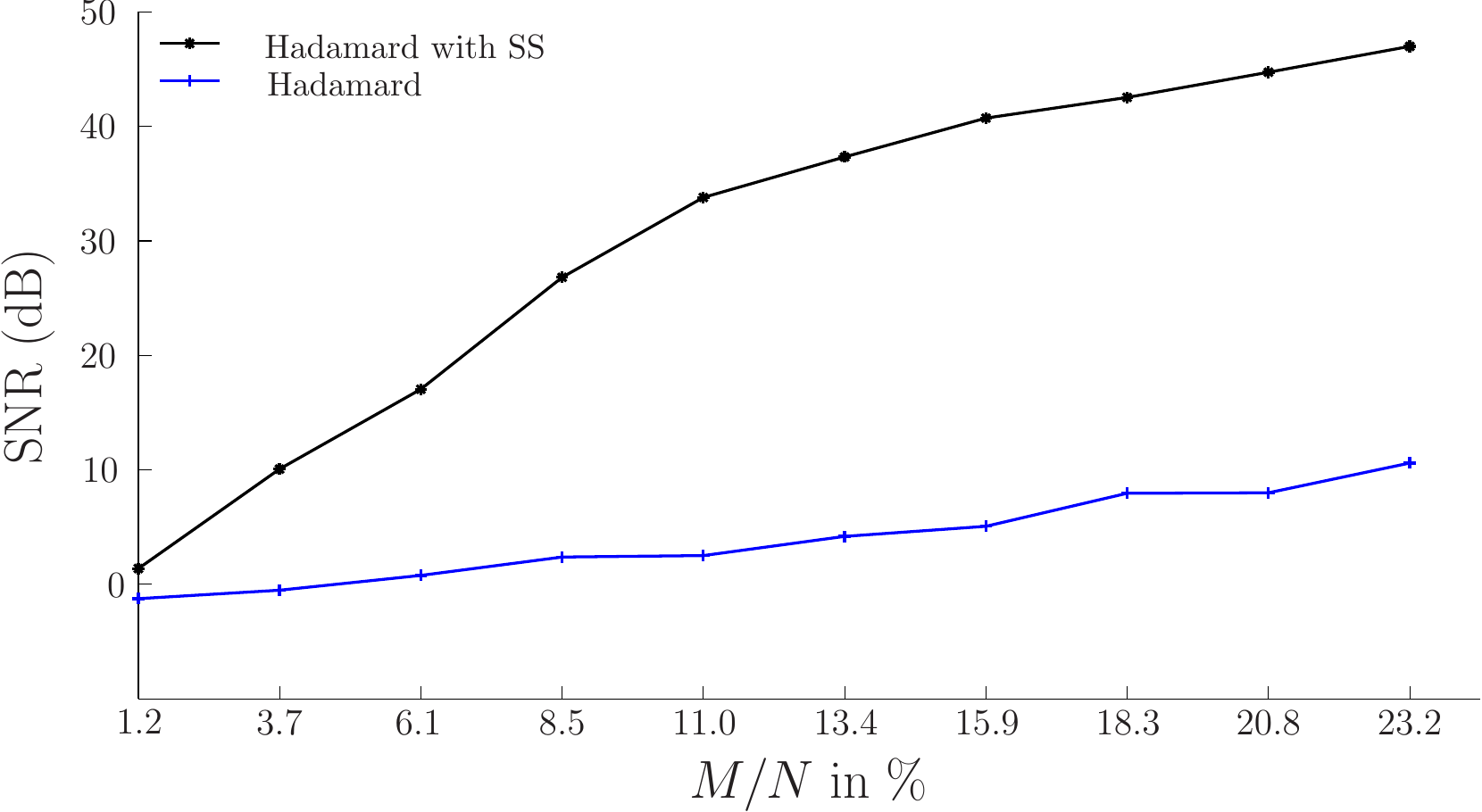}
	\caption{Reconstruction performances of Hadamard sensing with and without spread spectrum modulation. }
	\label{fig:MultiSense}
\end{figure}

The synthetic experiments in the ideal conditions show that the method of spread spectrum compressed sensing is capable of reconstructing deflection spectra with good SNR, using a small number of measurements. With this positive note, we shall now present the reconstruction results obtained using actual deflectometric measurements.

\subsection{Spectrum reconstruction from deflectometric measurements}
\label{subsec:deflectoExperiments}
In this section, we present the spectrum reconstruction results for actual test objects. {All the deflectometric experiments were carried out using Lambda-X's NIMO\texttrademark\cite{lxwww} (see~\mbox{\cref{subfig:nimo}}) deflectometric system, which has a programmable interface to load optical modulation patterns into its SLM.} 

{We consider a SLM region of interest of $64\times64$ pixels positioned at its physical centre, for making the measurements. The SLM pixels outside this area were always set to be opaque. Hence, the size of the modulation patterns and the underlying deflection spectra are $64\times64$ pixels in size, which in the vectorized version have the size $\N=4096$. This makes the size of the Hadamard matrix $4096\times 4096$.}

\paragraph*{Data generation} For experiments with the deflectometric data, we chose two plano-convex lenses of optical powers\footnote{The optical power of a lens is defined as $\mathfrak{D}=1/f$, the reciprocal of its focal length $f$ and it is measured in the unit $m^{-1}$ or \emph{dioptres} ($D$)} $9.99D$ and $60D$. The size of the deflection spectrum to be reconstructed was set to $64\times 64$ pixels, so that $\N=4096$. All the $4096$ possible modulation patterns were generated according to Eq.~\eqref{eqn:sensingbias} and the corresponding number of measurements for all locations $\k$ were collected on the CCD by loading the modulation patterns one by one into the SLM. For a given value of $\M$, a set of indices $\senset\subseteq[\N], |\senset| = \M$, was drawn uniformly and the measurements corresponding the indices in $\senset$ were selected to form the vector $\vecy_\k$, for a given CCD location $\k$. The spectrum was reconstructed by solving \eqref{eqn:analysisbpdnNN} with the UDWT analysis-based prior described in \ref{sec:problemformulation}. The bias $\bs b$ and the value of $\epsilon$ were appropriately set in all the experiments as described in Sec.~\ref{sec:upno-impact-rand} with the estimators of Table~\ref{table:noise-param}. 

{
	\paragraph*{Input SNR} It is important in our reconstruction to evaluate the level of noise corrupting the compressive observations of a deflection spectrum at pixel $k$, \ie the corresponding \emph{input} Signal-to-Noise (SNR) ratio. Having no access to the pure observations of the spectrum, this SNR is approximated by 
	%\begin{equation}
	%\label{eqn:inSNR}
	%{\rm SNR}_{\rm in}(M,N)\ =\ 20 \log_{10} \tfrac{\|\bs \Phi_{\rm opt} \bs s_k\|}{\mathcal E(M,N)}\ \approx\ 20 \log_{10} \tfrac{\|\bs \Phi_{\rm opt} \bs s^{\rm no}\|}{\mathcal E(M,N)},
	%\end{equation}
	\begin{equation}
	\label{eqn:inSNR}
	{\rm SNR}_{\rm in}\ =\ 20 \log_{10} \tfrac{\|\bs \Phi_{\rm ss} \bs s_\k\|}{ \|\vecz_\k - \bs \Phi_{\rm ss} \bs s_\k\|}\ \approx\ 20 \log_{10} \tfrac{\|\bs \Phi_{\rm ss} \bs s^{\rm no}\|}{\|\vecz_\k - \bs \Phi_{\rm ss} \bs s^{\rm no}\|},
	\end{equation}
	with the ``no object'' spectrum $\bs s^{\rm no}$ obtained in Sec.~\ref{sec:uppo-spre-funct}. Note that the input SNR is computed using $\vecz_\k$, not $\vecy_\k$, because this is the vector that will be finally used for spectrum reconstruction. As both $\|\vecz_\k - \bs \Phi_{\rm ss} \bs s_\k\|^2$ and $\|\bs \Phi_{\rm opt} \bs s_k\|^2$ are (approximately) proportional to $M$, we observed that this input SNR has an almost constant value of about 4.8 dB, for most $\k$.} 

\paragraph*{Results}

\begin{figure}[t]
	\centering
	\subfigure[]{
		\includegraphics[width=0.2\textwidth]{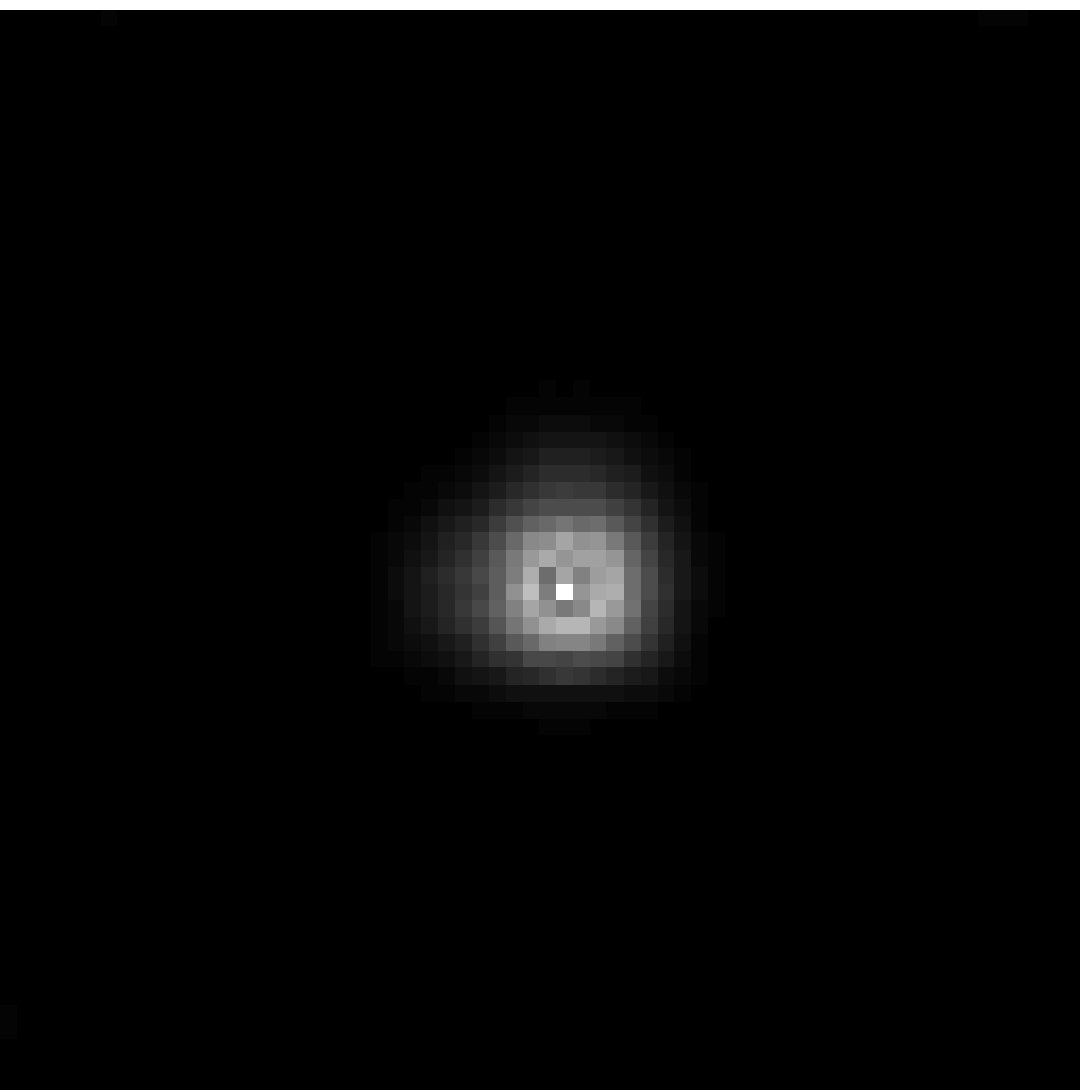}
		\label{subfig:M4096AnaUDWT}
	}
	\subfigure[]{
		\includegraphics[width=0.2\textwidth]{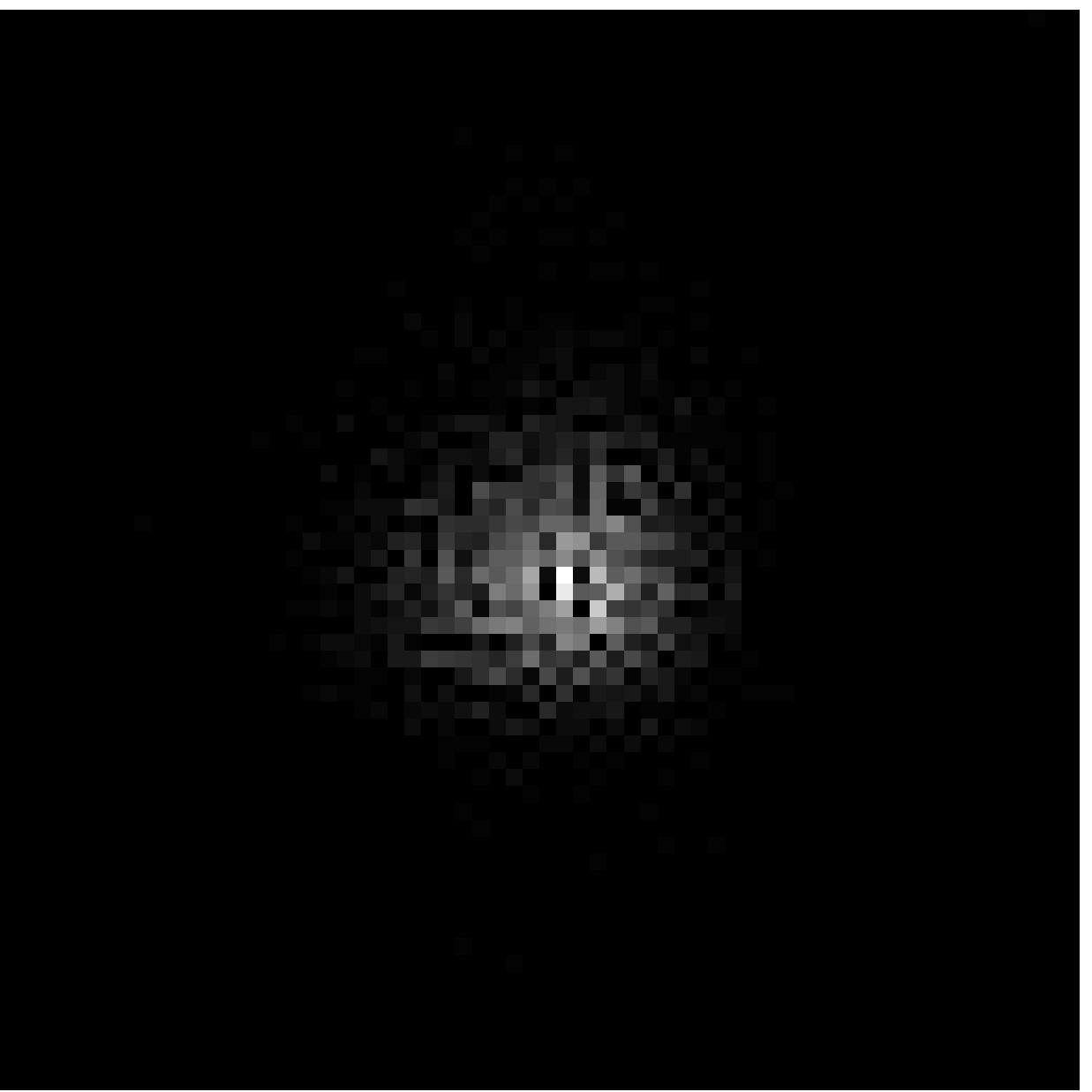}
		\label{subfig:M150AnaUDWT}
	}
	\subfigure[]{
		\includegraphics[width=0.2\textwidth]{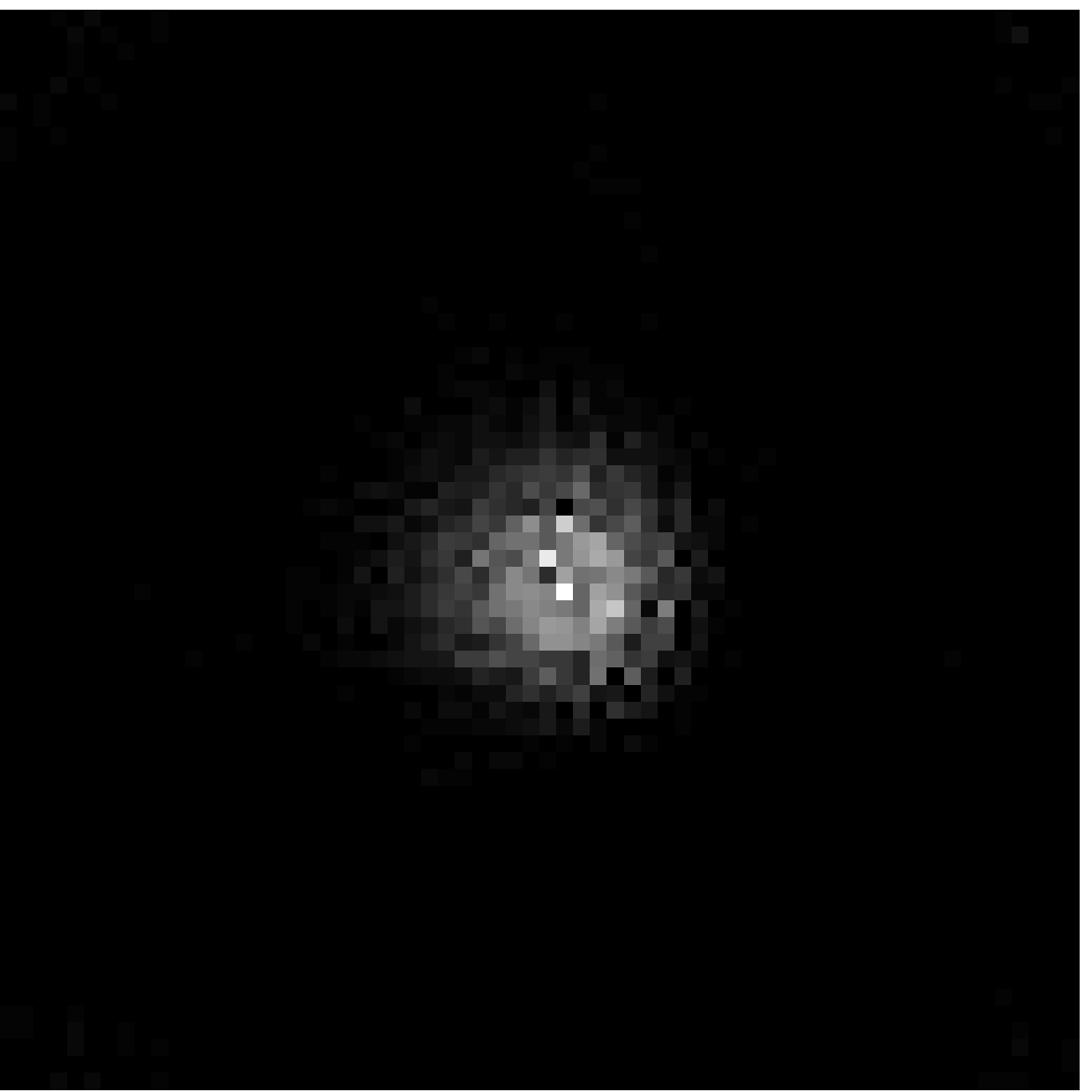}
		\label{subfig:M250AnaUDWT}
	}
	\caption{{Reconstruction examples of a spectrum of a $9.99D$ plano-convex lens:~\subref{subfig:M4096AnaUDWT} corresponds to $100\%$ of measurements,~\subref{subfig:M150AnaUDWT} corresponds to $11\%$ of measurements and ~\subref{subfig:M250AnaUDWT} corresponds to $15.8\%$ of measurements.}}
	\label{fig:CALI2006reconExamples}
\end{figure}

\begin{figure}[t]
	\centering
	\includegraphics[width=0.5\textwidth]{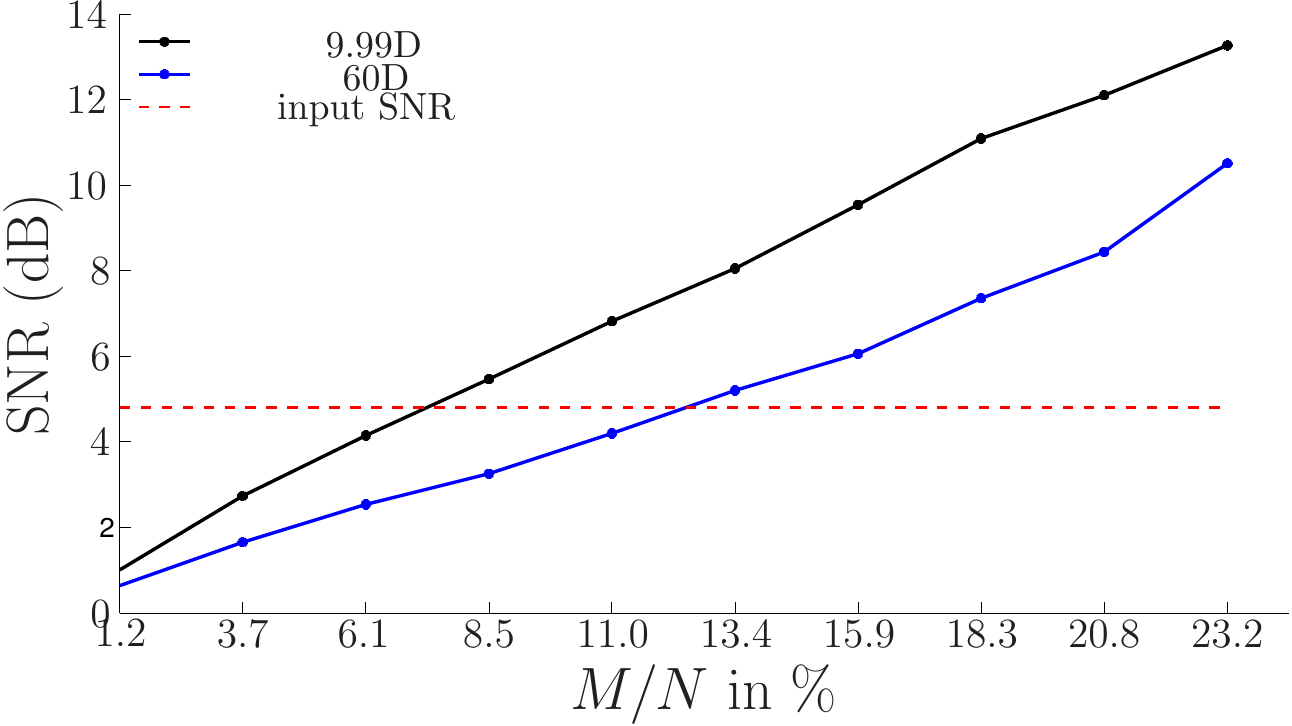}
	\caption{Output SNR versus number of measurements for analysis UDWT reconstruction mode for plano-convex lenses of optical powers $9.99D$ and $60D$.}
	\label{fig:64x64SpectrumRecon}
\end{figure}

Fig.~\ref{fig:CALI2006reconExamples} shows typical reconstructions of a deflection spectrum for the $9.99D$ lens, at some arbitrary location $\k$, {for $100\%$, $11\%$ and $15.8\%$ measurements}. It can be seen from these example reconstructions that the quality of reconstruction improves as the number of measurements increases. 

Fig.~\ref{fig:64x64SpectrumRecon} shows the SNR curves for the spectra of the two specified lens ($9.99D$ and $60D$), as a function of the ratio $\M/\N$ in $\%$. Each point on the plot was obtained by considering 4 different locations on the CCD and performing 20 independent reconstruction trials on each of them for every value of $\M/\N$ considered and subsequently averaging over all locations and trials. For each new trial, a new index set $\senset$ was generated and used. 

For assessing the reconstruction quality gain, the input SNR is represented by an horizontal dotted line near $5$ dB. To recollect, the input SNR tells us the relative strengths of signal and noise in the input which is used for reconstruction. From the plots, we can infer that the compressive sensing method produces a reconstruction that has better SNR compared to the input SNR. This performance is owed to the fact that the reconstruction procedure has the capability to take advantage of the sparse structure of the signal to carefully discriminate it from noise and suppress the latter. 

{T}he number of measurements needed for a $10$ dB reconstruction is about $\M/\N=16\%$ for the $9.99D $ lens and $\M/\N=21\%$ for the $60D$ lens, which is far lesser than the number of measurements needed according to classical Shannon sampling.

On a Macbook Pro\texttrademark with a 2GHz processor and 4GB RAM, {the reconstruction} takes about 3 minutes. However, when the ratio $\M/\N$ increases, the time required for reconstruction marginally decreases. Reconstruction of spectra for all the locations on the object surface is still a bottleneck due to computational requirement, but quicker characterization of objects can be achieved without reconstruction as described in the next section.

\section{Compressive characterization of light deflection spectra}
\label{sec:CPE}
The final goal of measuring and reconstructing deflection spectra is to characterize and understand the object being studied. Though deflection spectra at individual locations themselves contain rich information about the local geometry of the object, it is essential to see how the spectrum evolves as a function of the spatial location $\k$. {To this end} we would like to summarize each deflection spectrum by a few parameters. In this section, we characterize a spectrum by defining the notion of \emph{centrum} of a deflection spectrum, \emph{by assuming that each deflection spectrum contains only one deflection spot}, and present a method to compute it using compressive measurements.

The PSS method in~\cref{sec:pss} is also aimed at obtaining local deflection information across the whole surface. However, it is effective and accurate only when each deflection spectra contains only one Dirac spike. On the other hand, our proposed method considers generic deflection patterns and hence it is expected to be more effective. 

\subsection{Parametric characterization of a deflection spectrum}
\label{subset:parametricCharacterization}

For smooth objects, the deflection spectrum $\vecs_\k$ is simply a bright spot centered at a mean deflection angle that depends on the location $k$. The evolution of the mean deflection angle across $k$, which can be studied via the locations of the spectral spot, provides a global characterization of the object's surface. 

We fix the spectral origin $O$ (co-ordinates of zero deflection with respect to optical axis) and locate the spot with respect to $O$. In an ideal noiseless case, we could simply compute the geometric centroid to localize, but this fails in our current situation. Alternatively, we fixed a template 2-D Gaussian $\matg_\rho$ of variance $\rho$\footnote{The radius $\rho$ was set by analyzing the width of the spectral spot in a sample reconstruction of deflection spectrum.}, located on the spectral origin $O$. By letting $\vecg^\rho_\paramtrans$ to denote the vectorized form of $\matg_\rho$, translated in 2 dimensions by $\paramtrans=(r_x, r_y)^T\in\realno^2$, we propose to find the position of the spectral spot within the deflection spectrum by solving
\begin{equation}
\label{eqn:centroid}
{\paramtrans}_\k^\ast = \arg\max_{\paramtrans}\,|\langle\hat{\vecs}_\k,\vecg^\rho_\paramtrans\rangle|.
\end{equation}

Finding ${\paramtrans}_\k^\ast$ in this manner involves a fully reconstructed spectrum $\hat{\vecs}_\k$ and it is computationally challenging. However, thanks to the embedding properties of compressed sensing matrices, characterized by the restricted isometry property~\cite{Foucart:2013}, a similar matched filtering operation can be instead performed on the vector of measurements $\vecy_\k$ itself, without involving reconstruction. 

\subsection{Matched filtering using compressive samples}
\label{subsec:smashedFiltering}
Many common signal processing tasks such as detection, classification and parameter estimation can be performed using the compressive samples, without fully reconstructing the signals~\cite{Davenport:2007p2548, Davenport:2010p589}. On similar lines, we propose to localize the spectral spot of a deflection spectrum directly from its measurements $\vecy_\k$ by posing the matched filtering problem~\eqref{eqn:centroid} in the measurement domain, appropriately named \emph{smashed filtering}~\cite{Davenport:2007p2548}. 
{Considering a limited impact of the noise on the procedure, this is done by first computing a de-biased measurement vector
	\begin{equation}
	\label{eq:zk-estim}
	\bs z_k := \big(2\bs y_k - \bs 1_M (\bs 1_N \bs s_k)\big)/\sqrt N\ \approx \ \bs \Phi \bs s_k,
	\end{equation}
	where $(\bs 1_N \bs s_k)$ is actually measured by the schlieren deflectometer by setting the SLM with the pattern $\bs 1_N$, and then estimating
	\begin{equation}
	{\paramtrans}_\k^\ast= \arg\max_{\paramtrans}\,|\scp{\bs z_\k}{\slmpat\vecg^\rho_\paramtrans}| = \arg\max_{\paramtrans}\,|\langle\slmpat^T\bs z_\k, \vecg^\rho_\paramtrans\rangle|.
	\label{eqn:compcentroid}
	\end{equation} 
	Under certain assumptions, solving~\eqref{eqn:compcentroid} is similar to solving~\eqref{eqn:centroid}. If the sensing matrix $\bs\Phi$ is RIP or even D-RIP with respect to a sparsity basis/dictionary $\bs\Psi$ where both $\vecg^\rho_\paramtrans$ and $\bs s_k$ can be sparsely represented (see Sec.~\ref{sec:problemformulation}), then, $\scp{\bs\Phi\bs s_k}{\bs\Phi \vecg^\rho_\paramtrans} \approx \scp{\bs s_k}{\vecg^\rho_\paramtrans}$, where the level of approximation is actually regulated by the associated Restricted Isometry Constant (RIC) \cite{Davenport:2007p2548, Davenport:2010p589,Foucart:2013}: smaller the RIC, tighter the approximation.} 

Eq.~\eqref{eqn:compcentroid} is solved in two steps: (1) a coarse estimate $\hat{\paramtrans}_\k$ via convolution and peak detection and (2) finer sub-pixel estimate ${\paramtrans}_\k^\ast$ by a gradient descent around $\hat{\paramtrans}_\k$.

\subsection{Results}
For the experimental evaluation of spectral spot estimation using compressive measurements, we retained the configuration described in~\cref{subsec:deflectoExperiments}. For each of the 4 CCD locations $\k$, the centroid ${\paramtrans}_\k$ was computed by solving Eq.~\eqref{eqn:compcentroid}. A ``ground truth" centroid ${\paramtrans}_\k^\ast$ was also found by solving Eq.~\eqref{eqn:centroid}, using the fully reconstructed spectrum $\vecs_\k$ (solving~\eqref{eqn:bpdn}) using $\M = 4096$ ($100\%$) measurements with UDWT dictionary.

Fig.~\ref{fig:compcentroid} shows the centroid computation error $\ltwoof{{\paramtrans} - {\paramtrans}^\ast}$ as a function of the number of measurements $\M/\N$. Each data point is obtained by averaging 50 independent trials for each value of $\M$ over all the four locations. The horizontal dotted line indicates a unit pixel error and it can be seen that the compressive centroid estimation achieves sub-pixel accuracy, even with the number of measurements as low as $2.4\%$ ($50$ measurements) for the $60$D lens, and about $3.7\%$ for the $9.99$D lens.  

Compressive centroid computation takes less than a second on the same hardware used for reconstruction and this faster method can be used to study the global geometry of object surfaces.

\begin{figure}
	\centering
	\includegraphics[width=0.55\textwidth]{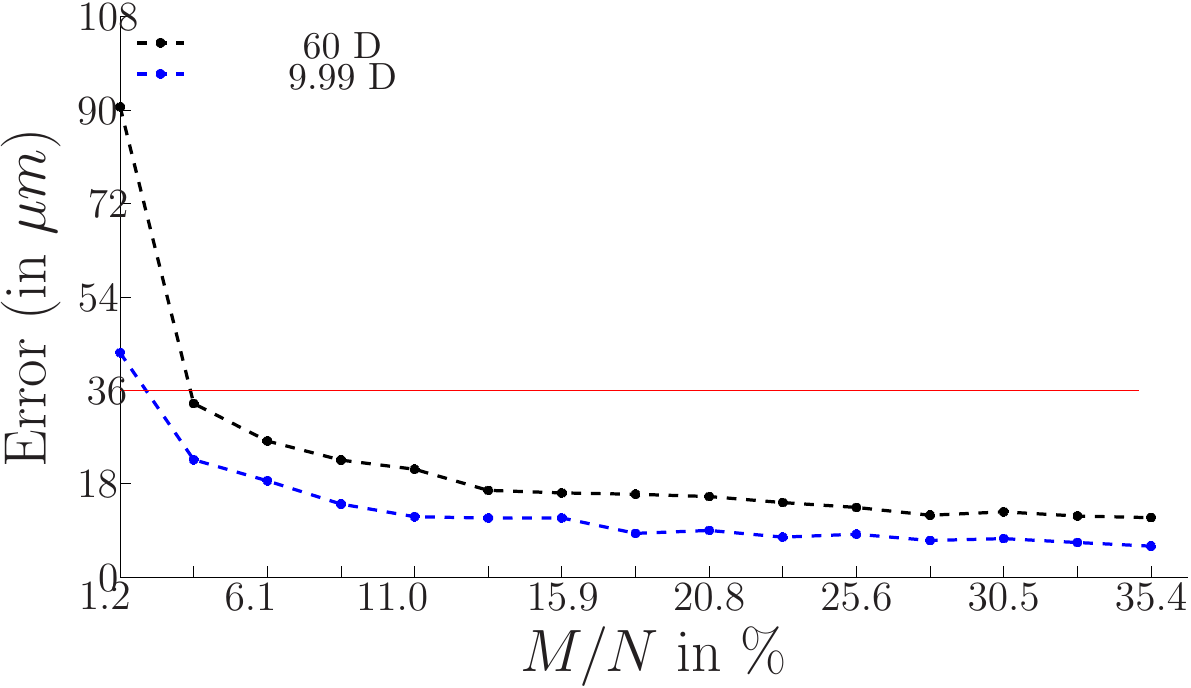}
	\caption{Average error in centroid estimation (in pixels) using compressive measurements, as a function of number of measurements, in \%.}
	\label{fig:compcentroid}
\end{figure}

\section{Characterization of objects from deflection spectra}
\label{sec:geom}

\begin{figure}
	\centering
	\includegraphics[width=0.6\textwidth]{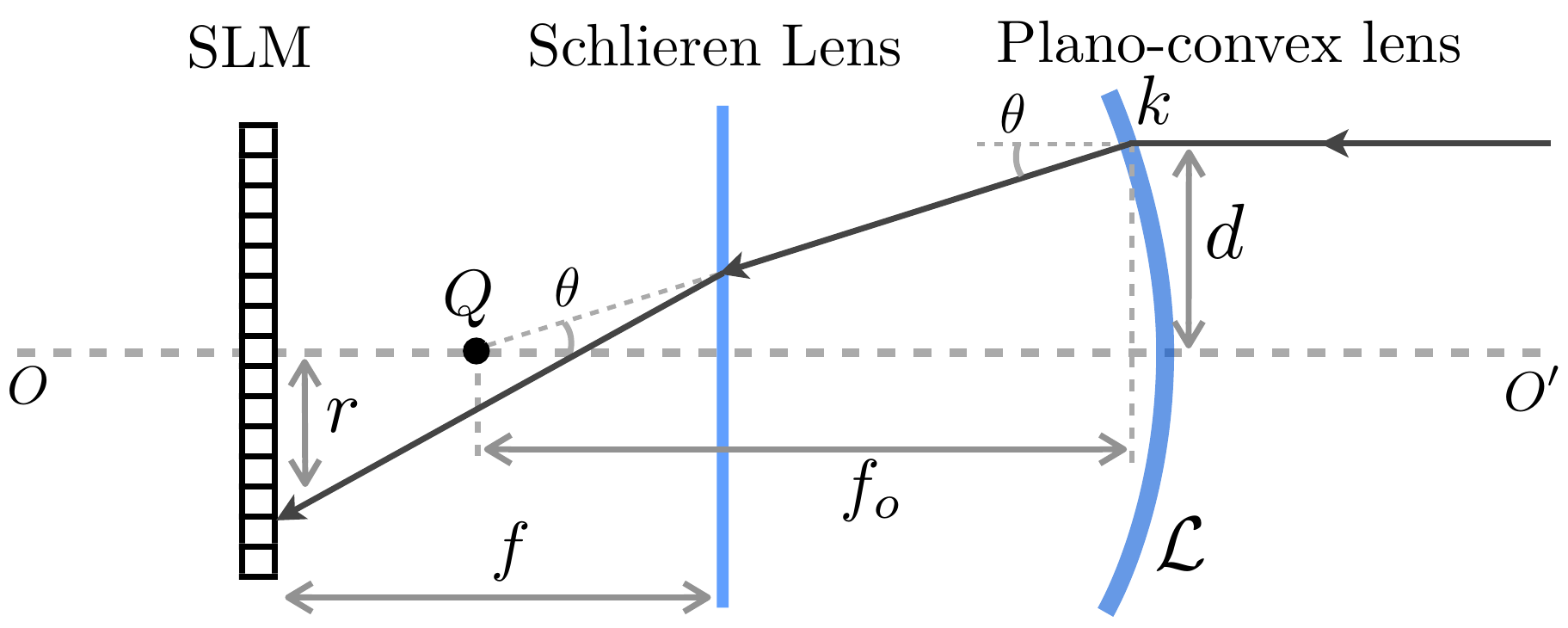}
	\caption{Relationship between deflection angle and the position of the deflection spectrum on the SLM.}
	\label{fig:geometry}
\end{figure}
The relationship between the object locations $\k$ and the position $\paramtrans_\k$ of the spectral spot in the spectrum $\vecs_\k$ can be understood with the 1-D schematic in~\cref{fig:geometry}\footnote{This is exactly the same as described in~\cref{subfig:pss}, without the tele-centric arrangement, which is not necessary for the following discussion}. By making the thin lens assumption, the relationship between the focal length of the schlieren lens $f$, dioptric power of the test object $1/f_o$ and the vertical distance $d$ of a considered point $\bs k$ on the test object is given by
%Inverting the relationship, we have 
\begin{equation}
\frac{1}{f_o} \approx \frac{1}{f}\frac{r}{d},
\label{eqn:powerExpression}
\end{equation}
where $r$ is the linear displacement on the SLM of the light ray incident on $\bs k$, parallel to the optical axis.

By computing the distances $r$ and $d$ in physical units (with the knowledge of the pitch of the CCD and the SLM pixels) and knowing the focal length $\f$ of the schlieren lens, we can compute the \emph{local optical power} for each location $\k$ on the object. For a plano-convex lens, the local optical power is the same across all the object locations. 

\subsection{Characterization of plano-convex lenses}
\label{subsec:geomplanoconvex}
For the
two plano-convex lenses of optical powers $9.99D$ and $60D$, already
used in~\cref{subsec:deflectoExperiments}, we considered an array of
$65\times 65$ CCD pixels ($513.42\times513.42 \mu m^2$ with $7.8989\mu m$ CCD pixel pitch, around the centre of the full CCD array of
size $1392\times1040$). Then, the
positions $\paramtrans_\k$ of all the spectral spots $\vecs_\k$
corresponding to the $65\times 65$ CCD locations were found using the
compressive matched filtering method, described
in~\cref{subsec:smashedFiltering}, for different values of the number
of measurements $\M$. The total time required to find the centroid of
all the $65\times 65$ spectra was about one hour and ten minutes on
the same machine that was used for the experiments
in~\cref{sec:recoveryResults}.

\cref{subfig:CALI2012_3D_m4096}-\subref{subfig:CALI2012_3D_m250} show the surface plots of $\|\paramtrans_\k\|_2$ as the function of location $\k\in\realno^2$, computed using $\M/\N = 100\%$ and $8.5\%$ measurements for the $60D$ plano-convex lens. Let us refer to these plot as \emph{centroid maps}. The centroid maps are symmetric due to the symmetry of plano-convex lenses. To robustly estimate the constant rate of centroid evolution, a two dimensional inverted cone is fit to each deflection map, in the least squares sense. The inverted cone is fit by optimizing its slope to minimize its squared error with the deflection map.~\cref{tab:slope} shows the values of the slopes computed for the centroid evolution plots corresponding to the two lenses.  

\begin{figure}
	\centering
	
	\subfigure[]{
		\includegraphics[width=0.3\textwidth]{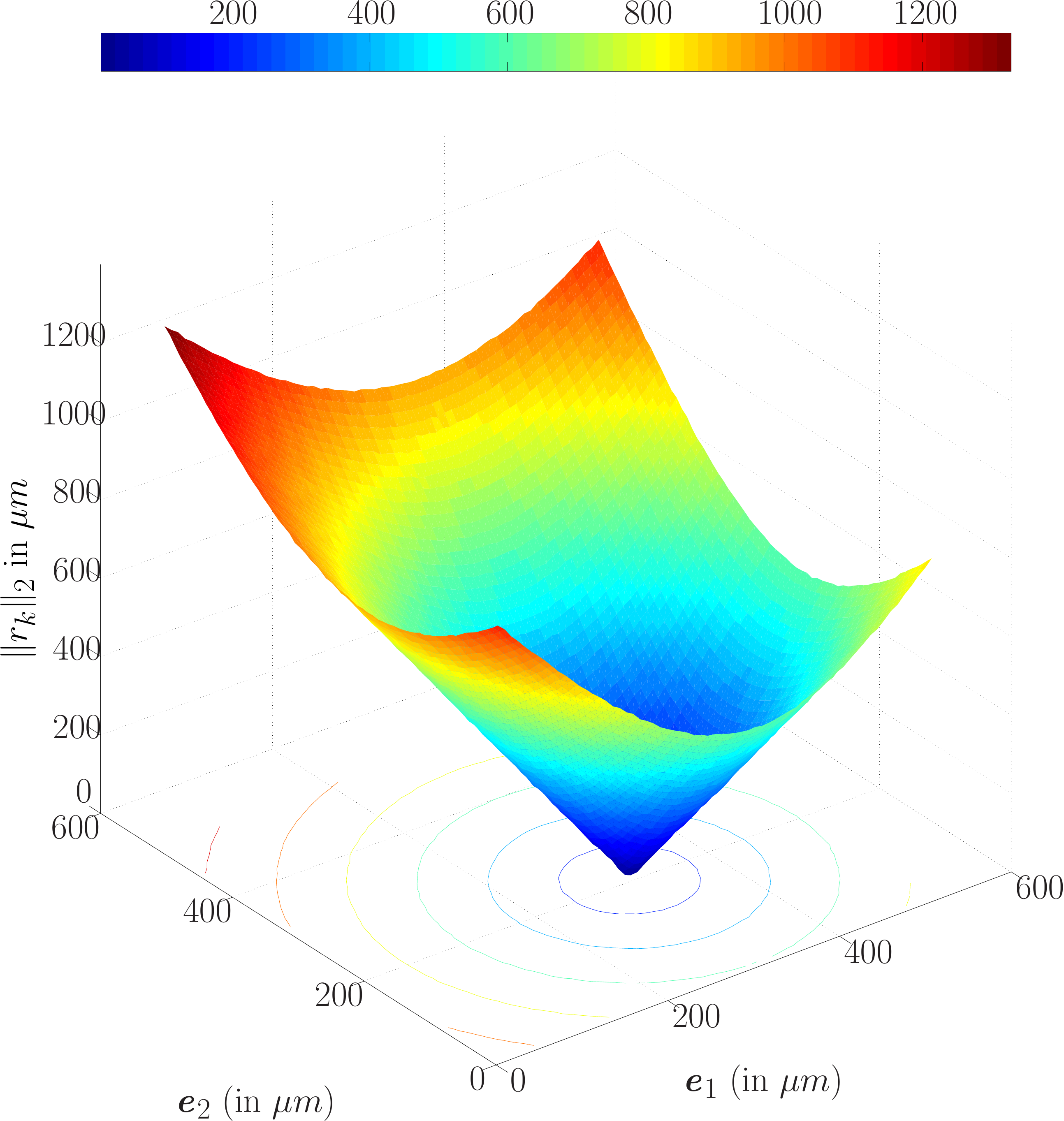}
		\label{subfig:CALI2012_3D_m4096}
	}
	\subfigure[]{
		\includegraphics[width=0.3\textwidth]{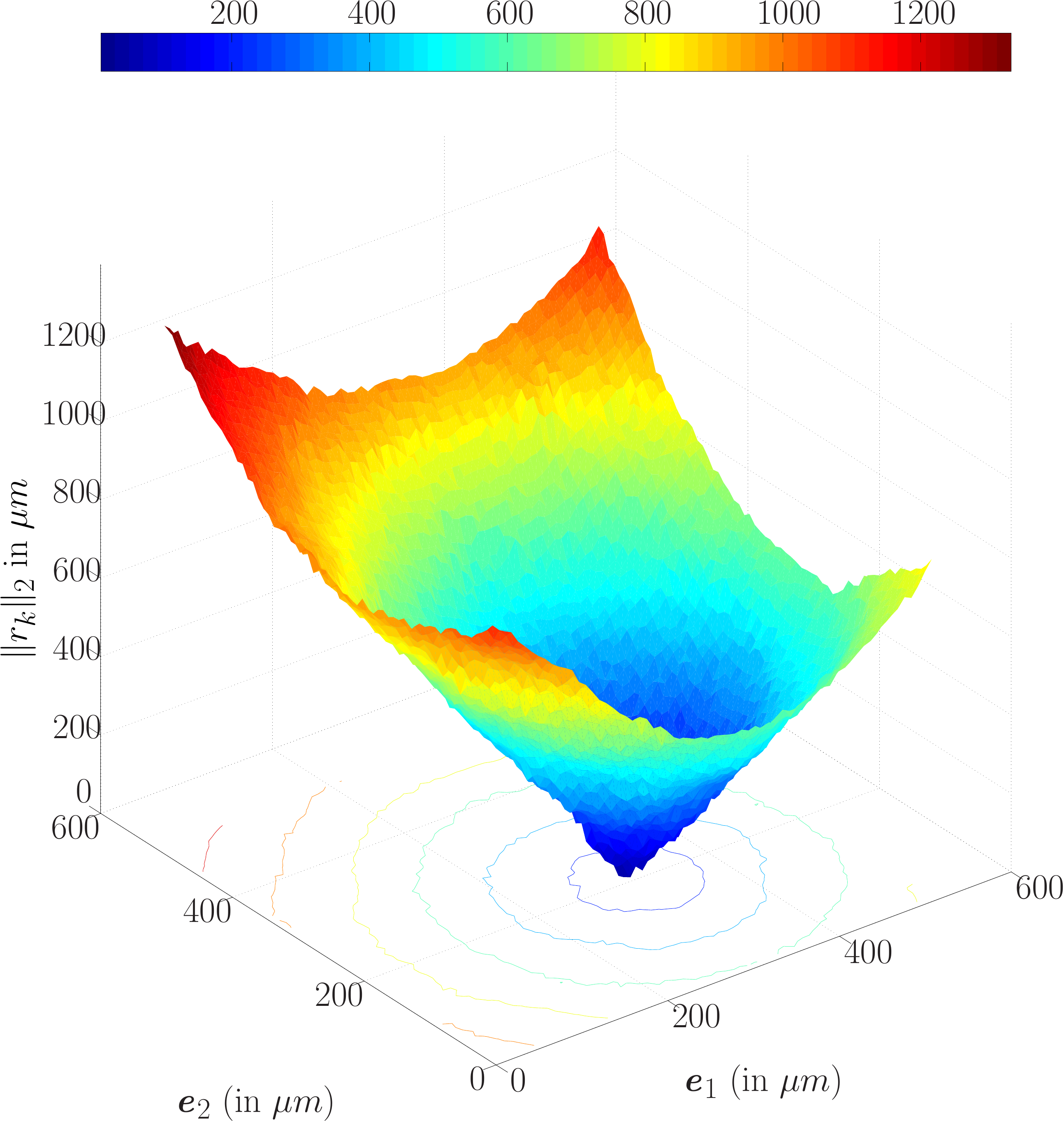}
		\label{subfig:CALI2012_3D_m250}
	}
	\caption{Evolution of deflection spectrum along the spatial locations of the object, computed using\subref{subfig:CALI2012_3D_m4096} $\M/\N=100\%$ and \subref{subfig:CALI2012_3D_m250} $\M/\N=8.5\%$ of measurements.}
	\label{fig:centrumEvolM4096}
\end{figure}

\begin{table}[h!]
	\centering
	\begin{tabular}{|c|c|c|c|c|}
		\hline
		Optical power & \multicolumn{2}{c|}{$m_o$} &  \multicolumn{2}{c|}{Estimated power} \\
		\hline
		& $\M/\N=100\%$ & $\M/\N=8.5\%$ & $\M/\N=100\%$& $\M/\N=8.5\%$ \\
		\hline
		$9.99D$ & $0.1128$ & $0.1065$ & $10.2785D$ & $9.7115D$\\
		\hline
		$60D$ & $0.6731$ &     $0.6743$ &$61.3544D$ & $61.4612D$\\
		\hline
	\end{tabular}
	\caption{Table of computed slopes and dioptric powers using compressive measurements with $\M/\N=100\%$ and $\M/\N=8.5\%$ measurements.}
	\label{tab:slope}
\end{table}

The dioptric power of the lens is computed using (due to Eq.~\eqref{eqn:powerExpression}) $1/\f_o = m_o/\f$, with schlieren lens of focal length $50 mm$ and SLM pixel pitch $36\mu m$. The third column of~\cref{tab:slope} lists the estimated dioptric powers of the two lenses. 
 
~\cref{fig:powerError} plots the absolute percentage error (with respect to the true power) as a function of the number of measurements.  As expected, in general the accuracy of the estimated optical power of the plano-convex lens, using compressive samples improves as the number of measurements increases. The systematic error in the power estimation of the $60D$ lens is unexplained and it may be arising from the bad calibration of the system parameters. 

\begin{figure}
	\centering
	\includegraphics[width=0.6\textwidth]{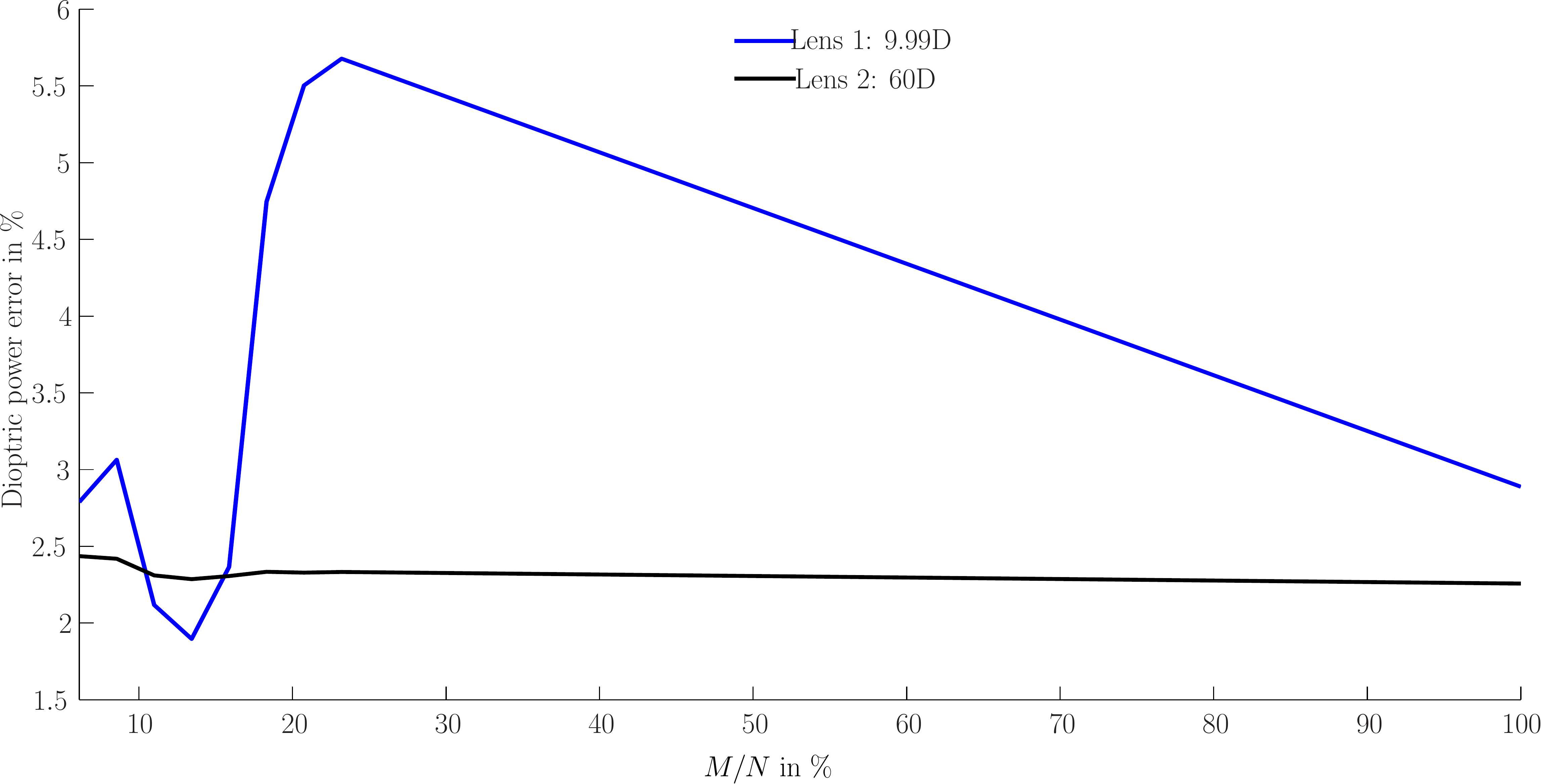}
	\caption{Percentage error in dioptric power estimation as a function of the number of measurements.}
	\label{fig:powerError}
\end{figure}

\subsection{Characterization of diffractive multi-focal intra-ocular lenses}
A diffractive Multifocal Intra-Ocular Lens (MIOL) has multiple foci, achieved by gratings carved on its surface as shown in~\cref{fig:MIOLShapeAndProfile}. The underlying curvature of the lens (blue dotted curve), defines the refractive aspect of the diffractive lens and controls the location of zero-order diffraction, which is utilized for distant vision. 

We considered a diffractive MIOL of dioptric powers $28D$ and $30.25D$ (two foci). A region of $256\times 256$ was considered on SLM for sensing the deflection maps. On the CCD array, the deflections were computed using compressive centroid method on an array of  $500 \times 500$ pixels (physical area of $3.95 mm \times 3.95 mm$). On a computational grid, the computations took about five hours.

\begin{figure}
	\centering
	\subfigure[]{
		\centering
		\raisebox{1.5\height}{\includegraphics[width=0.3\textwidth]{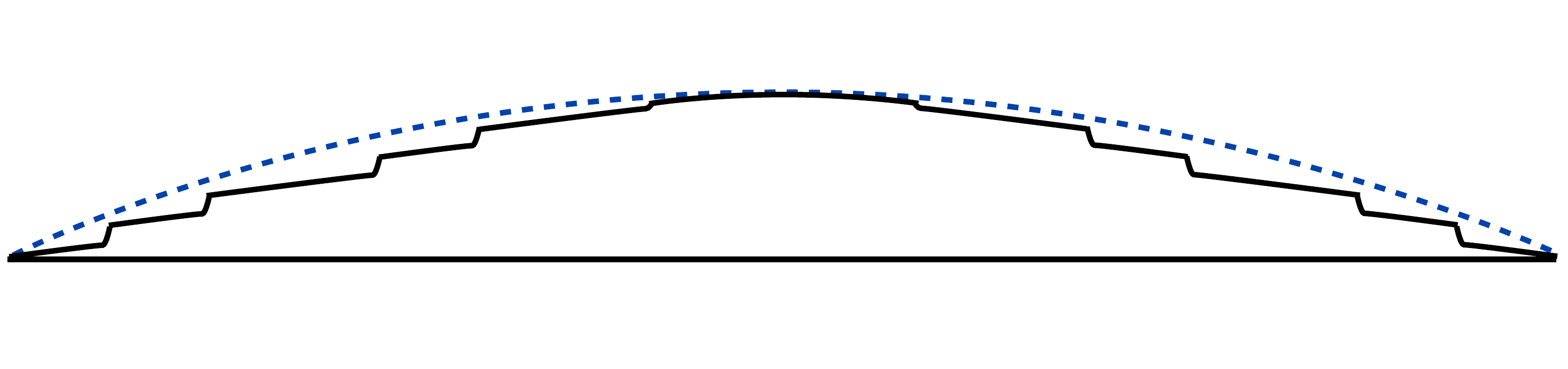}}
		\label{fig:MIOLShapeAndProfile}
	}
	\subfigure[]{
		\centering
		\includegraphics[width=0.3\textwidth]{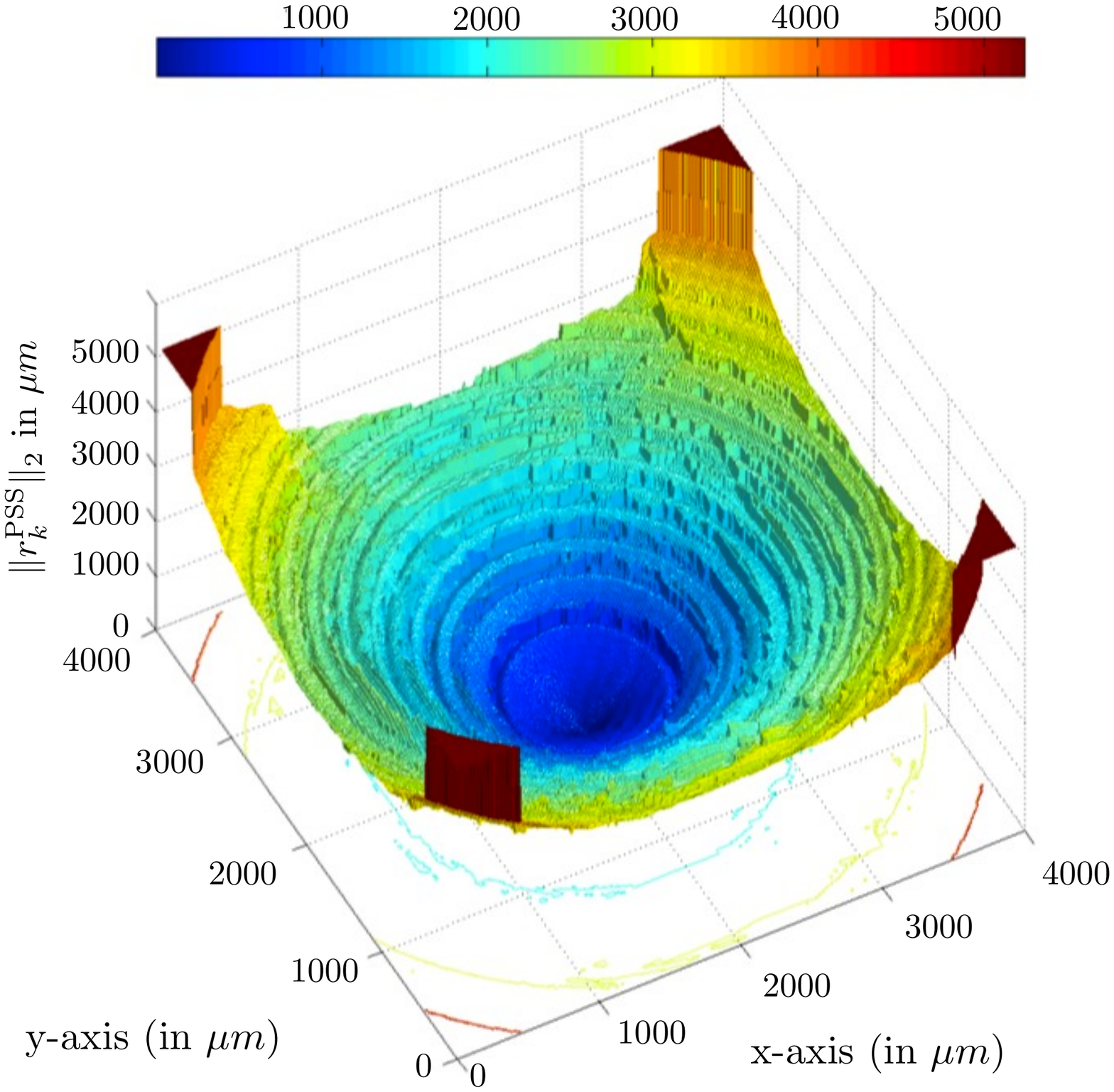}
		\label{subfig:PSS_MFD_3D}
	}
	\subfigure[]{
		\centering
		\includegraphics[width=0.3\textwidth]{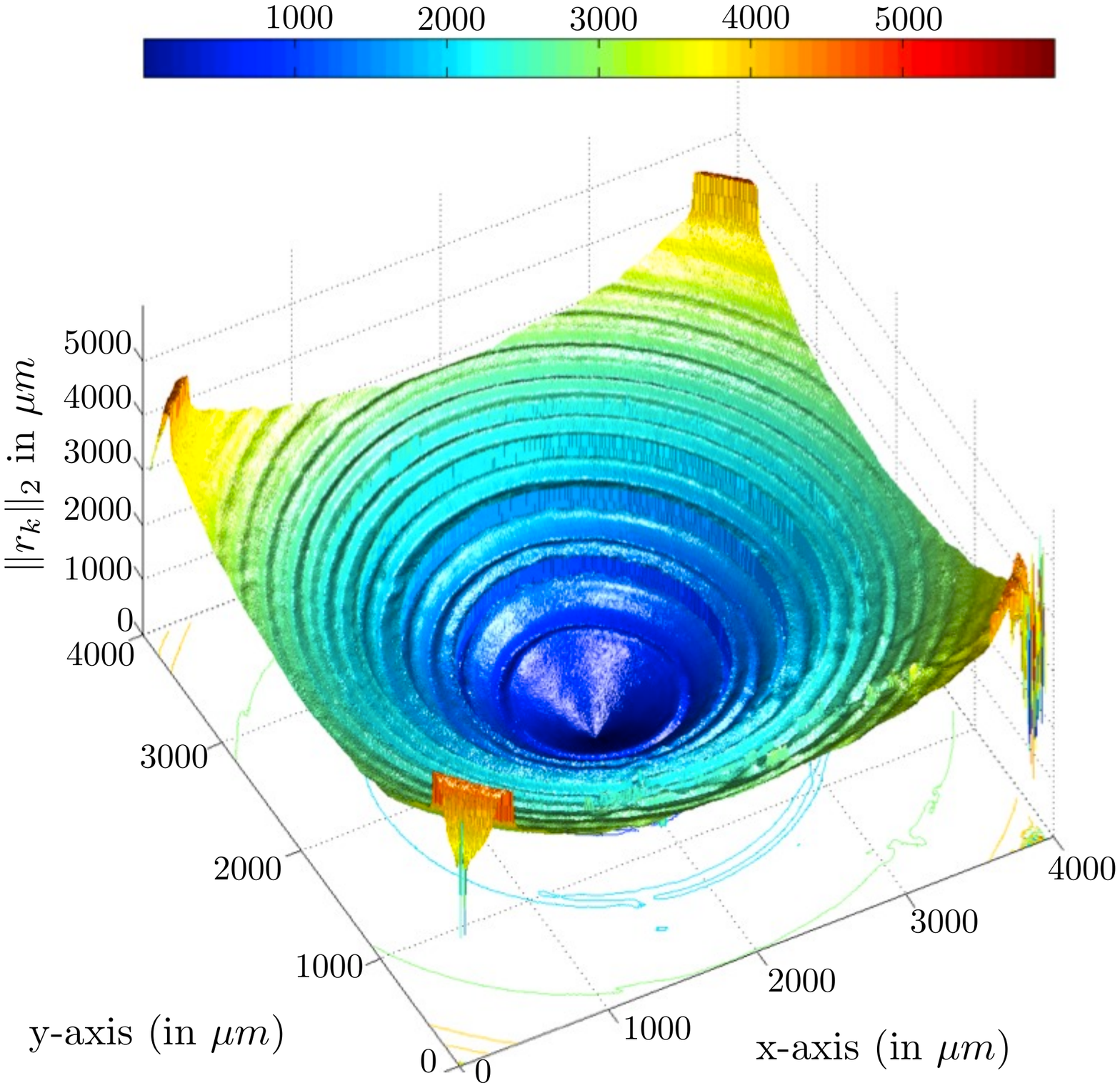}
		\label{subfig:MFD_3D_m5000}
	}
	\caption{\subref{fig:MIOLShapeAndProfile} Typical profile of a diffractive multifocal intra-ocular lens. 3D plots of centroid maps obtained using~\subref{subfig:PSS_MFD_3D} phase shifting schlieren method and~\subref{subfig:MFD_3D_m5000} compressive sensing method with $7.63\%$ measurements.}
	\label{fig:MFD_3D}
\end{figure}

\begin{figure}
	\centering
	\subfigure[]{
		\centering
		\includegraphics[width=0.5\textwidth]{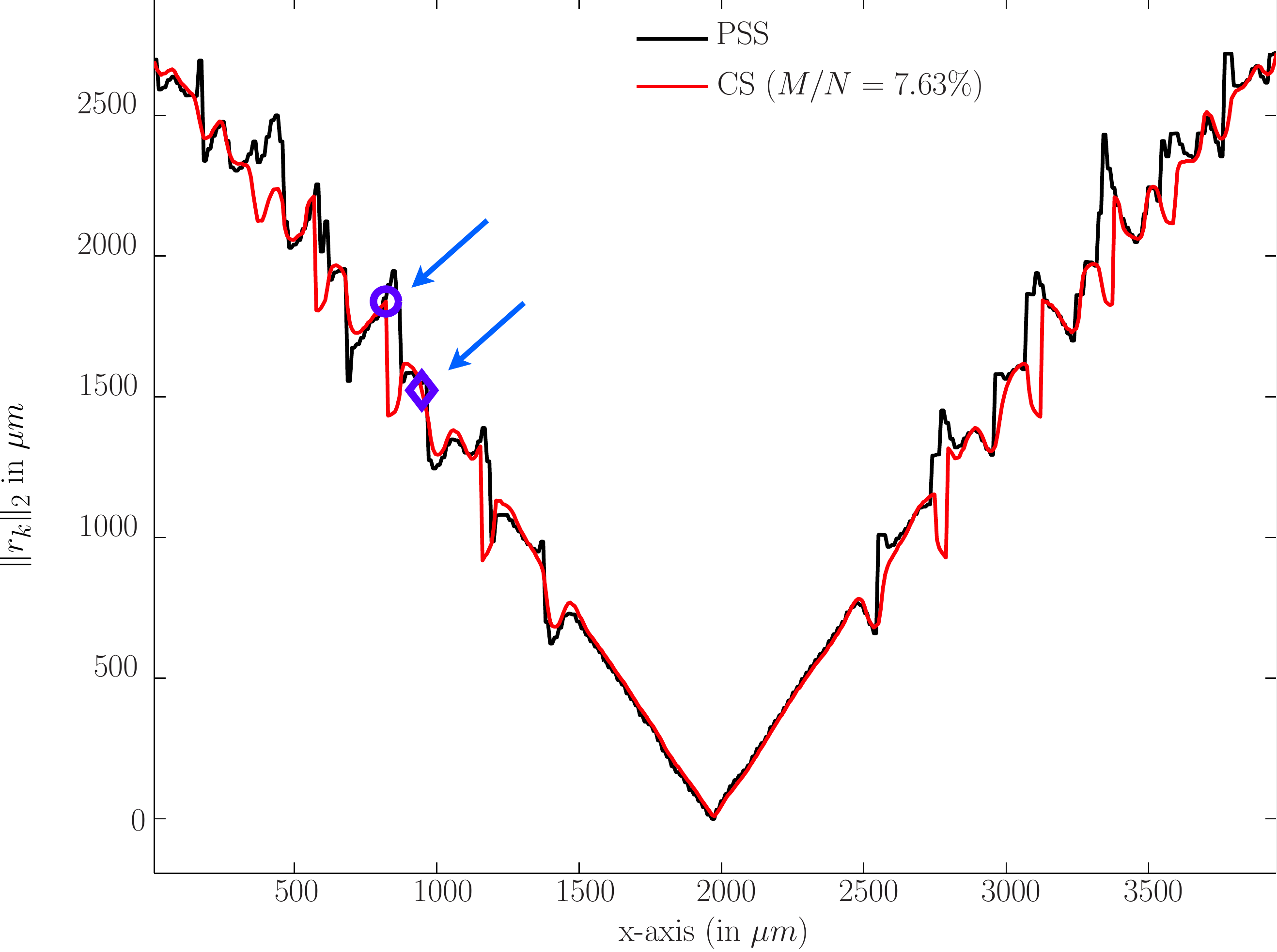}
		\label{subfig:MFD_profile_m5000}
	}\\
	\subfigure[]{
		\centering
		\includegraphics[width=0.3\textwidth]{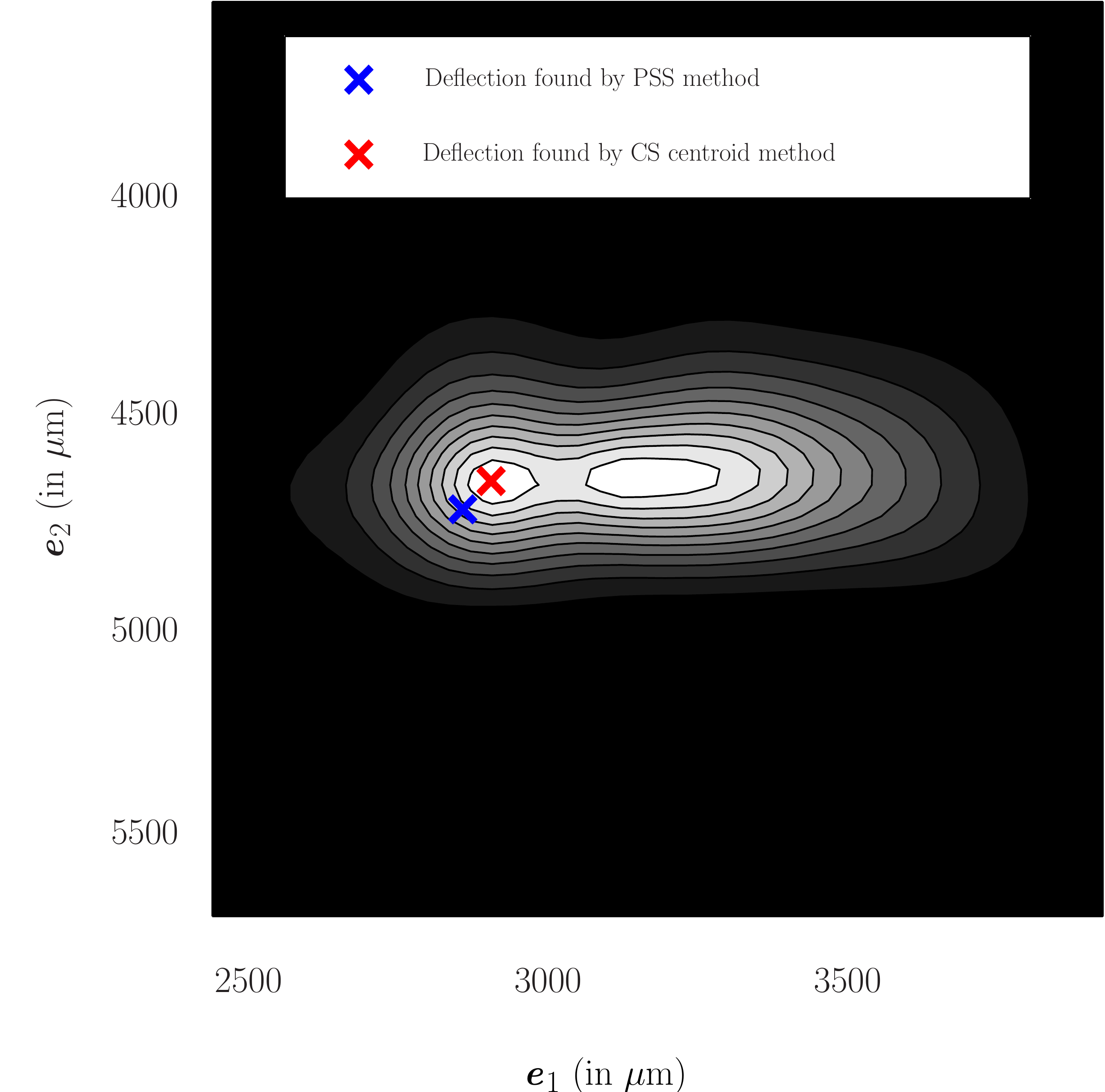}
		\label{subfig:MFD_combinedPlot_zoom_rr1_cc104_M5000}
	}
	\subfigure[]{
		\centering
		\includegraphics[width=0.3\textwidth]{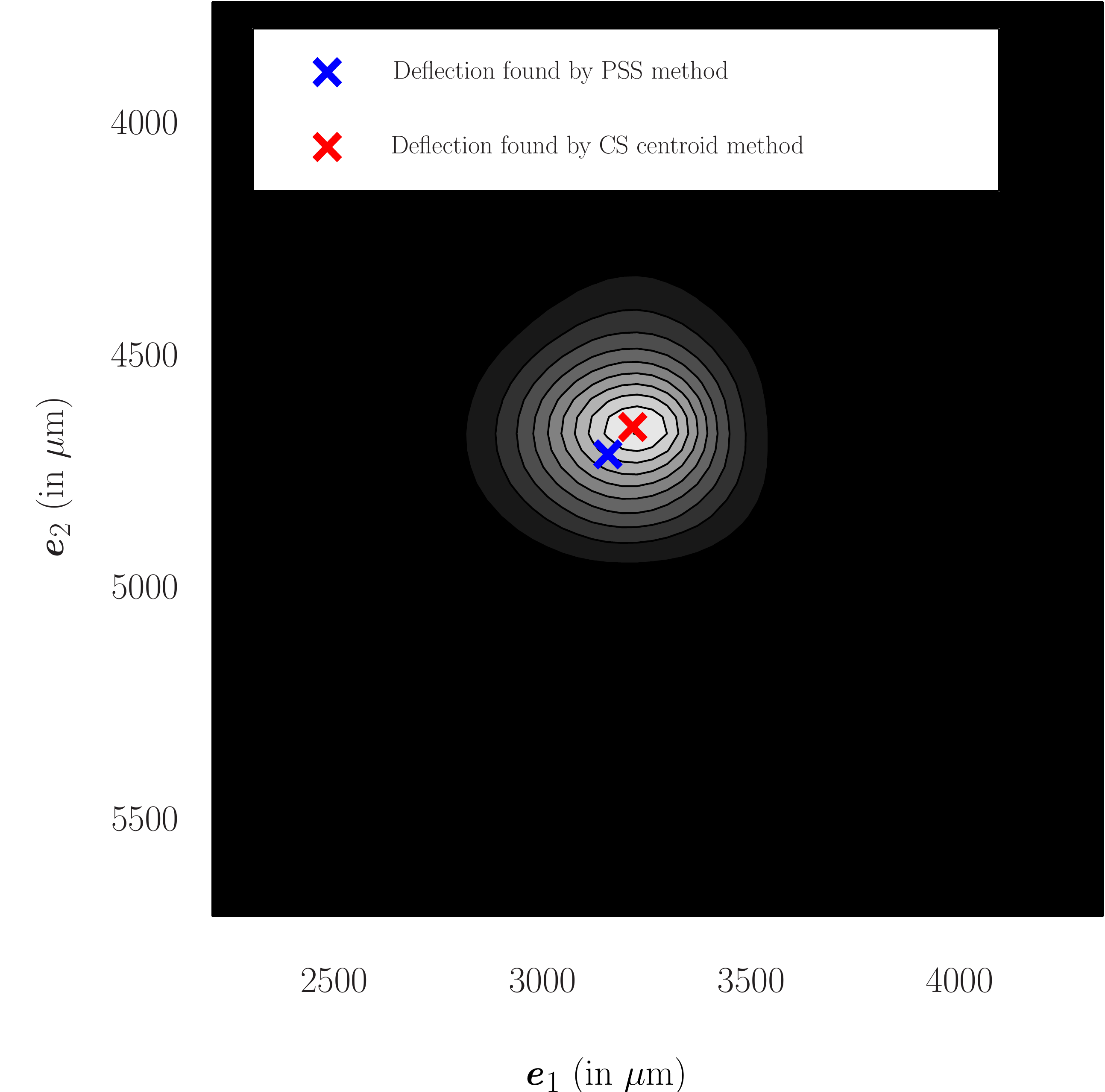}
		\label{subfig:MFD_combinedPlot_zoom_rr1_cc120_M5000}
	}
	%\subfigure[]{
	%\centering
	%\includegraphics[width=0.4\textwidth]{./MFD_profile_m500}
	%\label{subfig:MFD_profile_m500}
	%}
	\caption{\subref{subfig:MFD_profile_m5000} Centroid map along a cross section of a diffractive MIOL with $7.6\%$ measurements. The two arrows refers to the two points where spectra are further analyzed. Example reconstructions of deflection spectrum: (b) for the location indicated by a circle ($\circ$) mark in~\subref{subfig:MFD_profile_m5000} and (c) for the location indicated by a diamond ($\diamond$) mark in~\subref{subfig:MFD_profile_m5000}.}
	\label{fig:MFD_profile}
\end{figure}

\cref{subfig:MFD_3D_m5000} shows the 3D surface plot of centroid
evolution of the considered diffractive MIOL, computed using $\M/\N = 7.63\%$ measurements measurements. The corner artifacts are due to CCD pixels outside the object boundary. The grating
pattern of the MIOL is reproduced in both the surface plots. The central
region which has the same structure as a refractive lens reproduces
the cone like structure observed in~\cref{subsec:geomplanoconvex}. ~\cref{subfig:PSS_MFD_3D} shows the 3D map of the diffractive MIOL obtained by the PSS method described in~\cref{sec:pss}, without disturbing the experimental setup. PSS provides horizontal and vertical deflection values $\bs r^{\text{PSS}}_\k\in\realno^2$ for each CCD pixel $\k$, from which a 3D map is constructed by first computing the norm $\ltwoof{\bs r^{\text {PSS}}_\k}$ and appropriately rescaling them (accounting for unknown PSS algorithmic parameters) to make comparison with the deflection map provided by the compressive sensing method.

\cref{subfig:MFD_profile_m5000} is a cross section\footnote{The cross-section is selected in the horizontal direction to include the minimum.} of the deflection map obtained from CS method using $M/N=7.63\%$ measurements. For comparison, the corresponding cross-section of the deflection map obtained using PSS method is also overlaid. Even though both the methods agree in the refractive region of the lens, the CS method provides a smoother deflection map in the diffractive region compared to the PSS method. The PSS method involves a phase-unwrapping stage where neighbouring pixels influence the deflection information at a given pixel. However, the CS method estimates the deflections independently at each pixel and hence they are more robust and reliable. 

\cref{subfig:MFD_combinedPlot_zoom_rr1_cc104_M5000} shows a zoom{ed contour plot} of
the reconstructed deflection spectrum from $M/N=7.63\%$ measurements for the
location indicated by a circle ($\circ$) mark
in~\cref{subfig:MFD_profile_m5000}. The blue cross mark (X) indicates
the deflection as estimated by the PSS method and the red cross mark
indicates the deflection estimated by the compressive centroid
estimation method. Interestingly, the deflection spectrum clearly has
an elongated profile and appears to be made up of two spots. However,
the PSS method has no way of exploiting this multi-foci structure in its
computations. Similarly,
\cref{subfig:MFD_combinedPlot_zoom_rr1_cc120_M5000}  shows a zoom{ed contour plot}
of the reconstructed spectrum at the location pointed by a diamond ($\diamond$) mark in~\cref{subfig:MFD_profile_m5000}. Here, the deflection spectrum resembles a deflection spectrum of a plano-convex lens. However, the estimate due to the CS method seems more accurate at the centre of the spectral spot than the one provided by the PSS method. This is possibly a reason for the smoothness of the overall deflection map computed using the CS method.

In order to measure the surface smoothness, we compute the Total-Variation (TV) norm of the deflection maps. The TV norm of a 2-dimensional discrete signal $\bs I(x, y)$ is defined as~\cite{books/daglib/0025275, Rudin:1992tm}
%\begin{equation}
$\|\bs I\|_{\text{TV}} = \sum_{x, y}\ltwoof{ \nabla \bs I(x, y)}$,
%\end{equation}
where $\nabla\bs I(x, y)\in\realno^2$ is the gradient (horizontal and vertical) vector of the image at coordinates $(x, y)$. Smaller TV norm means smoother signal.~\cref{fig:MFD_smoothness} plots TV norm (computed by avoiding the corner artifacts) of the deflection map obtained by the CS method as a function of the number of measurements $\M/\N$. For comparison, the TV norm of the deflection map obtained by the PSS method is also plotted. We see that the CS method provides smoother deflection maps as the number of measurements increases. When $\M/\N$ is at least $3\%$, the  CS method produces deflection maps smoother than the one provided by the PSS method. 

Even though the CS method needs more number of measurements to provide a smoother centroid map than the PSS method, richer information about deflections can be obtained via reconstruction, especially when deflection spectrum contains multiple spots. PSS is totally ineffective in such cases. Apart from this advantage, CS method avoids phase unwrapping steps making it robust. Moreover, unlike PSS method, it relies on binary modulation patterns avoiding SLM non-linearities and is amenable for fast implementation with digital micro-mirror arrays, instead of slower SLMs. 

\begin{figure}[t]
	\centering
	\includegraphics[width=0.65\textwidth]{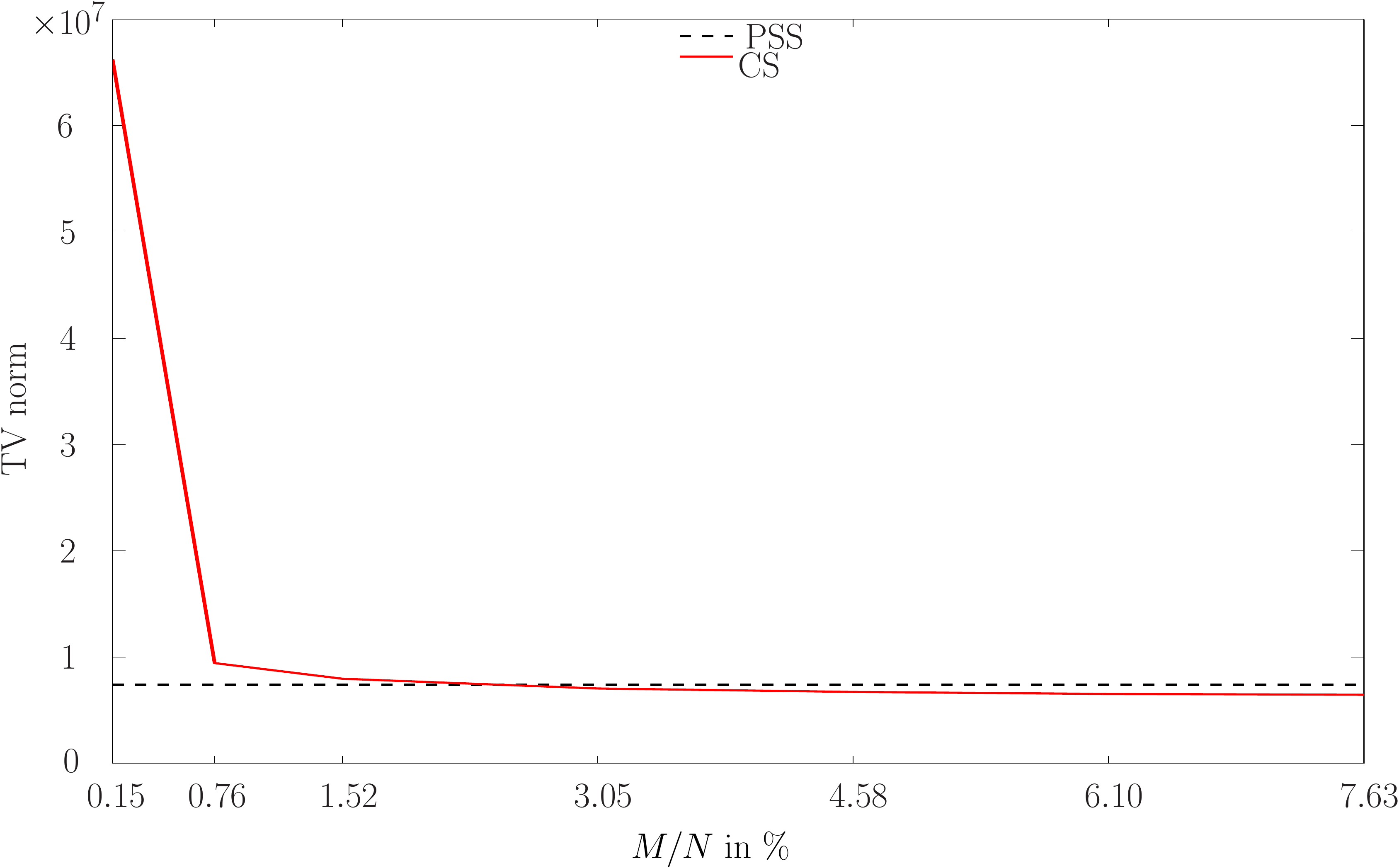}
	\caption{Smoothness of the deflection spectrum computed using the TV norm as a function of the number of measurements.}
	\label{fig:MFD_smoothness}
\end{figure}

% %
\section{{Resolution limit of deflectometry}}
\label{sec:resolution}

{We observe that any linear combination of several measurement vectors, corresponding to distinct CCD locations, is exactly the measurement of same linear combination of the spectra. That is, with $\bs\Psi = \bs I$, if $\bs S = (\bs s_{k_1},
	\cdots, \bs s_{k_L})$ and $\bs Y = (\bs y_{k_1},
	\cdots, \bs y_{k_L})$ are $L$ distinct spectra/measurement vectors
	taken on locations $\{k_j: 1\leq j \leq L\}$, we have $\bs Y = \bs \Phi \bs
	S$. Therefore, given $\bs v \in \bb R^L$,  $\sum_j v_j \bs y_{k_j} =
	\bs Y \bs v = \bs \Phi \bs
	S \bs v = \bs \Phi (\sum_j v_j \bs s_{k_j})$.} 

{In this light, we can explain the nature of elongated spectrum in~\cref{subfig:MFD_combinedPlot_zoom_rr1_cc104_M5000}. Each CCD pixel has a fixed resolution and probes a definite area on the surface of the object. Ideally, this area can be further broken down into smaller areas and we can associate a deflection spectrum to each of this sub-area. However, the CCD pixels at our disposal cannot resolve this finer division and hence at best it probes an \emph{average} spectrum, which is a linear combination of several spectra at smaller resolution. Owing to the relation described in the previous paragraph, each measurement vector at a CCD pixel is to be interpreted as a linear combination of  several measurement vectors associated to finer CCD resolution. The elongated spectrum is a consequence of the limited resolution of the CCD.}

{To substantiate, a \emph{coarse} CCD pixel is simulated by averaging the measurements over a $20\times 1$ neighbourhood of CCD pixels and the reconstruction is performed on the averaged measurement. \cref{fig:CALI2006_multi} shows the reconstructed spectrum using a coarse pixel for a $60D$ plano-convex lens, which consists of an elongated spot due to averaging of several underlying spectra.}
\begin{figure}
	\centering
	\includegraphics[width=0.45\textwidth]{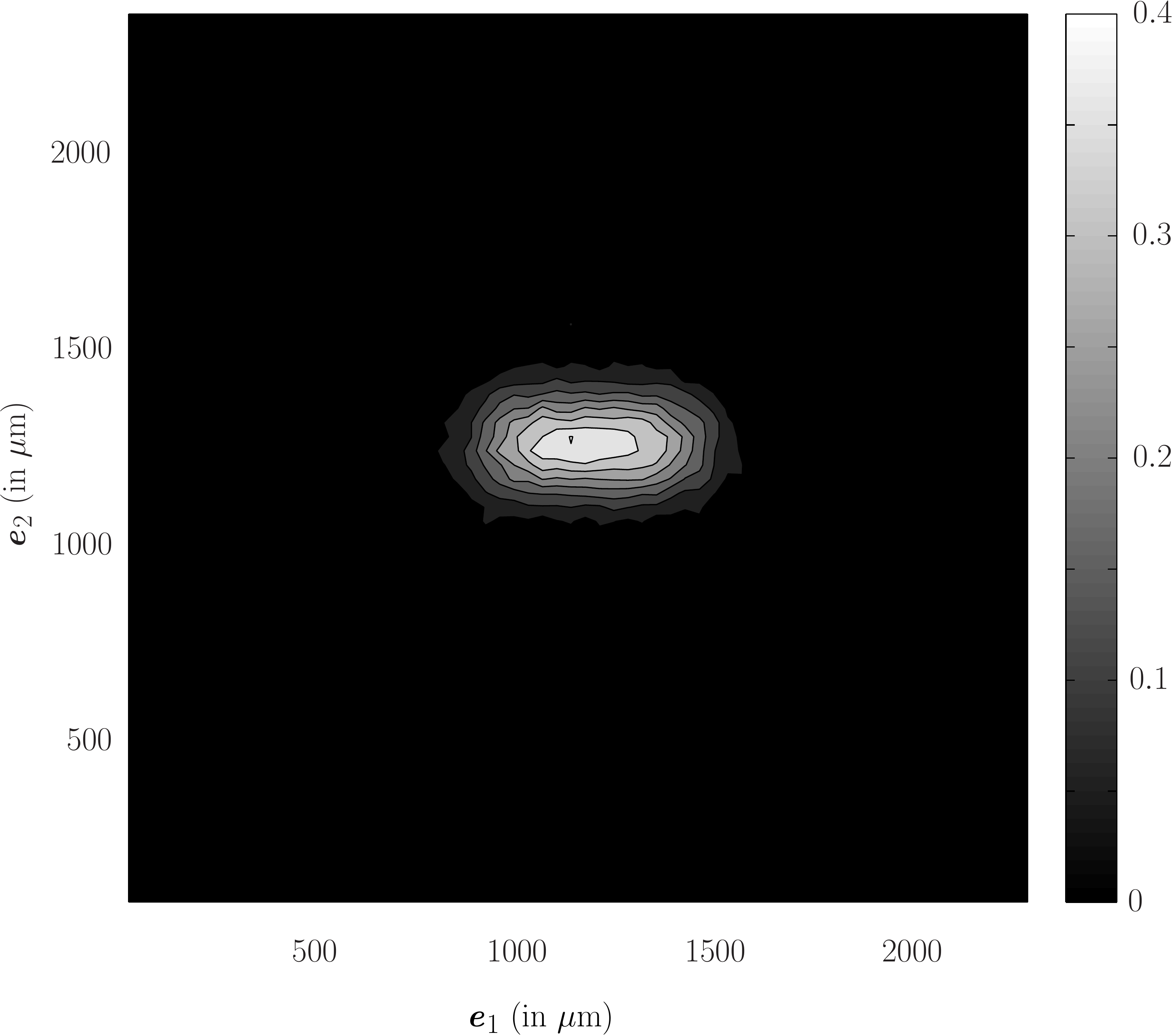}
	\caption{Deflection spectrum reconstructed using a simulated coarse CCD pixel, consisting of 10 true CCD pixels.}
	\label{fig:CALI2006_multi}
\end{figure}
{The deflection spectrum in \cref{subfig:MFD_combinedPlot_zoom_rr1_cc104_M5000} corresponds to a point on the grating edge of MIOL surface. At this location, the mean deflection transits from one angle to another and the CCD has insufficient resolution to distinguish the spectral spots giving rise to an elongated spot. However, thanks to the CS method, such deflection spectra can be visualized via reconstruction while PSS is oblivious to such spectra.}

\section{Conclusions and perspectives}
\label{sec:conclusions}
This paper presents a compressed sensing approach to acquire and reconstruct deflection spectra of transparent objects, using schlieren deflectometry. The design of the sensing matrix considers the practical aspects of optical implementation and also algorithmic implementation. A noise calibration procedure is also described which provides a reasonable bound on the inherent system noise, which is then used to tune the reconstruction algorithm. {To achieve this, we primarily exploit the inherent multiple measurement vector structure of the deflectometric system.} The experimental results show a great reduction in the required number of measurements for a good reconstruction SNR, thereby demonstrating the power of compressed sensing. Moreover, by decoupling the sensing and the reconstruction stages, compressed sensing framework provides a flexibility to tailor the reconstruction method by using appropriate sparsity prior. 

As a second major contribution, the paper contains a method to extract relevant low-dimensional parameters of the deflection spectra that can be directly extracted from the compressive measurements, thereby saving computations. This helps in quick characterization of the shape of the object under study and also provides estimates of optical parameters such as dioptric power. 

A fully reconstructed deflection spectrum is certainly rich in information and contains a lot more than simply a mean deflection angle. Though reconstruction is a computationally intensive task, it can be selectively used to pinpoint features at particular locations on the object. These locations could be guided by a first analysis provided by the compressive characterization which is computationally economical. The accuracy of the compressive characterization method is certainly dependent on the dioptric power of the object as well as the number of measurements one can afford. However, accuracy versus complexity trade-off is always a design choice. 

While the applicability of compressed sensing for deflectometric imaging is very promising, this is at best a good starting point for further exploration. Firstly, the reconstruction method can possibly be improved for speed so that it is viable to reconstruct deflection spectra at all CCD locations, thereby enabling the user to exploit the rich information in deflection spectra. {The multiple measurement vectors model developed in the paper can be further exploited to impose relevant data priors such as the low-rankness in order to simultaneously reconstruct the spectra.} A thorough analysis of the optical system noise also helps in better reconstruction quality. 

While the compressive characterization of deflection spectra presented in this paper is capable of providing reasonable information about the objects, it can be further made more robust by having more parameters in addition to the translation parameters. Modelling the underlying deflection spectra with complex functions that are closer to reality will result in much better estimation of parameters. This requires a thorough understanding of the physics of deflection spectrum formation. 

\section*{Acknowledgements}
Prasad Sudhakar is funded by the DETROIT project (WIST3), convention no. 1017073, Walloon Region, Belgium. Laurent Jacques is supported by the Belgian FRS-FNRS fund. Computational resources were provided by the supercomputing facilities of the Universit\'e catholique de Louvain (CISM/UCL) and the Consortium des \'Equipements de Calcul Intensif en F\'ed\'eration Wallonie Bruxelles (C\'ECI) funded by the F.R.S.-FNRS under convention 2.5020.11.

%	\bibliographystyle{plain}
%	\bibliography{biblio}
	
\end{document}